\documentclass[sn-mathphys-num]{sn-jnl}

\usepackage{xcolor}
\usepackage{graphicx}%
\usepackage{multirow}%
\usepackage{amsmath,amssymb,amsfonts}%
\usepackage{amsthm}%
\usepackage{mathtools}
\usepackage{mathrsfs}%
\usepackage[title]{appendix}%
\usepackage{xcolor}%
\usepackage{textcomp}%
\usepackage{manyfoot}%
\usepackage{booktabs}%
\usepackage{algorithm}%
\usepackage{algorithmicx}%
\usepackage{algpseudocode}%
\usepackage{listings}%
\usepackage{booktabs}
\usepackage{multirow}
\usepackage{subfigure}
\usepackage{caption}
\usepackage{subcaption}
\usepackage{graphicx}
\usepackage{lipsum}
\usepackage{subcaption}
\usepackage{enumitem}
\usepackage{empheq}
\usepackage{tikz}
\usetikzlibrary{calc}

\definecolor{mycustomcolor}{HTML}{6270F5}
\definecolor{mycustomcolor2}{HTML}{F58662}

\theoremstyle{thmstyleone}%
\newtheorem{theorem}{Theorem}
\theoremstyle{thmstyletwo}%
\newtheorem{remark}{Remark}%

\theoremstyle{thmstylethree}%

\definecolor{lightpurple}{rgb}{0.96, 0.96, 1}

\newcommand{\cC}{\mathcal{C}}

\newcommand{\cL}{\mathcal{L}}

\newcommand{\cU}{\mathcal{U}}

\newcommand{\bN}{\mathbb{N}}

\newcommand{\bR}{\mathbb{R}}

\newcommand{\dV}[1][u]{\cU\left(\Omega;\bR^{d_v}\right)}
\newcommand{\dVt}[1][u]{\cU\left(\left[0,\infty\right)\times\Omega;\bR^{d_v}\right)}

\newcommand{\bx}{\mathbf{x}}
\newcommand{\by}{\mathbf{y}}
\newcommand{\bv}{\mathbf{v}}
\newcommand{\bb}{\mathbf{b}}

\DeclareMathOperator*{\argmin}{\arg\min}

\newcommand*\diff{\mathop{}\!\mathrm{d}}

\graphicspath{{./Figures_arXiv/}{./}} 

\begin{document}

\title[Neural shortest path]{Neural Shortest Path for Surface Reconstruction from Point Clouds}

\author[1,2]{\fnm{Yesom} \sur{Park}}\email{yeisom@math.ucla.edu}

\author[3]{\fnm{Imseong} \sur{Park}}\email{parkis@snu.ac.kr}

\author[4]{\fnm{Jooyoung} \sur{Hahn}}\email{jooyoung.hahn@stuba.sk}

\author*[3]{\fnm{Myungjoo} \sur{Kang}}\email{mkang@snu.ac.kr}

\affil[1]{\orgdiv{Department of Mathematics}, \orgname{University of California, Los Angeles}, \orgaddress{\street{Portola Plaza}, \city{LA}, \postcode{90095}, \state{CA}, \country{United States}}}

\affil[2]{\orgdiv{Research Institute of Mathematics}, \orgname{Seoul National University}, \orgaddress{\street{Gwanakro 1}, \city{Seoul}, \postcode{08826}, \country{South Korea}}}

\affil*[3]{\orgdiv{Department of Mathematical Sciences}, \orgname{Seoul National University}, \orgaddress{\street{Gwanakro 1}, \city{Seoul}, \postcode{08826}, \country{South Korea}}}

\affil[4]{\orgdiv{Department of Mathematics and Descriptive Geometry}, \orgname{Slovak University of Technology in Bratislava}, \orgaddress{\street{Radlinskeho 11}, \city{Bratislava}, \postcode{81005}, \country{Slovak Republic}}}


\abstract{
In this paper, we propose the neural shortest path (NSP), a vector-valued implicit neural representation (INR) that approximates a distance function and its gradient. The key feature of NSP is to learn the exact shortest path (ESP), which directs an arbitrary point to its nearest point on the target surface. The NSP is decomposed into its magnitude and direction, and a variable splitting method is used that each decomposed component approximates a distance function and its gradient, respectively. Unlike to existing methods of learning the distance function itself, the NSP ensures the simultaneous recovery of the distance function and its gradient. We mathematically prove that the decomposed representation of NSP guarantees the convergence of the magnitude of NSP in the $H^1$ norm. Furthermore, we devise a novel loss function that enforces the property of ESP, demonstrating that its global minimum is the ESP. We evaluate the performance of the NSP through comprehensive experiments on diverse datasets, validating its capacity to reconstruct high-quality surfaces with the robustness to noise and data sparsity. The numerical results show substantial improvements over state-of-the-art methods, highlighting the importance of learning the ESP, the product of distance function and its gradient, for representing a wide variety of complex surfaces.}


\maketitle

\section{Introduction}
Surface reconstruction from a three-dimensional (3D) point cloud represents a long-standing and fundamental challenge in the fields of computer vision and computer graphics~\citep{berger2013benchmark, berger2017survey}. The aim is to restore a reliable representation of the surface that faithfully reflects the geometric features of the point cloud. Various classical numerical methods have been developed over the years and have been significant in advancing surface reconstruction, thereby providing a solid foundation for surface reconstruction~\citep{bernardini1999ball, carr2001reconstruction, zhao2001fast, dey2003tight, kazhdan2006poisson,kazhdan2013screened, digne2011scale}. Nevertheless, these methods often depend on fixed grid resolutions~\citep{hoppe1992surface,kazhdan2006poisson}, which may require the use of finer grids to capture high-resolution details. As they rely on the given points in triangulation processes, these methods might be sensitive to noise and incomplete data~\citep{hoppe1994piecewise, amenta1998new, gopi2002fast, cazals2006delaunay}. Furthermore, they may require substantial preprocessing, adaptations, or supplementary data, such as surface normal vectors, to handle intricate geometries or incomplete data. 

Recent advances in implicit neural representations (INRs) \citep{park2019deepsdf, mescheder2019occupancy, erler2020points2surf, sitzmann2020implicit, zhang2022critical, chou2022gensdf} have achieved high-quality reconstruction, whereby surfaces are represented as the level set of a learned neural network. This approach has overcome the previous limitations of the classical methods related to resolution and diversity of data, thereby enabling the reconstruction of complex geometries with high-quality. Typically, INRs learn implicit geometry by minimizing a loss function based on partial differential equations (PDEs)~\citep{gropp2020implicit, lipman2021phase, park2024p} or by reducing discrepancies between the estimated surface and the given point cloud~\citep{ma2020neural, zhang2022critical}. Many of these methods are focused on reconstructing watertight closed surfaces, relying on representations by signed distance functions~\citep{gropp2020implicit,ma2020neural,michalkiewicz2019deep} or occupancy fields~\citep{mescheder2019occupancy, peng2020convolutional,mi2020ssrnet} that require dividing the domain into inside and outside regions. This approach restricts applicability in real-world scenarios that often involve shapes with open surfaces, such as garments, automotive styling designs, or 3D-scanned indoor scenes. 

Recent research has explored INR methods capable of representing various shapes~\citep{chibane2020neural, chen20223psdf, zhou2023learning, zhao2021learning, michalkiewicz2019deep, long2023neuraludf, guillard2022meshudf, ye2022gifs}. A distance function has emerged as an effective tool for continuously representing open surfaces, leading to the development of a neural implicit function that approximates a distance function to represent the surface through the zero level set of the learned implicit function~\citep{chibane2020neural, zhao2021learning, venkatesh2021deep, zhou2024cap, long2023neuraludf}. These methods typically use the ground truth distance values or metrics that measure direct distances from the given point cloud. This poses challenges when dealing with incomplete or corrupted data. Approaches that estimate local differential geometric properties, such as the tangent plane and surface normal, encounter difficulties due to the non-differentiability of the distance function at the zero level set. Additionally, since the distance function is a $H^1$ function~\citep{adams2003sobolev} in a bounded domain~$\Omega$, approximating it as the $H^1(\Omega)$ norm is critically essential. The convergence of the gradient of the distance function has not yet been established in many previous studies, which may result in distorted gradients that could negatively affect the accuracy of the approximation. The most closely related work to our study is closest surface-point (CSP)~\citep{venkatesh2021deep} and it relies on estimated ground truth distance. However, it has not yet been shown that the estimated normal vector field is a conservative vector field, which may lead to inaccuracies in gradient learning and further limit the method's performance.

In this paper, we introduce the \textit{neural shortest path} (NSP), a vector-valued INR, that approximates the distance function and its gradient. The key feature of NSP is its parameterization of the exact shortest path (ESP) through a neural network, which attracts an arbitrary point to the nearest point on the target surface. This differentiates the NSP from existing studies that directly approximate a distance function by a network. Applying magnitude-directional decomposition (MDD) of the NSP, which decomposes the vector into its magnitude and direction, we use a variable splitting method~\citep{park2023resdf, park2024p}, assigning the magnitude and direction as the primary and auxiliary variables, respectively, with the constraint that the auxiliary variable represents the gradient of the primary variable. We prove that the integration of MDD and variable splitting method ensures the convergence of the magnitude of the NSP in the $H^1(\Omega)$ norm. We also devise a novel loss function that fully incorporates the properties of the shortest path, showing its global minimum is the desired ESP. The mathematically supported NSP facilitates the accurate approximation of the distance function and its gradient, thereby enabling high-quality representation of a broad range of surface reconstruction from point clouds, including those with open or non-watertight characteristics.

The distance function does not inherently distinguish between inside and outside regions, posing challenges to use traditional surface extraction algorithms such as marching cubes~\citep{lorensen1987marching}. To address this issue, various methods have been developed to effectively extract surfaces from learned distance functions~\citep{chen2022neural, guillard2022meshudf, zhang2023surface, hou2023robust}. Initially, the ball-pivoting algorithm was adapted~\citep{bernardini1999ball, chibane2020neural}, but it proved inefficient with large datasets due to the extensive number of operations required, often resulting in discontinuous and low-quality meshes. Subsequent techniques have incorporated gradient information to artificially induce sign changes for using marching cubes \citep{guillard2022meshudf, zhou2024cap}, or to model tangent planes through least squares solutions~\citep{zhang2023surface}. Nonetheless, the use of gradients near the zero level set, where the distance function becomes non-differentiable, may introduce instability or inaccuracies. In response to these challenges, we provide a new surface extraction algorithm using the structure of NSP. Moreover, the proposed approach takes an advantage of the inherent mesh-free capabilities in deep learning, enabling the evaluation of arbitrary points, thereby overcoming the limitations of mesh dependence typical in classical numerical methods. While further improvements are possible, this algorithm effectively shows our findings and offers a promising solution for surface extraction in computer vision applications.

We conduct comprehensive experiments to evaluate the performance of the proposed NSP across a diverse datasets \citep{bhatnagar2019multi,shapenet2015,zhou2013dense}. The first set of experiments demonstrates that the NSP is capable of accurately learning the distance function and its gradient. Even on simple toy datasets, the NSP shows a superiority quality of results compared to existing approaches. Across a wide range of datasets, the experiments show that the NSP enables high-quality surface reconstructions for complex shapes, including garments, automotive models, and 3D indoor scenes. These experiments include both quantitative and qualitative comparisons with state-of-the-art INR methods. The results clearly indicate that the NSP outperforms existing techniques, providing more seamless shape representations and better preservation of intricate geometric details. Furthermore, we evaluate the robustness of the NSP against noisy and sparse point clouds. Unlike the existing methods, which are sensitive to data corruption, the NSP demonstrates better robustness to noise and sparsity. Collectively, these experimental results validate the robustness and performance of the proposed model in handling incomplete and complex open surfaces, highlighting its potential applicability in real-world scenarios.

\section{Preliminaries}
In this section, we describe the shortest path property of distance functions and review the standard structure of artificial neural networks.
\subsection{Shortest path property}
\begin{figure}
  \begin{center}
    \includegraphics[width=0.30\linewidth]{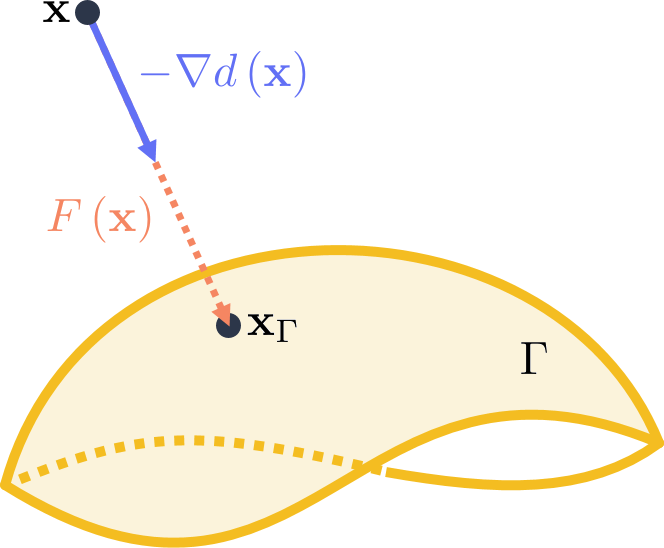}
  \end{center}
  \caption{Illustration of the shortest path property.}
  \label{fig:SP}
\end{figure} 
A distance function $d:\bR^3\rightarrow\bR$ from a bounded surface $\Gamma\subset\bR^3$ is defined as
\begin{equation}\label{eq:def_d}
d\left(\bx\right) = \underset{\bx_{\Gamma}\in\Gamma}{\min}\left\Vert\bx-\bx_{\Gamma}\right\Vert.
\end{equation}
Geometrically, for a given point $\bx\in\bR^3$, if we find the point $\bx_{\Gamma}$ on the surface $\Gamma$ nearest to $\bx$, then the distance  $d\left(\bx\right)=\left\Vert\bx-\bx_{\Gamma}\right\Vert$. The line segment, $\bx_{\Gamma}-\bx$, is the shortest path from $\bx$ to the surface $\Gamma$, also computed by the distance function $d$:
\begin{align}\label{eq:shortest_path}
\bx - \bx_{\Gamma} = d\left(\bx\right)\nabla d\left(\bx\right),
\end{align}
unless $\bx$ is located an equal distance from two or more points in $\Gamma$; see a schematic illustration of the shortest path in Figure~\ref{fig:SP}. Note that the set of equal distance points has measure zero in $\bR^3$. Since the above line segment is uniquely represented by a distanced function $d$~\eqref{eq:def_d}, we define the \textbf{exact shortest path} (\textbf{ESP}) as
\begin{equation}\label{eq:ESP}
	F\left(\bx\right) = d\left(\bx\right)\nabla d\left(\bx\right),
\end{equation}
except for the measure-zero regions where the distance function is not differentiable.
A notable feature of the distance function is that its gradient is of unit length:
\begin{subequations}
\begin{empheq}[left=\empheqlbrace]{align}
  \left\Vert \nabla d \right\Vert=1 & \text{ in } \Omega  \label{eq:eikonal_bc1}\\
 d=0 &  \text{ on }\Gamma\label{eq:eikonal_bc2}
\end{empheq}
\end{subequations}
which is also known as the eikonal equation~\citep{evans2022partial}. When we decompose the ESP into its magnitude and unit vector direction, we call this operation \textbf{Magnitude-Direction Decomposition} (\textbf{MDD}) of $F$,
\begin{align} \label{eq:MDD}
	F(\bx) = \left\Vert F\left(\bx\right)\right\Vert \cdot \frac{F\left(\bx\right)}{\left\Vert F\left(\bx\right)\right\Vert},
\end{align}
where the first term is the distance $d$ and the second term is its gradient $\nabla d$:
\begin{align}
    \left\Vert F\left(\bx\right)\right\Vert &= |d\left(\bx\right)| \cdot \left\Vert \nabla d\left(\bx\right)\right\Vert = |d\left(\bx\right)| \cdot 1 = d\left(\bx\right), \label{eq:reconst_from_SP1}\\     
    \frac{F\left(\bx\right)}{\left\Vert F\left(\bx\right)\right\Vert} &= \frac{d\left(\bx\right) \nabla d\left(\bx\right)}{d\left(\bx\right)} = \nabla d\left(\bx\right). \label{eq:reconst_from_SP2}
\end{align}
We serve the above properties from MDD of ESP as the building block for approximating the ESP by a neural network from point clouds.

\subsection{Neural network}
A multi-layer perceptron (MLP) with input dimension $d_{\text{in}}$ and output dimension $d_{\text{out}}$ is a function $f_{\theta}:\bR^{d_{\text{in}}}\rightarrow\bR^{d_{\text{out}}}$ defined by a composition of the basic unit known as a neural perceptron as follows:
\begin{equation}\label{eq:MLP}
f_{\theta}\left(\bx\right)= W\left(f_L\circ \cdots \circ f_0\left(\bx\right)\right) + \bb,\ \bx\in\bR^{d_{\text{in}}},
\end{equation}
where $L\in\bN$ is a given depth, $W\in\bR^{d_{\text{out}}\times d_{L+1}}$ is a weight of the output layer, and $\bb\in \bR^{d_{\text{out}}}$ is the last bias vector. The perceptron of $\ell$-th hidden layer $f_{\ell}:\bR^{d_{\ell}} \rightarrow \bR^{d_{\ell+1}}$ with $d_{0}=d_{\text{in}}$ is defined by 
\[
f_{\ell}\left(\by\right)=\sigma\left(W_\ell \by + \bb_\ell\right),\ \by\in\bR^{d_{\ell}},\ \text{for all } \ell=0,\dots,L,
\]
where $W_\ell\in \bR^{d_{\ell+1}\times d_{\ell}}$, $\bb_\ell\in\bR^{d_{\ell+1}}$, and a non-linear activation function $\sigma$. 
Since Lipschitz continuous activation functions $\sigma$ are used, the MLP $f_\theta$ is a Lipschitz function and is also bounded on a bounded domain.
The dimensions $d_\ell$ of the hidden layers are typically called by the width of the network. 
The output layer computes the output of the network $f_\theta$ by performing a matrix product between the output of the final hidden layer $f_L$ and $W$, and then adding a bias vector $\bb$.
For simplicity, we use a shorthand notation $\theta$ to denote all parameters, including the weights $\left\{W, W_0,\cdots, W_L\right\}$ and biases $\left\{\bb, \bb_0,\cdots ,\bb_L\right\}$.

Given the current parameter configuration, the parameters $\theta$ are iteratively adjusted by reducing a pre-defined training loss function $\cL\left(\theta\right)$, the minimization of which is typically performed according to the following gradient descent update
\begin{equation} \label{eq:gd}
    \theta_{n+1}=\theta_n-\eta\nabla_{\theta}\cL\left(\theta_n\right),
\end{equation}
where $\eta>0$ is a learning rate parameter.
The gradient calculation of the network in \eqref{eq:gd} is computed with auto-differentiation library (\texttt{autograd}) \citep{paszke2017automatic}.

\section{Proposed method}
Let $\Gamma\subset\bR^3$ be an open or closed bounded surface and $\Omega\subset\bR^3$ be a bounded, convex, and simply connected computational domain containing $\Gamma$. We aim to implicitly represent $\Gamma$ as the zero level set of distance function $d$ in $\Omega$ from an unorganized point cloud sampled from $\Gamma$.

\subsection{Neural shortest path}\label{sec:NSP}

\begin{figure}
	\begin{center}
		\includegraphics[width=0.80\textwidth]{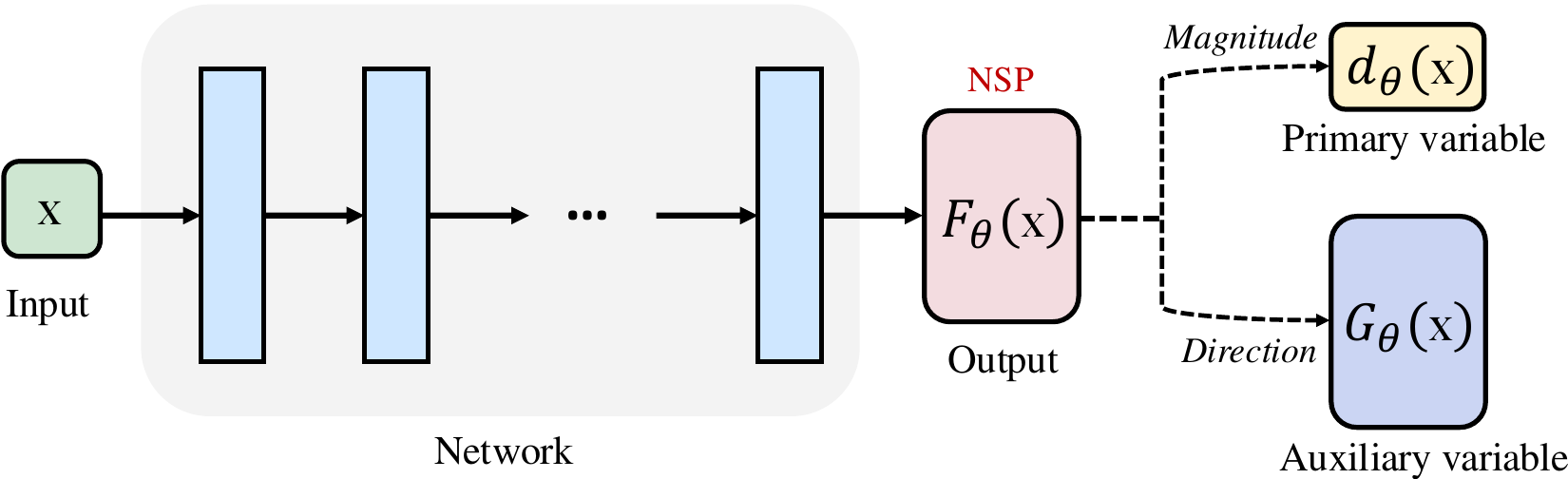}
	\end{center}
	\caption{The visualization of the network architecture of NSP with variable splitting.}
	\label{fig:architecture}
\end{figure}

We propose a vector-valued INR $F_{\theta}$ utilizing an MLP~\eqref{eq:MLP} with learnable parameters $\theta$ to learn the ESP \eqref{eq:ESP}. Through MDD, we decompose the output $F_{\theta}$ of the MLP into its magnitude and direction:
\begin{align}\label{eq:NSP}
	F_{\theta}:\bR^3\rightarrow\bR^3 \:\: \text{by} \:\: F_{\theta}\left(\bx\right) = \left\Vert F_{\theta}\left(\bx\right)\right\Vert \cdot \frac{F_{\theta}\left(\bx\right)}{\left\Vert F_{\theta}\left(\bx\right)\right\Vert}.
\end{align}
In order for MDD to be well-defined, we consider $F_\theta$ to be nonzero a.e. in $\Omega$.
The magnitude is designated as the primary variable, which we enforce to approximate the distance function in Section~\ref{sec:loss_ftn}. The direction is referred to as the auxiliary variable in the variable splitting method~\citep{park2024p, park2023resdf} to impose the property~\eqref{eq:reconst_from_SP2}:
\begin{align}\label{eq:NSP_form_1}
	F_{\theta}\left(\bx\right) = \left\Vert F_{\theta}\left(\bx\right)\right\Vert \cdot \frac{F_{\theta}\left(\bx\right)}{\left\Vert F_{\theta}\left(\bx\right)\right\Vert} = d_\theta\left(\bx\right) \cdot G_\theta \left(\bx\right), \quad \text{w.r.t.} \quad G_\theta \left(\bx\right) = \nabla d_\theta\left(\bx\right).
\end{align}
Then, the proposed INR exhibits the following characteristic:
\begin{align}\label{eq:NSP_form_2}
	F_{\theta}\left(\bx\right) = d_\theta\left(\bx\right) \nabla d_\theta\left(\bx\right).
\end{align}
With an analogy to the established form in the ESP~\eqref{eq:ESP}, we call the first output $F_{\theta}$ of the proposed network as \textbf{neural shortest path} (\textbf{NSP}); see Figure~\ref{fig:architecture} for a pictorial representation of the proposed NSP. Note that the constraint condition in~\eqref{eq:NSP_form_1} immediately shows that the magnitude of NSP satisfies the following eikonal PDE in \eqref{eq:eikonal_bc1}:
\begin{align}\label{eq:PDE_part_EE}
\left\Vert \nabla \left\Vert F_{\theta}\left(\bx\right)\right\Vert \right\Vert= \left\Vert \nabla d_\theta\left(\bx\right) \right\Vert = \left\Vert G_\theta \left(\bx\right) \right\Vert = \left\Vert \frac{F_{\theta}\left(\bx\right)}{\left\Vert F_{\theta}\left(\bx\right)\right\Vert} \right\Vert = 1.
\end{align}
Let us consider that the viscosity solution of the eikonal equation \eqref{eq:eikonal_bc1} and \eqref{eq:eikonal_bc2} is a distance function and take into account the relationship above. Then, we can see that in order for the magnitude of the NSP, $d_\theta = \left\Vert F_{\theta} \right\Vert$, to approximate a distance function, the boundary condition of the eikonal equation is necessarily required, that is, $d_\theta\left(\bx\right) = 0$, $\bx \in \Gamma$. In Section~\ref{sec:loss_ftn}, we impose this aspect when designing the loss function.

The proposed INR approach differs from previous INR methods by neither solely learning the distance function~\citep{chibane2020neural,zhou2024cap}, nor separately learning the distance function and its gradients as outputs from two different networks~\citep{park2024p}. Instead, it directly learns $F_\theta$ in a manner that enables the model to capture the property of ESP \eqref{eq:ESP}. By representing the ESP with a single $F_\theta$, we not only relax the product representation of the ESP but also simultaneously approximate the distance function and its gradient through the learned NSP. From the construction derived from the properties~\eqref{eq:reconst_from_SP1} and~\eqref{eq:reconst_from_SP2}, it follows that $d_\theta$ automatically becomes a non-negative function and $G_\theta$ has a unit length. The proposed network architecture inherently imposes hard constraints on characteristics of the distance function and its gradient into variables $d_\theta$ and $G_\theta$.

One of the main advantages of the proposed NSP is the effective control of numerical issues related to non-smoothness at the zero level set. Kinks, discontinuities in the first derivatives, always appear at the zero level set in the distance function, representing the target surface $\Gamma$. These kinks pose a substantial numerical challenge for neural networks approximating distance functions, leading to instability and significant errors near the surface, thus making inaccurate representation of the zero level set. The NSP effectively addresses this issue because of its formulations~\eqref{eq:NSP_form_1} and~\eqref{eq:NSP_form_2}. As $\bx \rightarrow \Gamma$, we have $d_\theta\left(\bx\right) \rightarrow 0$ and $\left\Vert G_{\theta}\left(\bx\right)\right\Vert = 1$, therefore $F_{\theta}\left(\bx\right)$ approaches to the zero vector. In the vicinity of the zero level set, issues of instability and significant errors are avoided in the NSP, enabling stable learning throughout the training process.

\subsection{Loss function} \label{sec:loss_ftn}
In this section, we devise the loss function in order to enforce, one of outputs in the NSP, $d_\theta$, in Figure~\ref{fig:architecture} to satisfy the characteristics of the distance function from $\Gamma$ and to impose the constraint condition in~\eqref{eq:NSP_form_1}:
\begin{equation}\label{eq:total_loss}
    \int_\Gamma \left\vert d_\theta\left(\bx\right)\right\vert \diff\bx  + \lambda_{\text{GM}}\int_\Omega \left\Vert \nabla d_\theta\left(\bx\right) - G_\theta\left(\bx\right) \right\Vert^2 \diff\bx
    + \lambda_{\text{SP}}\int_\Omega \Bigl\vert d_{\theta}\left( \bx-F_{\theta}\left(\bx\right)\right)\Bigr\vert^2\diff\bx,
\end{equation}
where $F_{\theta} = d_{\theta} G_{\theta}$, $d_{\theta} = \left\Vert F_{\theta} \right\Vert$, $G_{\theta} = \frac{F_{\theta}}{\left\Vert F_{\theta} \right\Vert}$, and $\lambda_{\text{GM}}$ and $\lambda_{\text{SP}}$ are positive constants. Note that Theorem~\ref{thm:global_min} proves that a minimizer of the proposed loss functional \eqref{eq:total_loss} is an ESP from $\Gamma'$, where $\Gamma \subseteq \Gamma'$. The main characteristics of each loss terms are summarized:
\begin{itemize}
\item The first term enforces $d_\theta$ to vanish on the target surface $\Gamma$.
\item The second term imposes the constraint condition in~\eqref{eq:NSP_form_1}.
\item The third term enforces the network to satisfy the property~\eqref{eq:shortest_path}.
\end{itemize}
Note that \eqref{eq:total_loss} is nonconvex and nonlinear. The Euler-Lagrange equation, which represents the optimality conditions for \eqref{eq:total_loss}, is highly nonlinear. As a result, it is not straightforward to find a minimizer using classical optimization methods. This situation underscores an advantage of the deep learning approach, known for its effectiveness in minimizing complex loss functions.

\paragraph{Manifold loss}
Manifold loss function,
\begin{equation}\label{eq:mnfld_loss}
    \cL_\Gamma\left(\theta\right)=\int_\Gamma \left\vert d_\theta\left(\bx\right)\right\vert \diff\bx,
\end{equation}
encourages $d_\theta$ to vanish on the target surface $\Gamma$, which means the boundary condition of the eikonal equation \eqref{eq:eikonal_bc2}. In practical implementation, $\cL_\Gamma$ is defined by means of a Monte Carlo approximation of the integral over the given point cloud.

\paragraph{Gradient matching loss}
Gradient matching loss function,
\begin{equation}\label{eq:GMloss}
	\cL_{\text{GM}}\left(\theta\right)=\int_\Omega \left\Vert \nabla d_\theta\left(\bx\right) - G_\theta\left(\bx\right) \right\Vert^2 \diff\bx,
\end{equation}
is the $L^2$ penalization of the constraint:
\begin{equation}\label{eq:constraint_VS}
	\nabla d_\theta  = G_\theta.
\end{equation}
As we already mentioned in Section~\ref{sec:NSP}, it is the constraint to have the key characteristic of the NSP~\eqref{eq:NSP_form_2}. 
Given that $G_\theta$ has been structured with a unit length, minimizing the gradient matching loss function $\cL_{\text{GM}}$ would ensure that $\nabla d_\theta$ also has a unit length, thereby satisfying the relation~\eqref{eq:PDE_part_EE} in the eikonal equation~\eqref{eq:eikonal_bc1}.

\paragraph{Shortest path loss}
Shortest path loss function,
\begin{equation}\label{eq:SP_loss}
	\cL_{\text{SP}}\left(\theta\right)= \int_\Omega \Bigl\vert d_{\theta}\left( \bx-F_{\theta}\left(\bx\right)\right)\Bigr\vert^2\diff\bx,
\end{equation}
encourages a point pulled by $F_{\theta}\left(\bx\right)$ to be a point in $\Gamma$:
\begin{equation}
	\bx'=\bx-F_{\theta}\left(\bx\right)\in\Gamma.
\end{equation}
More importantly, the point $\bx'$ in $\Gamma$ is directed to become the nearest point from $\bx$, as a result of the proposed NSP construction. Two components in MDD of $F_{\theta}$, its magnitude $d_{\theta}$ and direction $G_{\theta}$, are enforced to satisfy the characteristics of distance function and its gradient; see~\eqref{eq:NSP_form_1}, and eventually drives the property $F_{\theta} = d_{\theta} \nabla d_{\theta}$~\eqref{eq:NSP_form_2} with \eqref{eq:GMloss}. 
By deriving points $\bx'$ pulled by the NSP to lie on the zero level set of $d_\theta$, the NSP is enforced to be the ESP and satisfy the property~\eqref{eq:shortest_path} by minimizing the shortest path loss function~\eqref{eq:SP_loss}. The manifold loss~\eqref{eq:mnfld_loss} enforces $d_{\theta} \left(\bx\right) = 0$ only for $\bx \in \Gamma$, where $\Gamma$ is the measure zero in $\bR^3$. Comparing to the manifold loss, the shortest path loss enforces much stronger condition because, for $\bx \in \Omega \setminus \Gamma$, where is not a measure-zero region, $d_{\theta} \left(\bx'\right) = 0$ as long as the point $\bx'$ is pulled by  $F_{\theta}\left(\bx\right)$ from $\bx$. The shortest path loss makes a crucial rule to prove Theorem~\ref{thm:global_min}.
 
In practical implementation, the loss function is computed as follows:
\begin{equation}\label{eq:NSP_loss_pract}
    \int_\Omega \Bigl\vert d_{\hat{\theta}}\left( \bx-F_{\theta}\left(\bx\right)\right)\Bigr\vert^2\diff\bx,
\end{equation}
where $\hat{\theta}$ indicates that the gradient has been detached from the current computational graph. The form of $\cL_{\text{SP}}$ is inherently complex, comprising the composition of the network within the network. This intrinsic complexity, coupled with the use of stacked networks, presents significant challenges for optimization. To address these challenges, an actual implementation employs the disabling of the gradient calculation of the outer $d_\theta$. Subsequently, the computational graph is no longer associated with the parameters of $d_{\hat{\theta}}$, and it is regarded as a frozen function within the loss function.

One of the advantages to use shortest path loss function is that this method reduces sensitivity to the point cloud. Unlike to a loss function based on the explicit distance between the pulled points and the given point cloud \citep{chibane2020neural,venkatesh2021deep,zhou2024cap}, our approach implicitly imposes that the pulled points reach to the zero level set of the learned distance, which conforms to the target surface. Experimental evidence supporting this is provided in Section~\ref{sec:exp_corruption}. We also note that $d_\theta$ at arbitrary points $\bx-F_\theta\left(\bx\right)$ can be exactly evaluated by the deep learning model of the NSP. However, grid-based numerical methods encounter the approximation error to evaluate a function value at an arbitrary point which is not in the grids. 

\subsection{Mathematical properties} \label{sec:Math_prop}

\paragraph{Convergence in the $H^1(\Omega)$ norm}
In this section, we demonstrate the the $H^1(\Omega)$ norm convergence of the magnitude of the proposed NSP. A distance function lies in $H^1(\Omega)$ space in a domain $\Omega \subset \bR^3$:
\begin{align}
	H^1\left(\Omega\right) = \left\{ f \in L^2\left(\Omega\right) \mid \partial_i f \in L^2\left(\Omega\right),\ i=1,2,3 \right\},
\end{align}
where $\partial_i f$ is the weak derivative of $f$ with respect to the $i^{\text{th}}$ axis of $\Omega$. In order to achieve an accurate approximation of the distance function, it is essential to approximate it in $H^1(\Omega)$ norm:
\begin{align}
	\left\Vert f\right\Vert_{H^1(\Omega)} = \left( \left\Vert f\right\Vert_{L^2(\Omega)}^2 + \sum_{i=1}^{3} \left\Vert \partial_i f\right\Vert_{L^2(\Omega)}^2 \right)^{1/2}.
\end{align}
We demonstrate that the representation of the primary variable of NSP in~\eqref{eq:NSP_form_1} in our approach ensures the convergence of the primary variable in the $H^1(\Omega)$ norm. The proof of the theorem below is provided in Appendix \ref{appen:pf_Sobolev}. 

\begin{theorem}\label{thm:Sobolev_convergence}
Let $F_n:\Omega\rightarrow\bR^3$ be a uniformly bounded sequence of $H^1\left(\Omega\right)$ functions which are nonzero a.e. in $\Omega$ and consider the MDD $d_n\coloneqq \left\Vert F_n\right\Vert$ and $G_n\coloneqq \frac{F_n}{\left\Vert F_n\right\Vert}$ for $n\in\bN$.
If $F_n$ converges a.e. in $\Omega$, both $d_n$ and $G_n$ converge a.e. in $\Omega$. Moreover, if $\left\Vert \nabla d_n-G_n\right\Vert_{L^2\left(\Omega\right)}$ converges to zero, then $d_n$ converges in $H^1\left(\Omega\right)$ norm as $n\rightarrow\infty$.
\end{theorem}

In the context of training a deep learning model, the network parameters $\theta$ are typically updated recursively, generating a sequence of parameters $\left\{\theta_n\right\}_{n\in\bN}$ through a gradient-based optimizer. The NSP is trained through an iterative process whereby a sequence of network functions $\left\{F_{\theta_n}\right\}_{n\in\bN}$ minimizing the loss function is generated. The above theorem implies that, when the sequence $F_{\theta_n}$ converges as $n\rightarrow\infty$, the convergence of two components in MDD of $F_{\theta_n}$ -- the magnitude $d_{\theta_n}$ and the direction $G_{\theta_n}$ -- is also ensured. Additionally, if the sequence of parameters $\theta_n$ is learned to minimize the gradient matching loss $\left\Vert \nabla d_{\theta_n}-G_{\theta_n}\right\Vert_{L^2\left(\Omega\right)}$ as zero, the convergence of the gradient of the $d_{\theta_n}$ is achieved. It means the sequence $\left\{d_{\theta_n}\right\}_{n\in\bN}$ converges in the $H^1(\Omega)$ norm, that is, the limit function lies in $H^1(\Omega)$ space where a distance function exists.

In deep learning models, the completion of training is typically assessed by two main indicators: the convergence of the loss function towards zero and the convergence of the network itself. In existing studies of approximating a distance function in INR, the convergence of the network's derivatives has not been proved yet. This means that even if the network functions show convergence and the mentioned indicators are satisfied, there is no guarantee that the limit function will accurately approximate a function in the space where the distance function exists.

Theorem~\ref{thm:Sobolev_convergence} also explains that the proposed NSP approach enables the $H^1(\Omega)$ norm convergence of the primary variable in NSP without an additional higher-order derivatives. Sobolev training \citep{czarnecki2017sobolev, son2021sobolev}, which defines the loss function in terms of Sobolev norms, has been considered in INR for surface reconstruction~\citep{yang2023steik, wang2024aligning} in order to enhance the regularity of the predicted solution. However, this approach presents challenges in optimization and increases computational costs caused by using higher-order derivatives in the loss function. In \citep{park2024beyond}, a similar Sobolev convergence of variable splitting was established. However, unlike to our approach, which uses a single network, their method relies on separate networks for approximating the primary and auxiliary variables, thus depending on the boundary conditions of the considered PDE.

\paragraph{Global minimum}
Combining all loss functions in Section~\ref{sec:loss_ftn}, we propose a loss function for NSP defined as a weighted sum of them:
\begin{equation}\label{eq:NSP_loss}
   \cL_{\text{NSP}}\left(F_\theta\right)= \cL_{\text{NSP}}\left(\theta\right)=\cL_\Gamma\left(\theta\right)  + \lambda_{\text{GM}}\cL_{\text{GM}}\left(\theta\right)
   + \lambda_{\text{SP}}\cL_{\text{SP}}\left(\theta\right),
\end{equation}
where $\lambda_{\text{GM}}$ and $\lambda_{\text{SP}}$ are positive constants that determine the extent to which each loss term regulates the network during the training process. Now that the loss function has been introduced, it is essential to determine whether optimizing the proposed loss function $\cL_{\text{NSP}}$ actually leads to the desired ESP. 
Rewriting the the minimization of \eqref{eq:NSP_loss}
\begin{equation}\label{eq:minimization_NSP}
  F^\ast = \text{argmin}\ \left\{\cL_{\text{NSP}}\left(F\right) \mid F\in H^1\left(\Omega,\bR^3\right),\ F\neq 0 \text{ a.e. in } \Omega\right\},
\end{equation}
the following theorem demonstrates that a global minimizer of \eqref{eq:minimization_NSP} is the ESP a.e. in $\Omega$. The proof can be found in Appendix \ref{appen:pf_global_min}; see also a similar proof in \citep{park2023resdf}.
\begin{theorem}\label{thm:global_min}
An optimal solution $F^*$ in \eqref{eq:minimization_NSP} is an ESP to $\Gamma'$ which contains $\Gamma$, except on a set of measure zero.
\end{theorem}

\begin{remark}
In the context of PDEs, the distance function $d$~\eqref{eq:def_d} from $\Gamma$ is the unique viscosity solution of the eikonal equation~\eqref{eq:eikonal_bc1} and \eqref{eq:eikonal_bc2} on the domain $\Omega\subset\bR^3$. A possible attempt to obtaining a distance function by neural networks is to use physics-informed neural networks (PINNs), which utilize the residual of the eikonal equation as a loss function:
\begin{equation}\label{eq:eikonal_energy}
    \int_\Omega \left(\left\Vert \nabla d \left(\bx\right)\right\Vert - 1\right)^p \diff\bx,
\end{equation}
for $p=1$ or $2$. However, due to the ill-posed nature of the eikonal equation, which has multiple solutions, minimization of PINN loss \eqref{eq:eikonal_energy} does not guarantee the viscosity solution of the eikonal equation; see more details in~\citep{park2024p}.
\end{remark}

\paragraph{Minimal area regularization}
In addressing the surface reconstruction from point clouds, we additionally introduce a regularization term in our loss function~\eqref{eq:NSP_loss} that enforces the minimal area (MA) of the zero level set of the trained distance function $d_\theta$:
\begin{equation}\label{eq:loss_final}
    \cL _\text{total}\left(\theta\right)= \cL_{\text{NSP}}\left(\theta\right)+ \lambda_{\text{MA}} \int_\Omega \delta_\epsilon\left(d_\theta\left(\bx\right)\right) \diff\bx,
\end{equation}
where $\lambda_{\text{MA}}$ is a positive constant and $\delta_\epsilon(x) = 1-\tanh^2\left(\frac{x}{\epsilon}\right)$ is a smeared Dirac delta function with $\epsilon >0$.

The minimal area constraint arises from the ill-posed nature of surface reconstruction from point clouds, which inherently admits multiple solutions. The point samples only provide partial information, making it impossible to definitively determine the surface’s behavior in unsampled regions. As mentioned in Theorem~\ref{thm:global_min}, the distance function from any surface $\Gamma'$ containing the target surface $\Gamma$ can serve as an optimal solution. Particularly in the context of open surfaces, it is crucial to ensure that the learned function does not close off opened areas. By minimizing the area of the zero level set, we effectively target the actual gaps in the point cloud, enhancing the fidelity of the surface reconstruction. This approach, validated in previous works such as~\citep{fuchs1977optimal, pumarola2022visco} and further advanced by~\citep{Caselles20008Geom, he2020curvature, zhang2022critical, park2024p}, has proven effective in managing incomplete data and preventing the formation of incorrect surfaces.

\subsection{Shortest path-based surface extraction}\label{subsec:extract_alg}
In this section, we propose a surface extraction algorithm to use the strength of NSP. The algorithm is briefly summarized as follows:
\begin{enumerate}[label=S\arabic*:,ref=S\arabic*]
\item \textit{Rough Filtering of Cubic Cell}: Roughly filter out cubic cells that have no intersection with the target surface, based on the values obtained from the learned distance function.\label{step:first}
\item \textit{Determination of Cells Containing the Surface}: Identify the cells that include the target surface and determine representative points of the surface based on shortest path property~\eqref{eq:shortest_path}. \label{step:second}
\item \textit{Surface Mesh Extraction}: Apply dual contouring to extract the surface mesh from the determined representative points of the surface. \label{step:third}
\end{enumerate}
We now provide a detailed explanation of these steps.

\paragraph{Rough filtering of cubic cell}
We construct a discrete, regular grid by dividing the region of interest, typically 
$\Omega = \left[-1,1\right]^3$, into $N^3$ cubic cells. This division is achieved by creating $N$ equal subdivisions along each coordinate axis within the domain $\Omega$. Prior to the detailed selection of cells, a preliminary step is undertaken to roughly identify cells that potentially intersect with the surface, to reduce the computational cost in subsequent steps. For each cubic cell, if the values of learned distance function $d_\theta$ at all eight vertices exceed a specified threshold $\eta$,
\begin{equation}
	d_\theta\left(\bv\right) \geq \eta, \:\: \text{for all vertices $\bv$ of the cell},
\end{equation}
then the cell is treated as containing no part of the surface and is excluded from further consideration. Let us denote a collection of selected cells as $\cC=\left\{c_k\right\}$. In the implementation, we set $N=256$ and $\eta=2/N$.

\paragraph{Determination of cells containing the surface}
\begin{figure}
	\centering
	\includegraphics[width=0.4\linewidth]{
		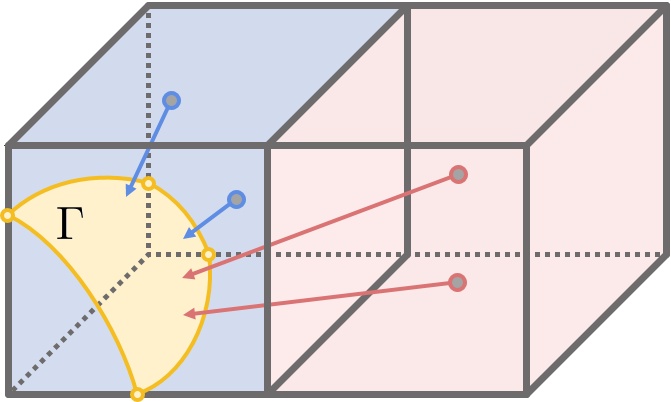}
	\caption{Illustration of shortest path-based cell determination of the surface extraction algorithm in Section \ref{subsec:extract_alg}. For points in the right red-colored cell, if all points pulled by the NSP stay out of the cell, then the cell does not contain the surface $\Gamma$. For a point in the left blue-colored cell, if the point pulled by the NSP stays in the cell, then the cell intersects with $\Gamma$.}
	\label{fig:surface_extraction}
\end{figure}
We determine whether the selected cells $c_k\in\cC$ contain a part of the surface $\Gamma$, that is the zero level set of $d_\theta$, by using the learned NSP $F_\theta$. The key property of the shortest path~\eqref{eq:shortest_path} is that a point pulled by the ESP, $\bx - F\left(\bx\right)$, is on the surface $\Gamma$. Based on this property, for a cell $c_k\in\cC$, if all points $\bx\in c_k$ pulled by the NSP, $\bx - F_\theta\left(\bx\right)$, stay out of the cell, we conclude that the cell $c_k$ does not intersect $\Gamma$. Conversely, if a point $\bx\in c_k$ pulled by the NSP stays in the cell $c_k$, we identify the pulled point $\bx-F_\theta\left(\bx\right)$ which has the minimum distance value $d_\theta$ as a representative point of the surface in the cell $c_k$; see Figure~\ref{fig:surface_extraction} for a pictorial explanation.

In the practical implementation, for a cell $c_k\in\cC$, we randomly sample $n$ number of points $\bx_k^i \in c_k$, $i=1,\ldots,n$. We also consider a slightly enlarged bounding cube $\tilde{c}_k$, specifically increasing the size of the cubic cell $c_k$ by $7\%$. If all pulled points stay out of the cell, that is, $\bx_k^i - F_\theta \left( \bx_k^i \right) \notin \tilde{c}_k$, then we discard $c_k$. Otherwise, we define the representative point that presents the surface within the cell of $c_k$ as follows as 
\begin{align}\label{eq:dual_pt}
p_k =\argmin\left\{ d_\theta\left(\by_k^i\right) \mid \by_k^i = \bx_k^i - F_\theta \left( \bx_k^i \right) \in \tilde{c}_k, \: i = 1, \ldots, n\right\},
\end{align}
which serves a pulled point to give the minimum of $d_\theta$. The enlarged bounding cube, $\tilde{c}_k$, adjusts the criteria for determining point inclusion and exclusion, making it less strict for excluding a cell but more rigorous for including points within it. The enlargement allows for a more inclusive check that can account for small discrepancies in point location due to slight variations in how the surface is represented. If a point, after being pulled by $F_\theta$, still falls within this larger cube, it still suggests a possible likelihood that the original cell, $c_k$, intersects with the target surface $\Gamma$. This method reduces the risk of prematurely discarding cells that do actually intersect with the surface due to minor sampling errors. In numerical experiments, we use $n=200$.

In this approach, we use the learned NSP rather than the gradient of the distance function. This facilitates more stable surface extraction compared to existing surface extraction algorithms tailored to distance functions~\citep{chen2022neural, guillard2022meshudf, zhang2023surface, hou2023robust}, which rely on the gradient of the learned distance function. The use of the gradient may result in inaccuracies and instability due to kinks at the zero level set. 
As mentioned in Section~\ref{sec:NSP}, the NSP, defined as $d_\theta G_\theta$, where $G_\theta$ is a unit vector auxiliary variable that approximates $\nabla d_\theta$, ensures that the computed NSP remains stable near kinks at the zero level set. By employing this strategy, we achieve a reliable choice of representative points~\eqref{eq:dual_pt}, leading to a high-quality surface mesh that conforms precisely to the target geometry. This method not only preserves geometric details of the surface but also significantly reduces errors in areas where existing gradient-based approaches often encounter difficulties.

\paragraph{Surface mesh extraction}
To construct the surface mesh from the selected representative points~\eqref{eq:dual_pt}, we adopt the principles of Dual Contouring in~\citep{ju2002dual}. For an edge of $c_k \in \mathcal{C}$, if $p_{k'}$ exists for all four $c_{k'} \in \mathcal{C}$ sharing the edge, then the four representative points are connected to generate a quad-face. The quad faces are then validated for correct connectivity and triangulated. When a quad-face is not a plane, there is an ambiguity for two triangulations of the quad-face. In order to reduce such ambiguity for better visualization and evaluation of the performance of the proposed surface mesh extraction, the most standard and simple local smoothing technique, Laplacian smoothing, is adopted to repair poor-quality meshes by adjusting the grid point placements without modifying the mesh topology and to reduce possible irregularities or highly distorted elements. In the implementation, we use the computationally inexpensive \texttt{trimesh library} (2019)\footnote{\url{https://github.com/mikedh/trimesh}} for this purpose.

\section{Numerical experiments}
This section presents a numerical evaluation of the proposed NSP. In Section~\ref{sec:exp_toy}, we begin by assessing the effectiveness of the NSP in learning the distance function and its gradient. Sections~\ref{sec:exp_synthetic} and \ref{sec:exp_indoor} demonstrate the capability of NSP in surface reconstruction for various point cloud datasets through experimental validation. Additionally, Section~\ref{sec:exp_corruption} evaluates the robustness of the NSP to noise and data sparsity. All experiments are compared with leading INR models. Implementation details are provided in the Appendix \ref{appen:exp_detail}.

\subsection{Evaluation on toy examples}\label{sec:exp_toy}
To evaluate the effectiveness of the proposed NSP in accurately learning distance function in the $H^1$ sense, we start with experiments on simple toy datasets. To show the advantages of distance function in managing open surfaces, we present simple three-dimensional open surfaces. In particular, we generate a dataset consisting of $1,000$ random points sampled from a hemisphere.
\[
\left\{\left(x,y,z\right)\in\bR^3 \bigm| x^2+y^2+z^2=0.6^2,\ z\geq 0\right\},
\]
and from a partial cylinder with its side cut away, which is given by
\[
\left\{\left(r,\theta,z\right) \,\middle|\, r=0.5, 0\leq\theta\leq\frac{3}{2}\pi, -0.5\leq z\leq 0.5 \right\}
\]
in the cylinderical coordinate.

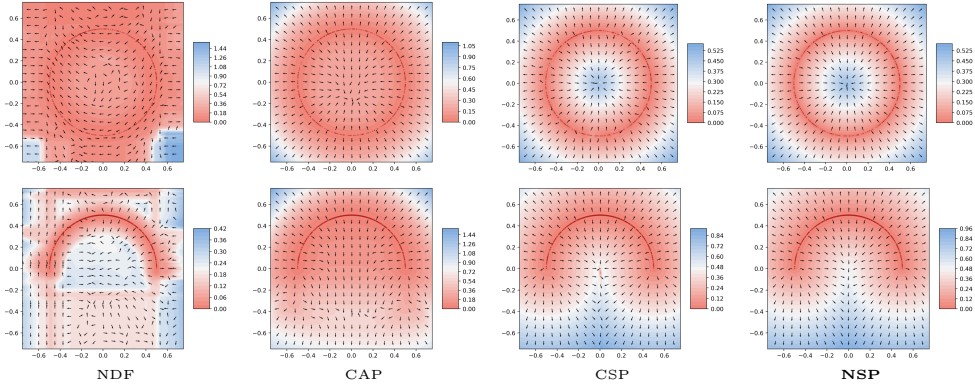
\begin{figure}
	\begin{tikzpicture}[every node/.style={font=\tiny}]		
		\node[anchor=south west,inner sep=0] (image1) at (0,0) 
		{\includegraphics[width=0.24\linewidth]{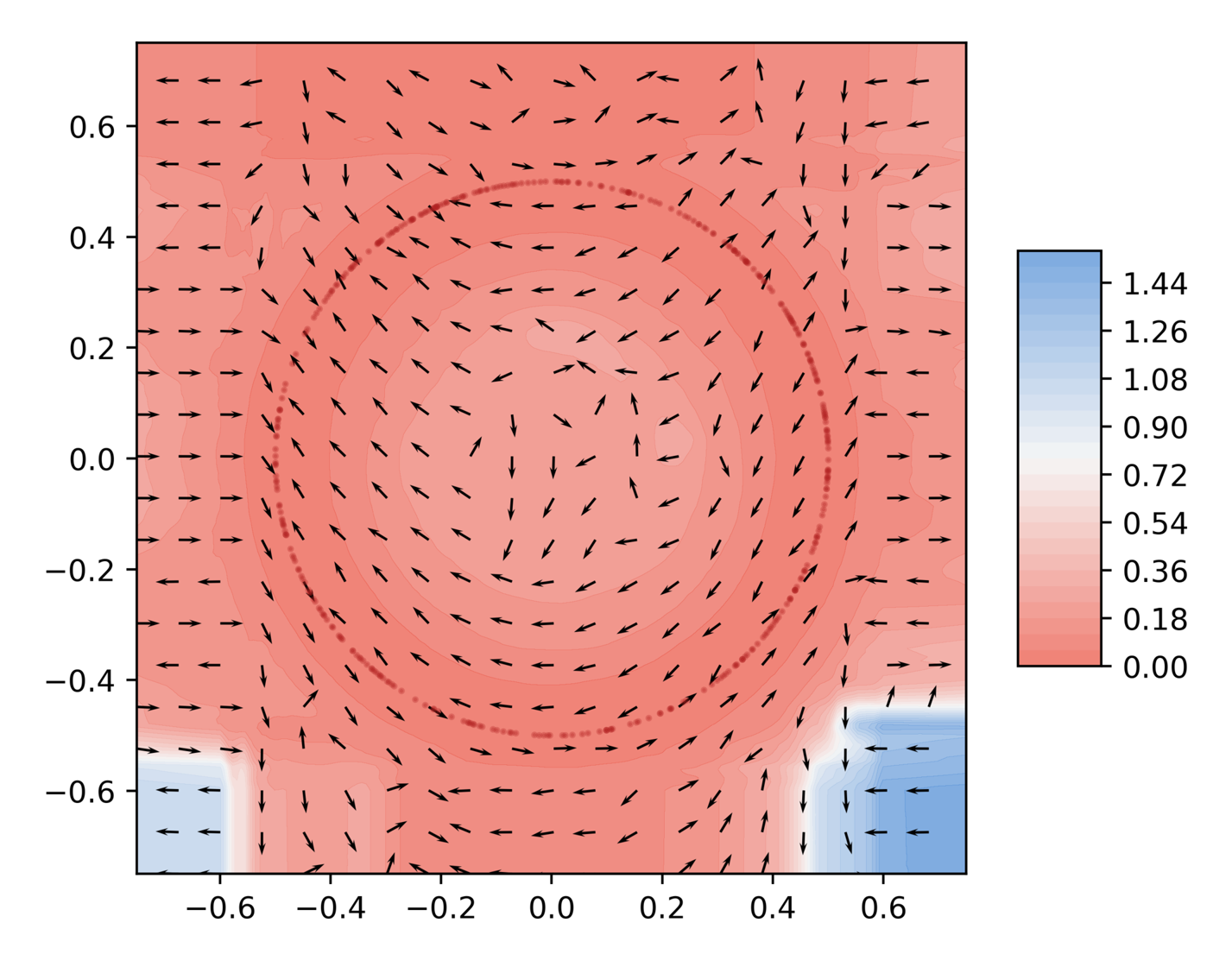}};
		\node[anchor=south west,inner sep=0] (image2) at (0.25\linewidth,0) 
		{\includegraphics[width=0.24\linewidth]{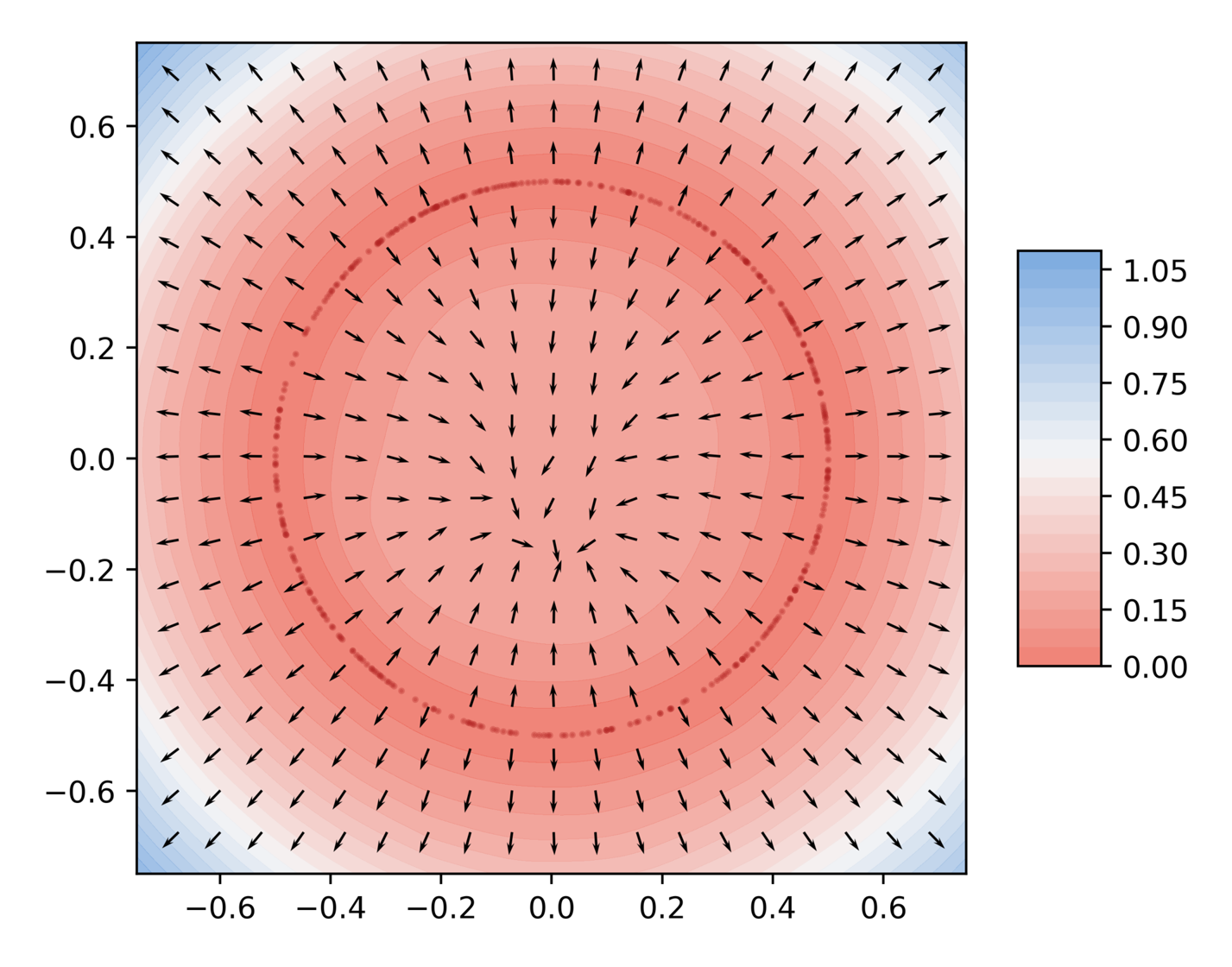}};
		\node[anchor=south west,inner sep=0] (image3) at (0.5\linewidth,0) 
		{\includegraphics[width=0.24\linewidth]{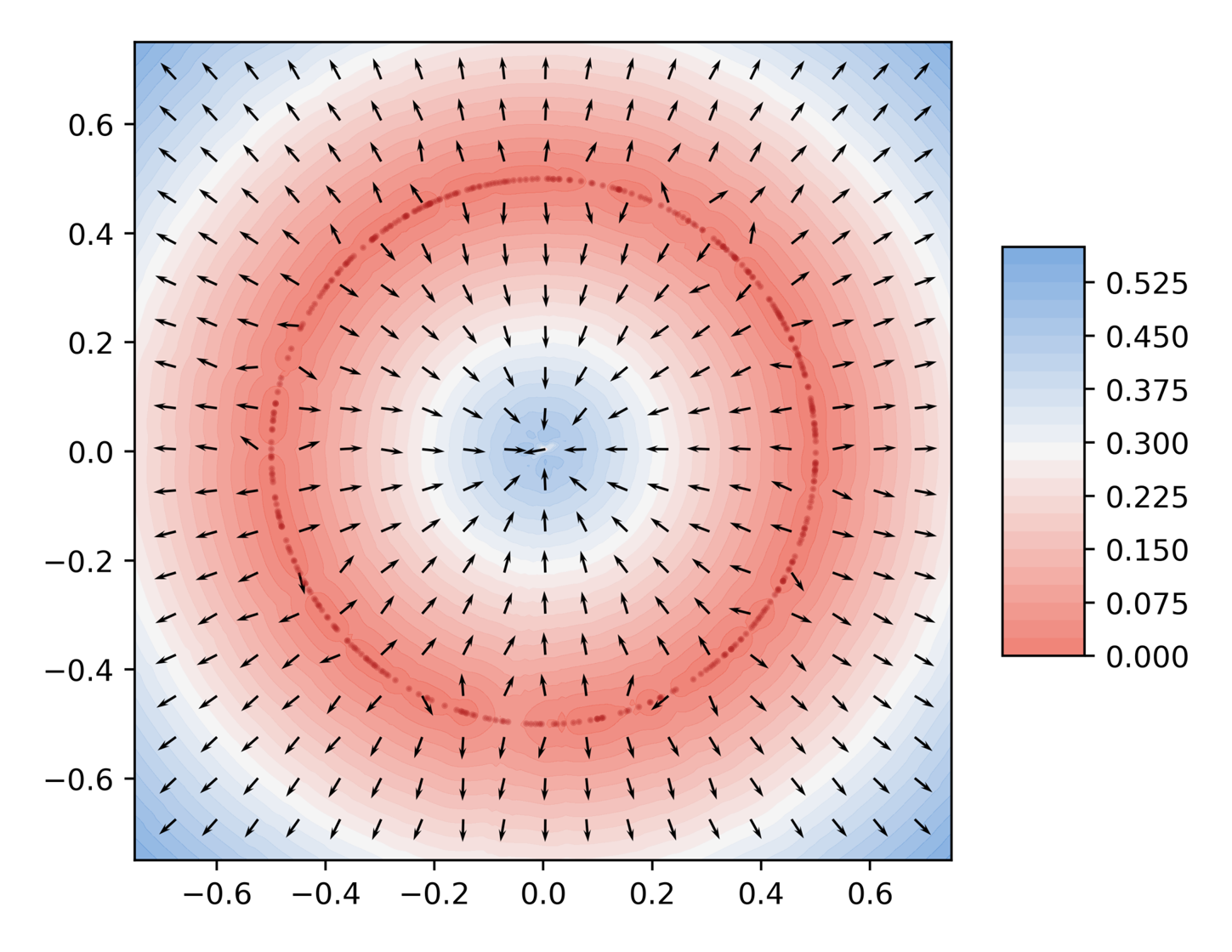}};
		\node[anchor=south west,inner sep=0] (image4) at (0.75\linewidth,0) 
		{\includegraphics[width=0.24\linewidth]{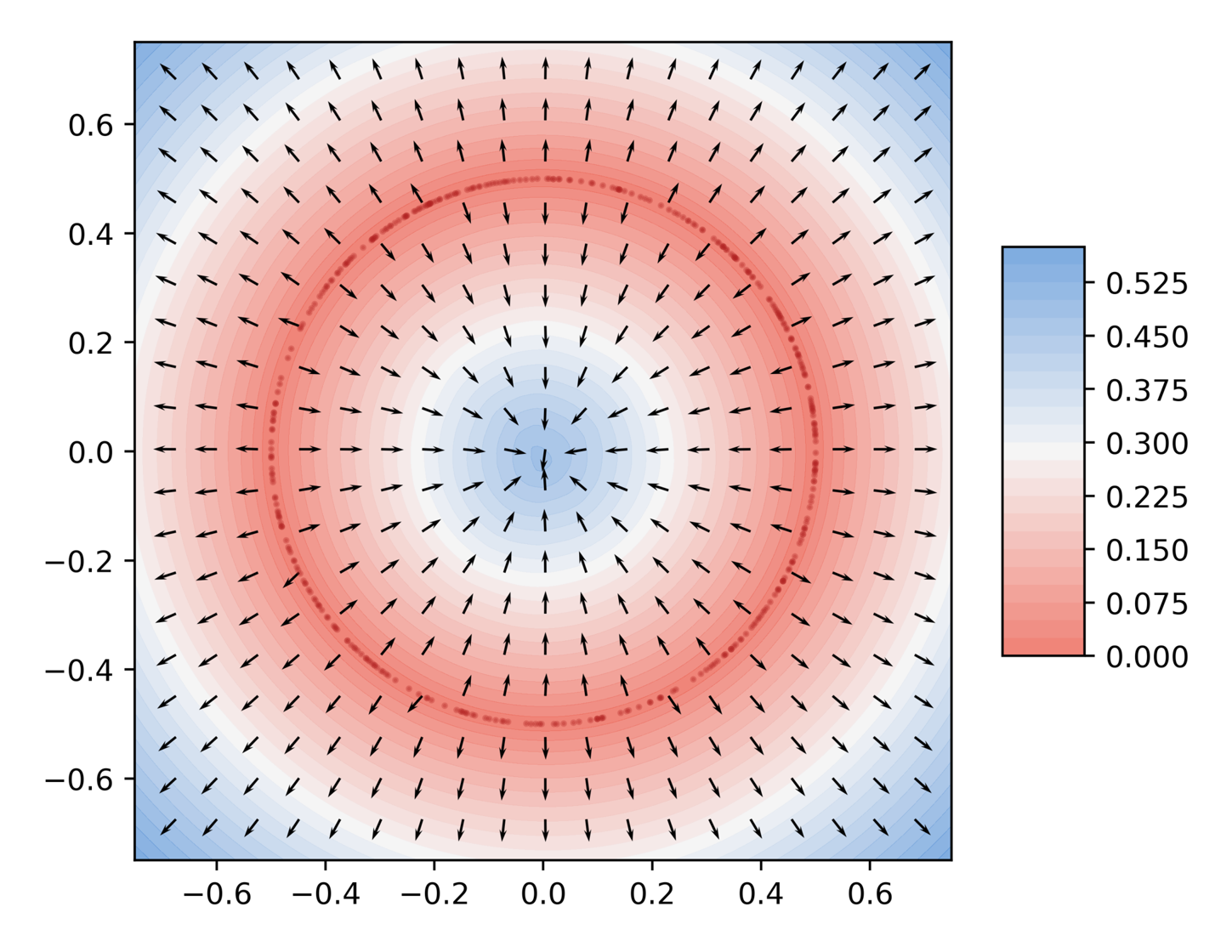}};		
		\node[anchor=north west,inner sep=0] (image5) at (image1.south west) 
		{\includegraphics[width=0.24\linewidth]{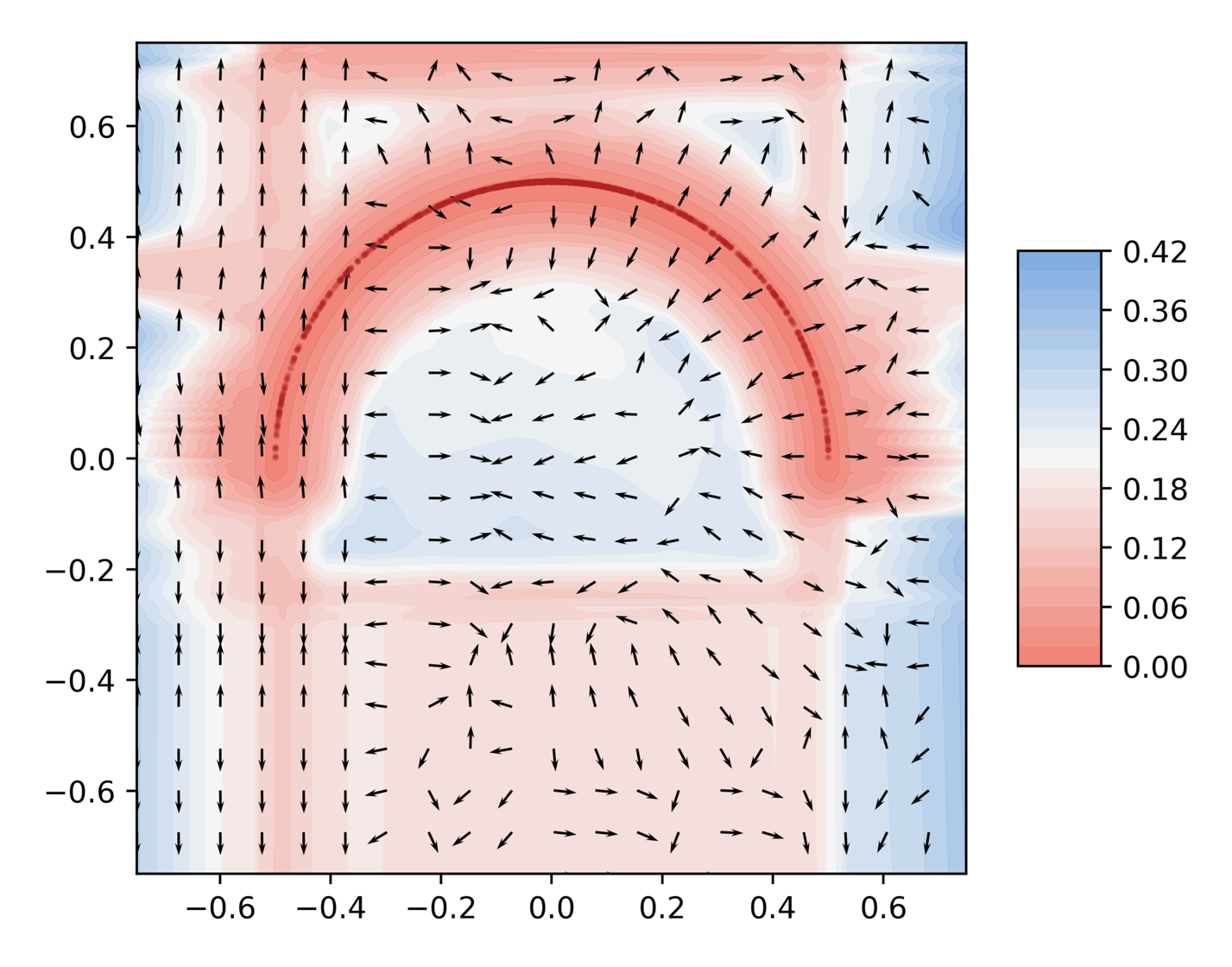}};
		\node[anchor=north west,inner sep=0] (image6) at (image2.south west) 
		{\includegraphics[width=0.24\linewidth]{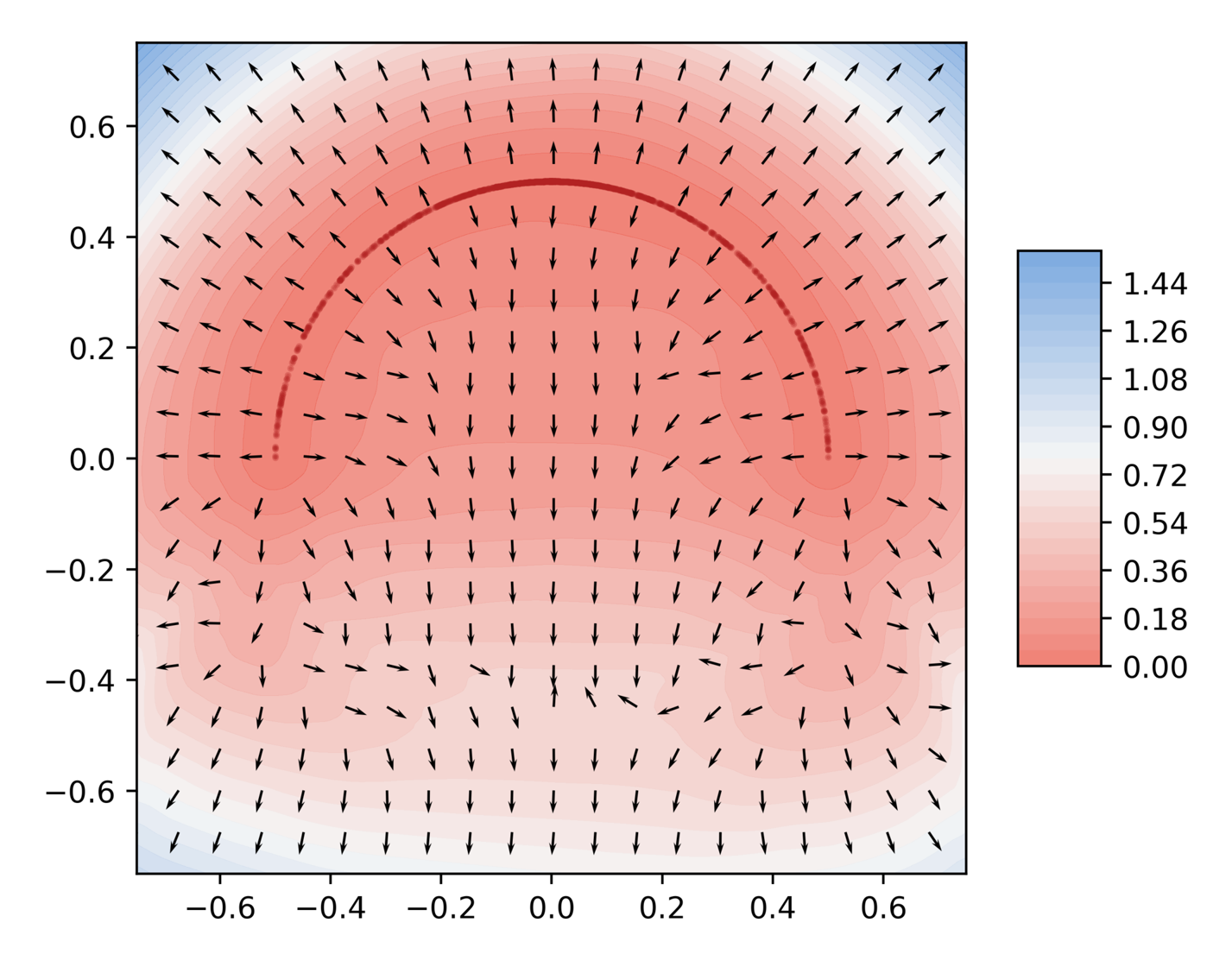}};
		\node[anchor=north west,inner sep=0] (image7) at (image3.south west) 
		{\includegraphics[width=0.24\linewidth]{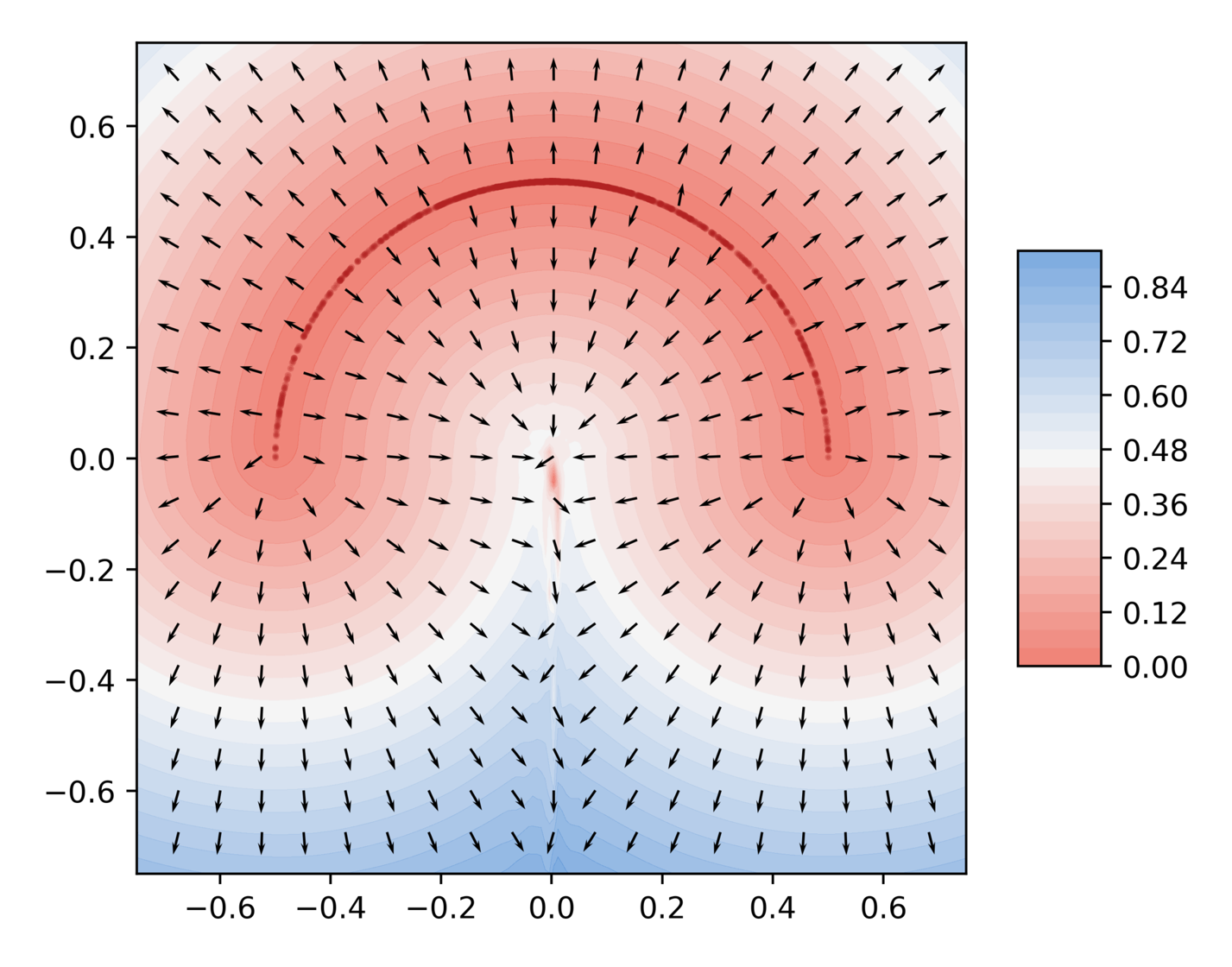}};
		\node[anchor=north west,inner sep=0] (image8) at (image4.south west) 
		{\includegraphics[width=0.24\linewidth]{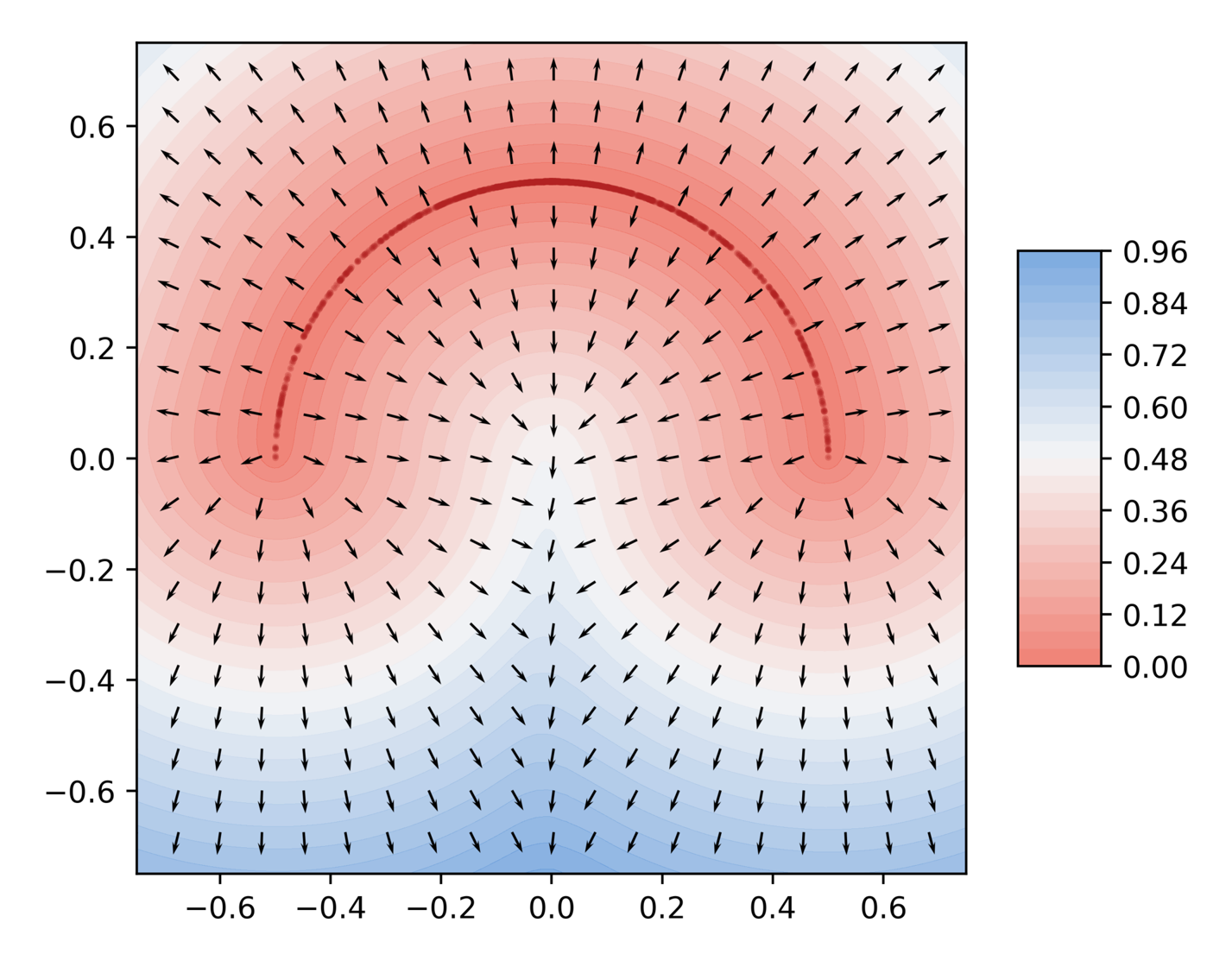}};		
		\node[anchor=north,inner sep=1pt] at (image5.south) {NDF};
		\node[anchor=north,inner sep=1pt] at (image6.south) {CAP};
		\node[anchor=north,inner sep=1pt] at (image7.south) {CSP};
		\node[anchor=north,inner sep=1pt] at (image8.south) {\textbf{NSP}};
	\end{tikzpicture}
    \caption{Results from the trained models on hemisphere data, showing cross section cuts in planes $z=0.1$ (top row) and $y=0$ (bottom row). The level sets illustrate the distance function learned from a given point cloud, represented by red dots. Quivers indicate the gradient field of the trained distance functions.}
    \label{fig:toy_hemisphere}
\end{figure}

\begin{figure}
	\begin{tikzpicture}[every node/.style={font=\tiny}]		
		\node[anchor=south west,inner sep=0] (image1) at (0,0) 
		{\includegraphics[width=0.24\linewidth]{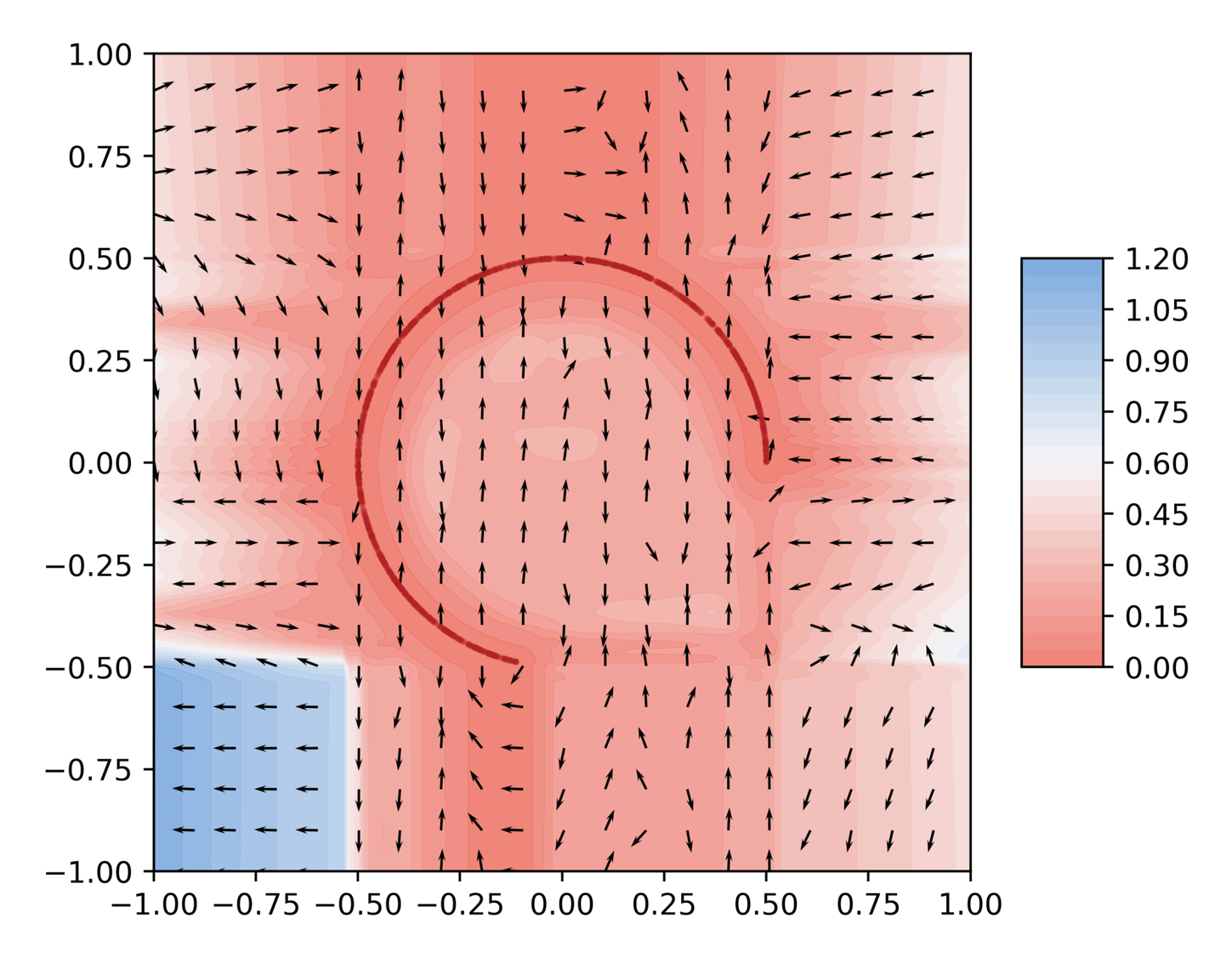}};
		\node[anchor=south west,inner sep=0] (image2) at (0.25\linewidth,0) 
		{\includegraphics[width=0.24\linewidth]{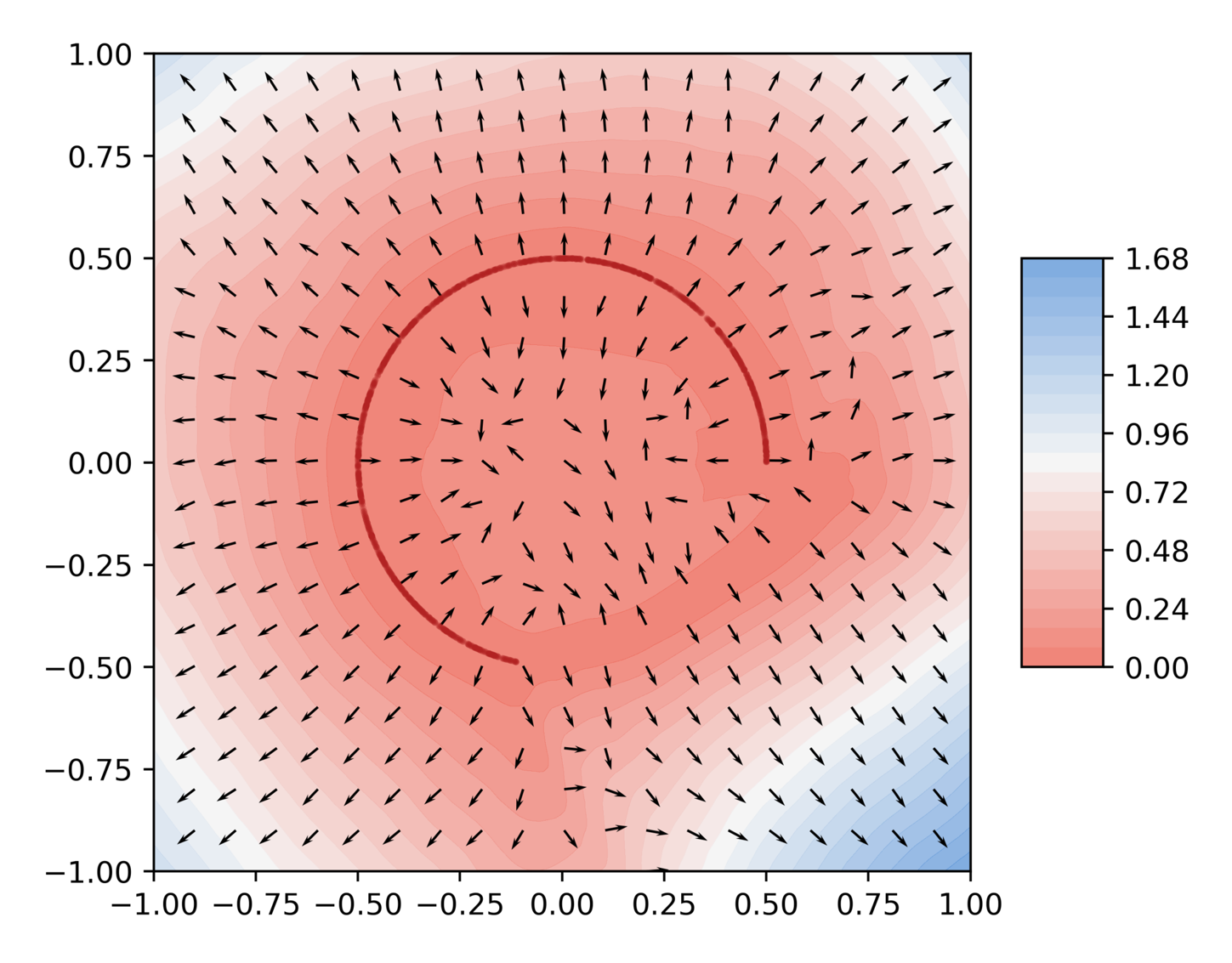}};
		\node[anchor=south west,inner sep=0] (image3) at (0.5\linewidth,0) 
		{\includegraphics[width=0.24\linewidth]{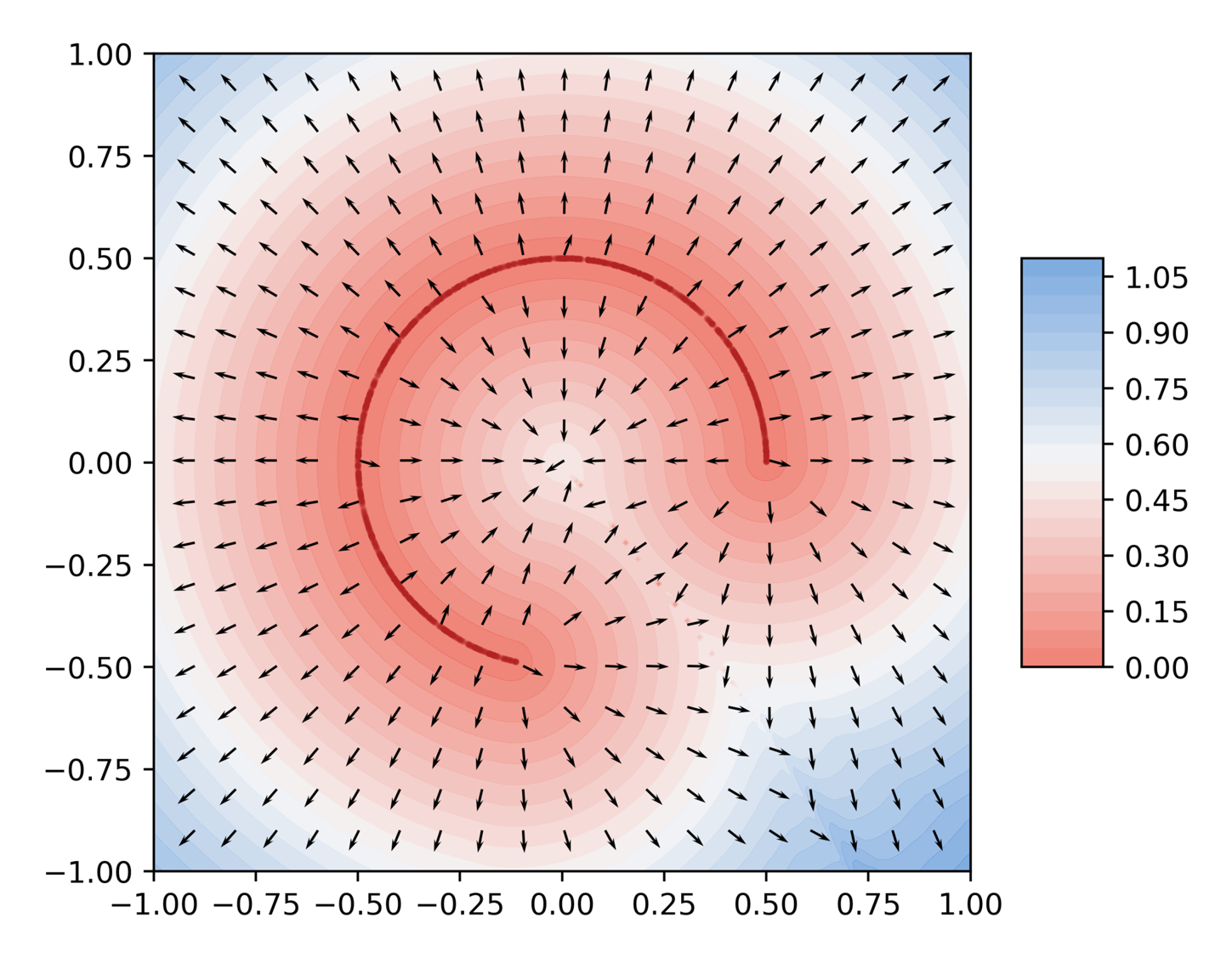}};
		\node[anchor=south west,inner sep=0] (image4) at (0.75\linewidth,0) 
		{\includegraphics[width=0.24\linewidth]{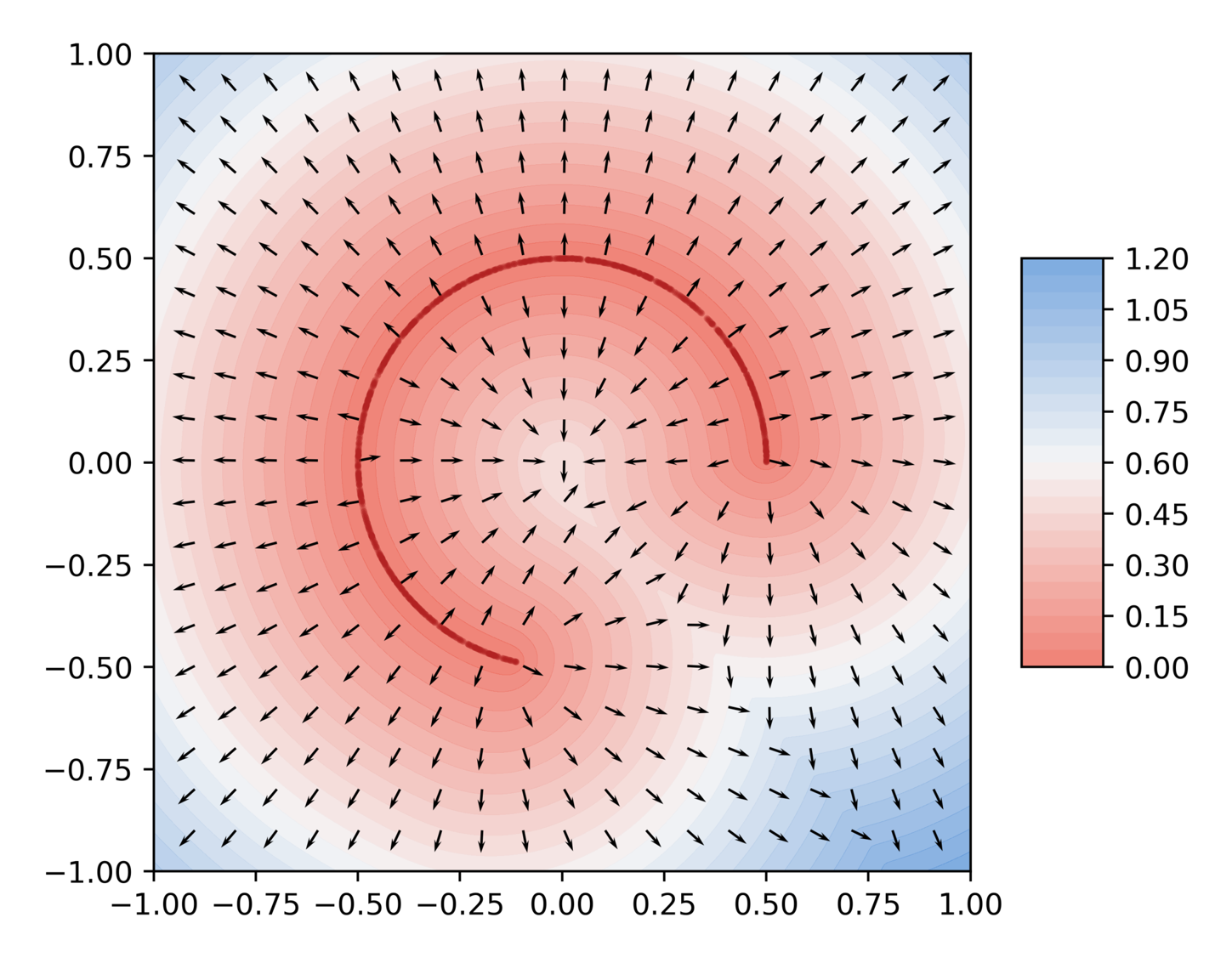}};		
		\node[anchor=north west,inner sep=0] (image5) at (image1.south west) 
		{\includegraphics[width=0.24\linewidth]{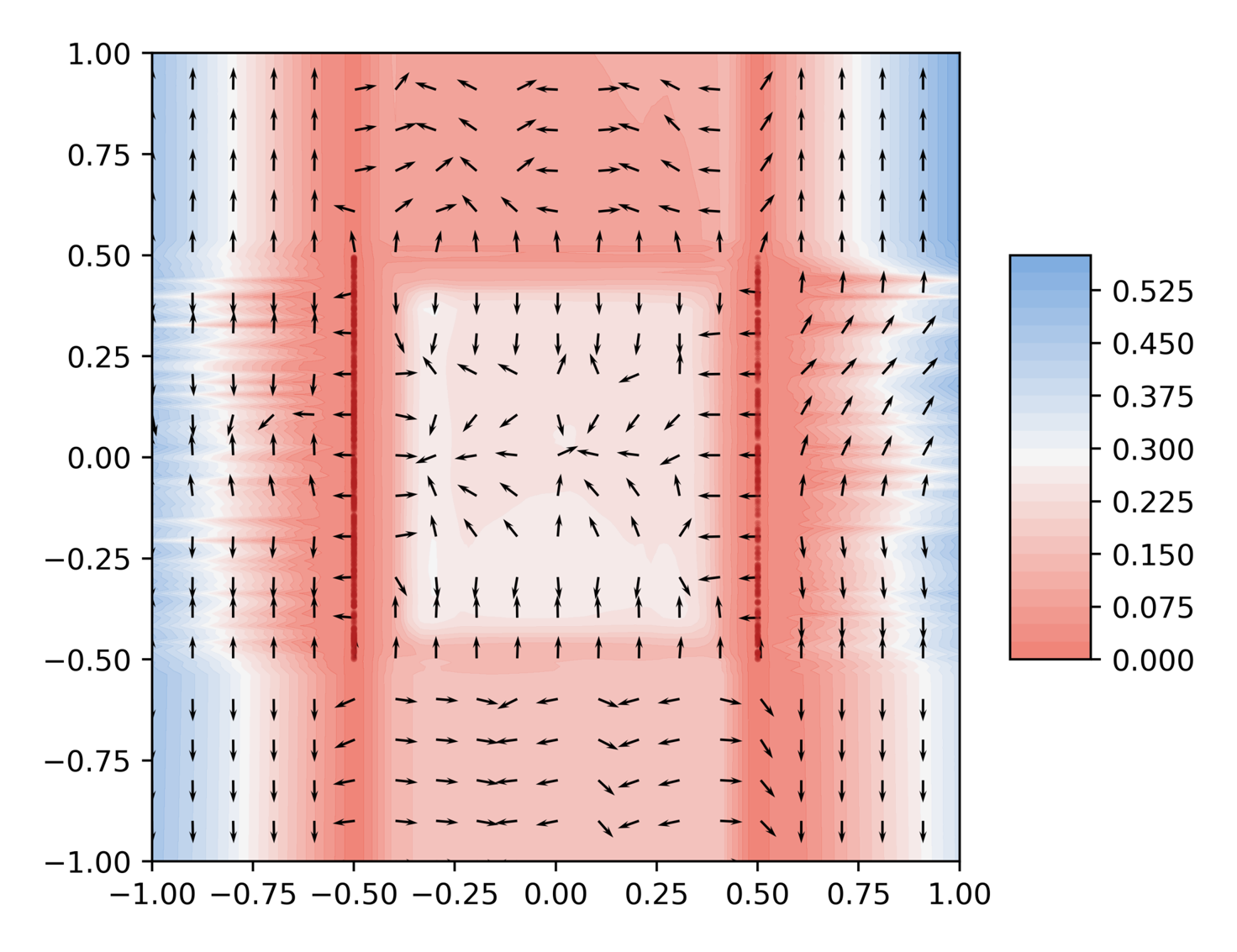}};
		\node[anchor=north west,inner sep=0] (image6) at (image2.south west) 
		{\includegraphics[width=0.24\linewidth]{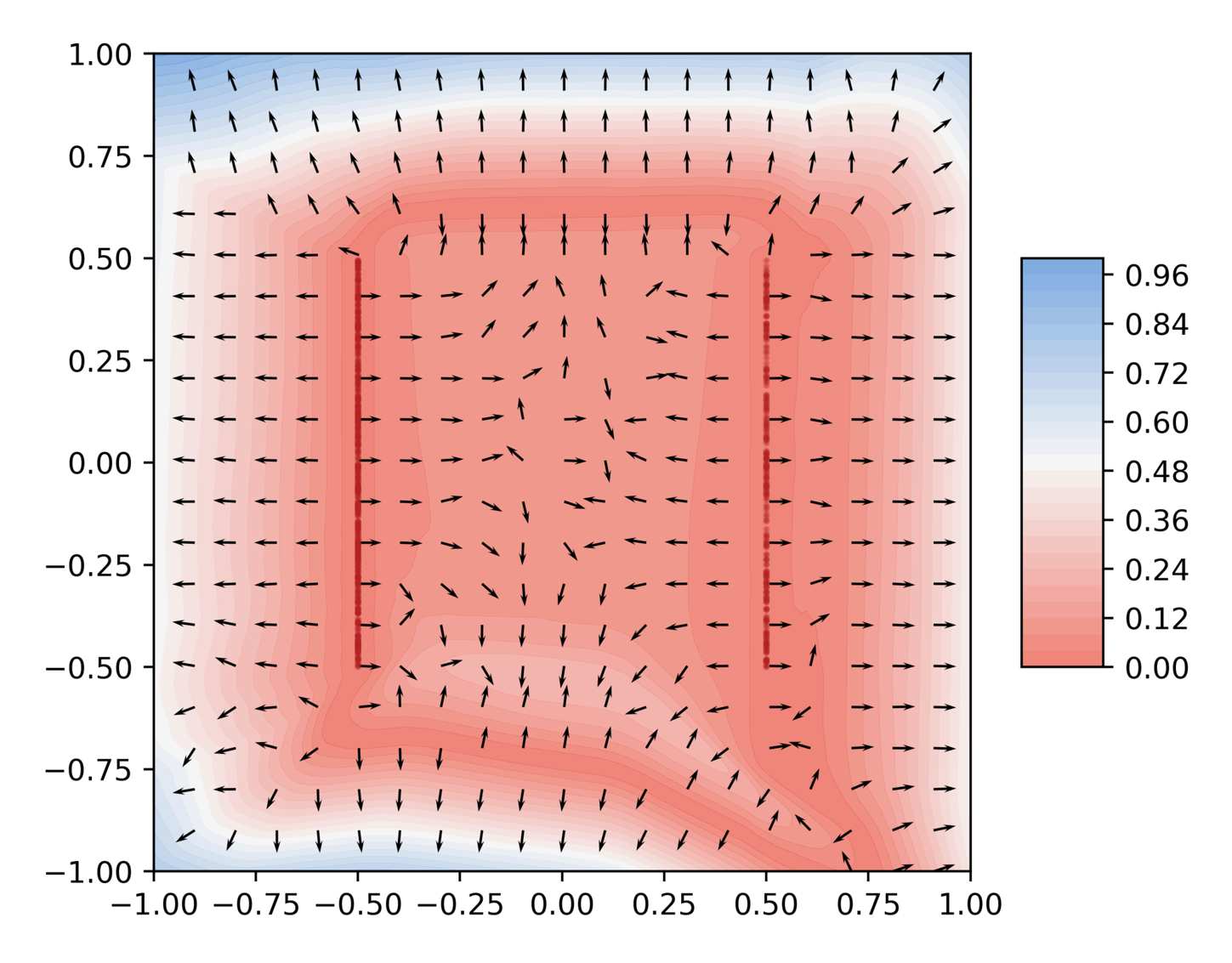}};
		\node[anchor=north west,inner sep=0] (image7) at (image3.south west) 
		{\includegraphics[width=0.24\linewidth]{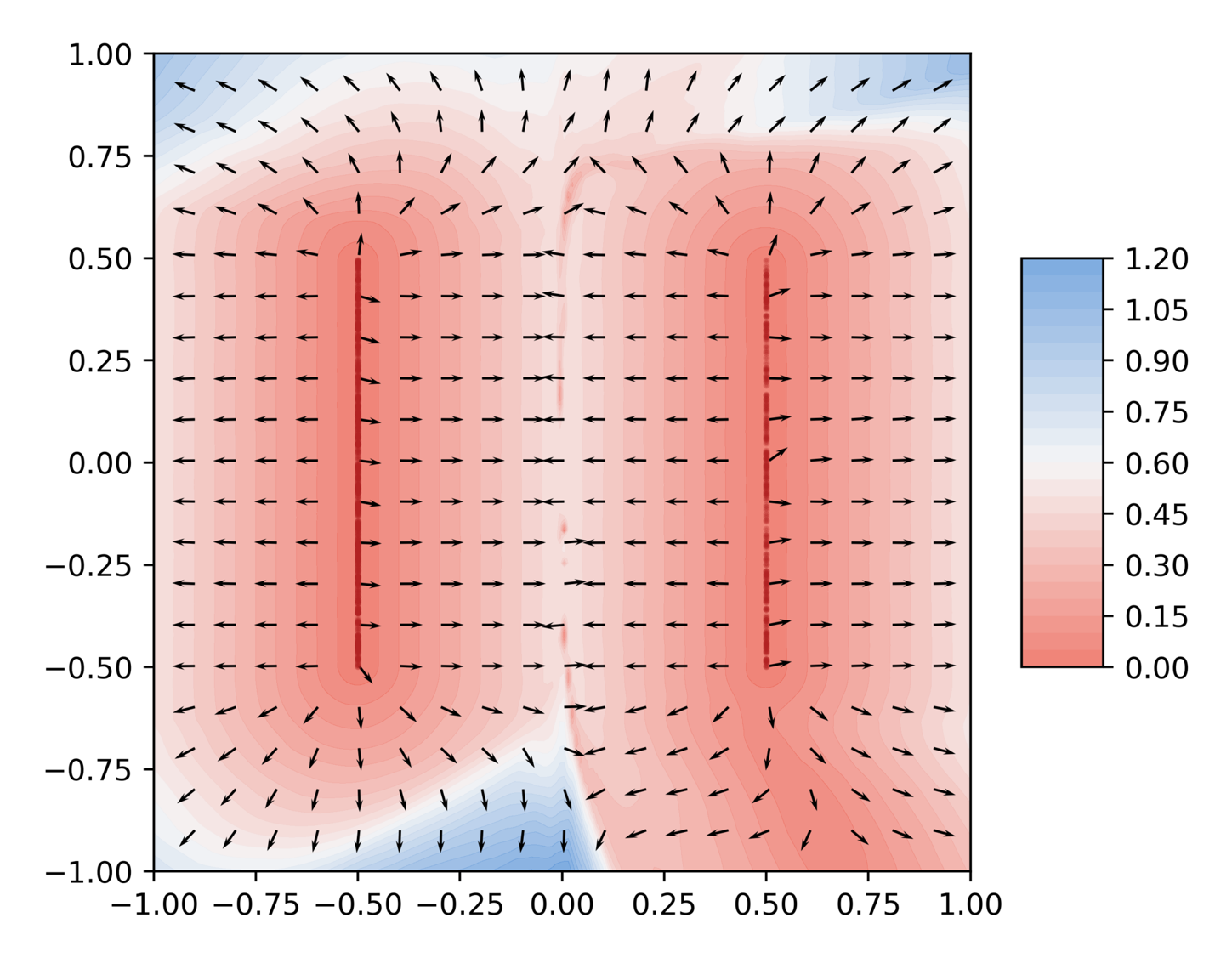}};
		\node[anchor=north west,inner sep=0] (image8) at (image4.south west) 
		{\includegraphics[width=0.24\linewidth]{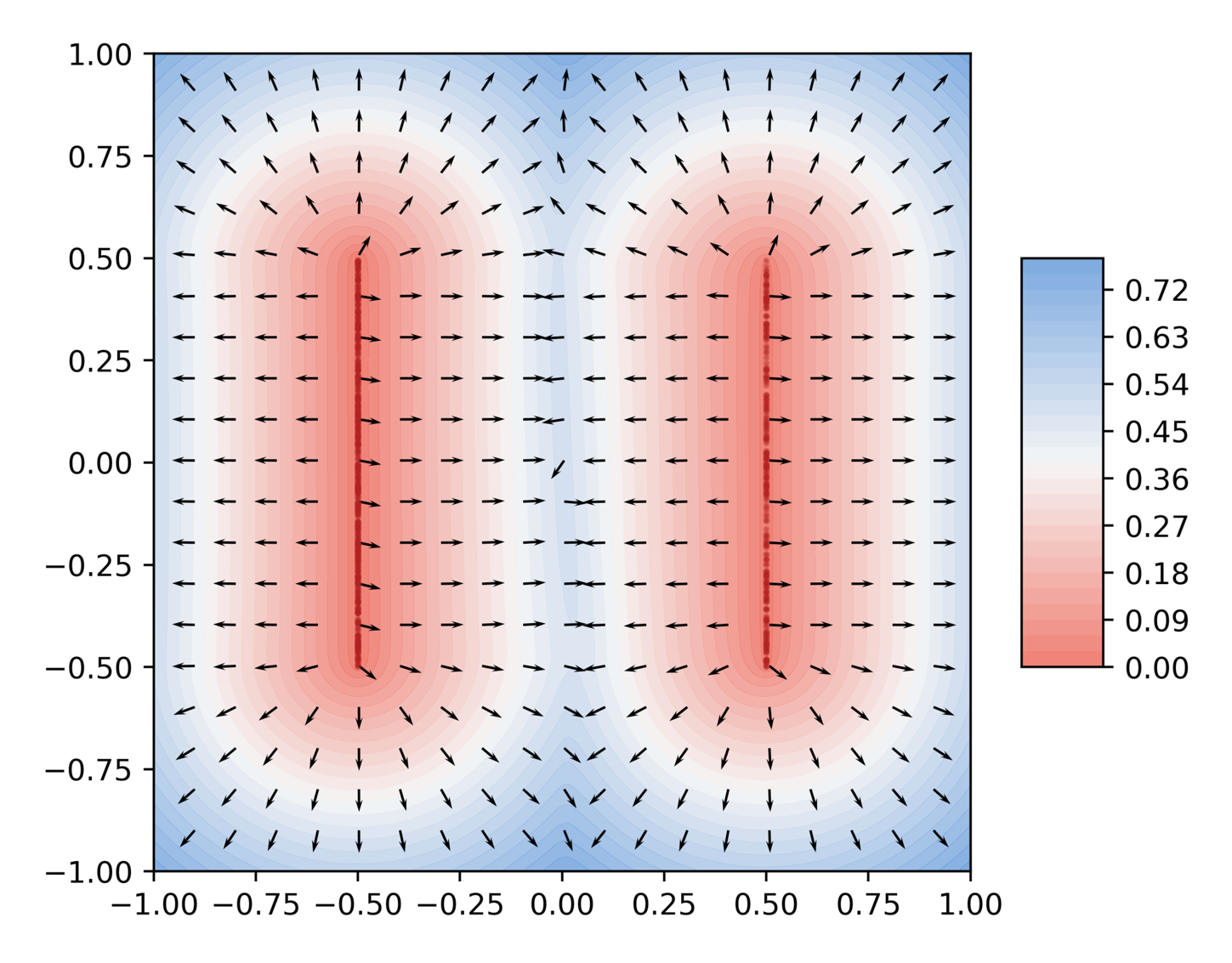}};		
		\node[anchor=north,inner sep=1pt] at (image5.south) {NDF};
		\node[anchor=north,inner sep=1pt] at (image6.south) {CAP};
		\node[anchor=north,inner sep=1pt] at (image7.south) {CSP};
		\node[anchor=north,inner sep=1pt] at (image8.south) {\textbf{NSP}};
	\end{tikzpicture}
    \caption{Results from the trained models on partial cylinder data, showing cross section cuts in planes $z=0$ (top row) and $y=0$ (bottom row). The level sets illustrate the distance functions learned from a given point cloud, represented by red dots. Quivers indicate the gradient field of the trained distance functions.}
    \label{fig:toy_cylinder}
\end{figure}
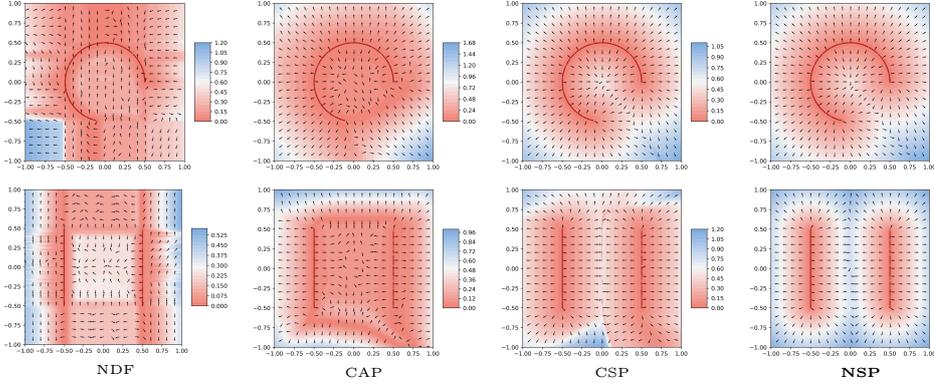

For comparative analysis, experiments are conducted for recent high-performing INR models for learning distance functions, namely NDF \citep{chibane2020neural}, CAP \citep{zhou2024cap}, and CSP \citep{venkatesh2021deep}. To visually evaluate how well each model learns the distance function and its gradient, we present the iso-contours of the cross-sections at the $xy$ and $zx$ planes in Figures \ref{fig:toy_hemisphere} and \ref{fig:toy_cylinder}, accompanied by quiver plots representing the learned gradients.
The results clearly demonstrate that even on such simple toy data, the proposed NSP approximates both the distance function and its gradient with considerably greater accuracy than existing approaches.
The baseline models -- NDF, CAP, and CSP -- exhibit distorted gradients. Additionally, the iso-contours appear irregular and imprecise as a consequence of inaccurate gradient approximations. Notably, NDF and CAP managed to learn the given point cloud to include it within the zero level set, but they failed to learn the overall distance function. CSP relatively learned the distance function, yet there was a disruption observed in the gradients. In particular, the iso-contours of CSP along the $y=0$ plane of the partial cylinder in Figure \ref{fig:toy_cylinder} appeared to be considerably distorted. In contrast, the NSP demonstrates precise learning of both the distance function and its gradient. These results show that NSP, which learns the ESP instead of the distance function and proves $H^1$ convergence of $d_\theta$, actually learns both the distance function and its gradient with high accuracy. This highlights the advantages of NSP, which, by training ESP, simultaneously learns both the distance function and its gradient, in contrast to baselines that directly learn the distance function using neural networks. 

\subsection{Surface reconstruction for synthetic shapes}\label{sec:exp_synthetic}
In our synthetic shape experiments, we examine the MGN dataset \citep{bhatnagar2019multi}, which consists of garment shapes, as suggested by \citep{zhou2024cap}. This dataset is suitable for demonstrating the performance of the proposed method in the context of open surfaces. Additionally, we adopt the approach outlined by \citep{chibane2020neural}, employing the car category from the ShapeNet dataset \citep{shapenet2015}. This dataset enables us to evaluate the capability of the proposed model to reconstruct surfaces with challenging properties such as multi-layered and non-closed shapes.

\begin{table}
    \centering
       \caption{Results of surface reconstruction on MGN dataset.} \label{tab:mgn}
    \begin{tabular}{lcccccccc}
   	\toprule 
   	& \multicolumn{4}{c}{Mesh} & \multicolumn{4}{c}{Points}\\
   	\cmidrule(lr){2-5} \cmidrule(lr){6-9}
   	& \multicolumn{2}{c}{$d_C$} & \multicolumn{2}{c}{$d_H$} & \multicolumn{2}{c}{$d_C$} & \multicolumn{2}{c}{$d_H$}\\
   	\cmidrule(lr){2-3} \cmidrule(lr){4-5} \cmidrule(lr){6-7} \cmidrule(lr){8-9}
   	Model & Mean & Median & Mean & Median & Mean & Median & Mean & Median\\
   	\cmidrule(lr){1-1} \cmidrule(lr){2-3} \cmidrule(lr){4-5} \cmidrule(lr){6-7} \cmidrule(lr){8-9}
     NDF    & 0.1138 & 0.1138 & 0.5494 & 0.5494 & 0.1125 & 0.1125 &  0.5493 & 0.5493\\
     CAP    &  0.0027 & 0.0013 & 0.0680 & 0.7190 & 0.0048 & 0.0036 & 0.0694 & 0.0722\\
     CSP    &  0.0524 & 0.0520 & 0.4134 & 0.4329 & 0.0524 & 0.0517 & 0.4139 & 0.4332\\
     \textbf{NSP} & 0.0012 & 0.0010 & 0.0242 & 0.2160 & 0.0033 & 0.0033 & 0.0257 & 0.0217\\
     \bottomrule
    \end{tabular}
\end{table}

\begin{table}
    \centering
    \caption{Results of surface reconstruction on ShapeNet cars dataset.} \label{tab:cars}
    \begin{tabular}{lcccccccc}
    \toprule 
    & \multicolumn{4}{c}{Mesh} & \multicolumn{4}{c}{Points}\\
    \cmidrule(lr){2-5} \cmidrule(lr){6-9}
    & \multicolumn{2}{c}{$d_C$} & \multicolumn{2}{c}{$d_H$} & \multicolumn{2}{c}{$d_C$} & \multicolumn{2}{c}{$d_H$}\\
    \cmidrule(lr){2-3} \cmidrule(lr){4-5} \cmidrule(lr){6-7} \cmidrule(lr){8-9}
    Model & Mean & Median & Mean & Median & Mean & Median & Mean & Median\\
    \cmidrule(lr){1-1} \cmidrule(lr){2-3} \cmidrule(lr){4-5} \cmidrule(lr){6-7} \cmidrule(lr){8-9}
     NDF    & 0.1042 & 0.1042 & 0.3666 & 0.3667 & 0.1073 & 0.1073 & 0.3445 & 0.3445\\
     CAP    & 0.0151 & 0.0151 & 0.0633 & 0.0633 & 0.0034 & 0.0035 & 0.0161 & 0.0161\\
     CSP    & 0.0131 & 0.0131 & 0.1069 & 0.1068 & 0.0067 & 0.0067 & 0.0671 & 0.0670\\
     \textbf{NSP} & 0.0148 & 0.0148 & 0.0619 & 0.0620 & 0.0040 & 0.0040 & 0.0268 & 0.0269\\
     \bottomrule
    \end{tabular}
\end{table}

We compare the performance of the proposed NSP to recent leading INR methods, including NDF, CAP, and CSP. 
To evaluate the quality of the reconstruction, we utilize two metrics: Chamfer distance $d_C$ and Hausdorff distance $d_H$ and sample one million points on the reconstruction to compute distances. These are used to quantify the discrepancy between the reconstructions and the ground truth (GT) meshes, as well as to evaluate the distance between the input point clouds and the reconstructions.
	
\begin{figure}
\centering
\begin{tikzpicture}[every node/.style={font=\tiny}]
	\node[anchor=south west, inner sep=0] (input1) 
	at (0,0) {\includegraphics[width=0.32\linewidth]{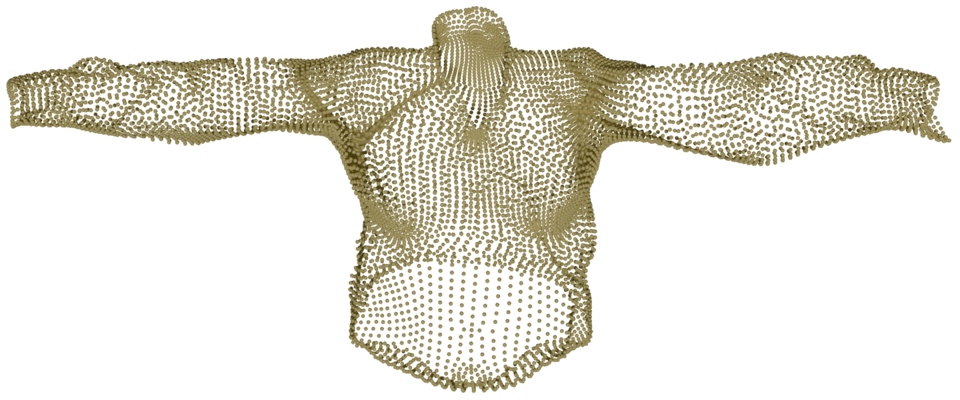}};
	\node[anchor=north east,inner sep=0, xshift=-5pt, yshift=25pt] at (input1.south east) {Input};
	\node[anchor=south west, inner sep=0] (gt1) 
	at (0.34\linewidth,0) {\includegraphics[width=0.32\linewidth]{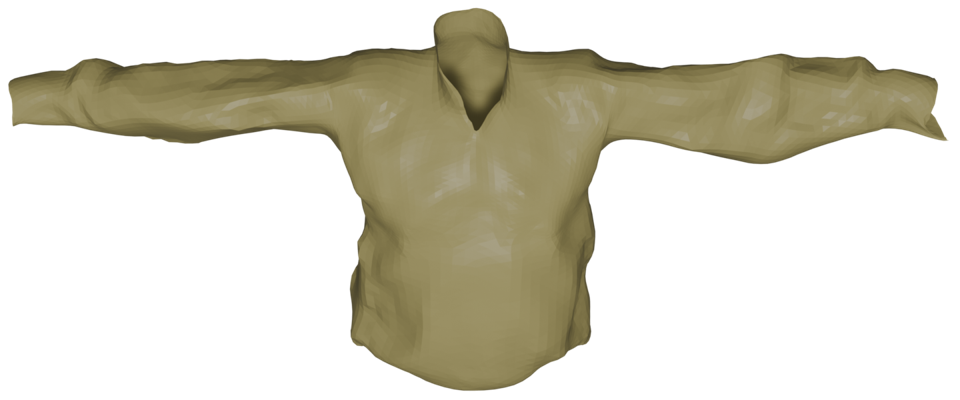}};
	\node[anchor=north east,inner sep=0, xshift=-5pt, yshift=25pt] at (gt1.south east) {GT};
	\node[anchor=south west, inner sep=0] (ndf1) 
	at (0.68\linewidth,0) {\includegraphics[width=0.32\linewidth]{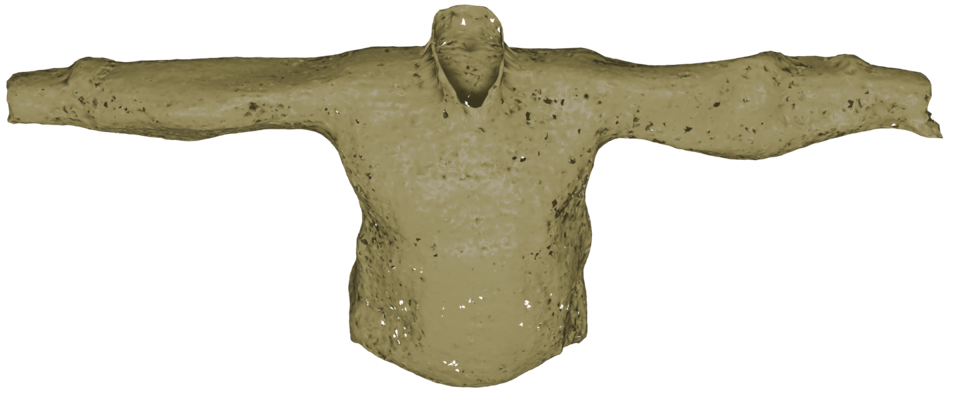}};
	\node[anchor=north east,inner sep=0, xshift=-5pt, yshift=25pt] at (ndf1.south east) {NDF};
	
	\node[anchor=north west, inner sep=0] (cap1) 
	at (input1.south west) {\includegraphics[width=0.32\linewidth]{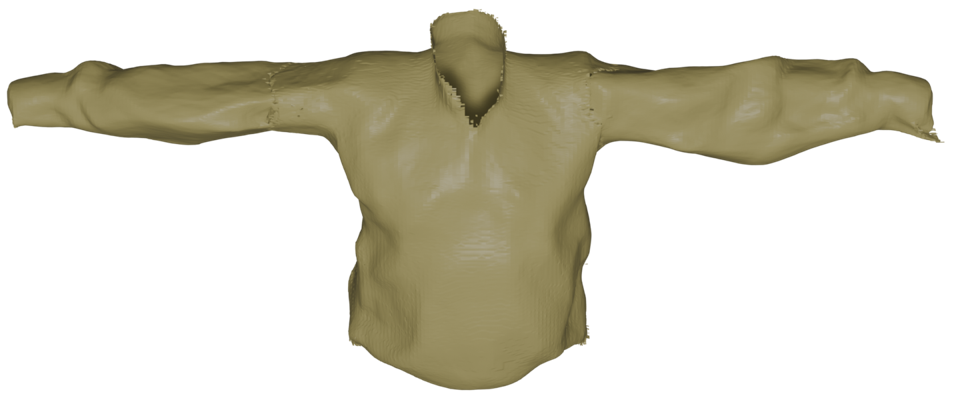}};
	\node[anchor=north east,inner sep=0, xshift=-5pt, yshift=25pt] at (cap1.south east) {CAP};
	\node[anchor=north west, inner sep=0] (csp1) 
	at (gt1.south west) {\includegraphics[width=0.32\linewidth]{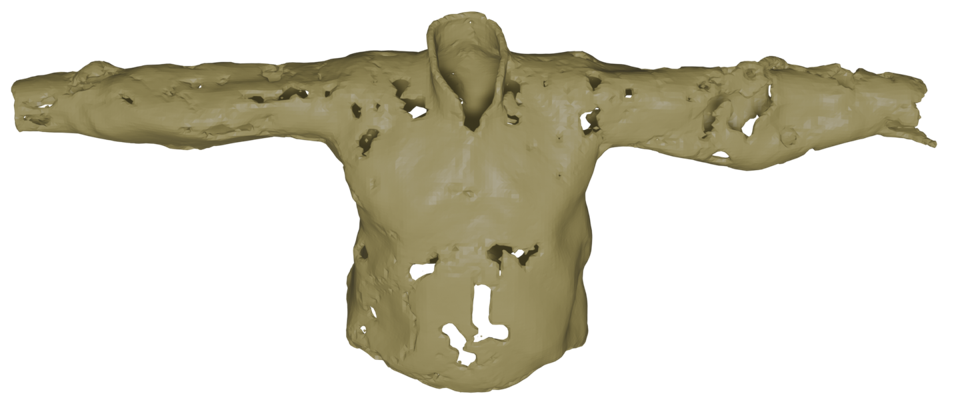}};
	\node[anchor=north east,inner sep=0, xshift=-5pt, yshift=25pt] at (csp1.south east) {CSP};
	\node[anchor=north west, inner sep=0] (nsp1) 
	at (ndf1.south west) {\includegraphics[width=0.32\linewidth]{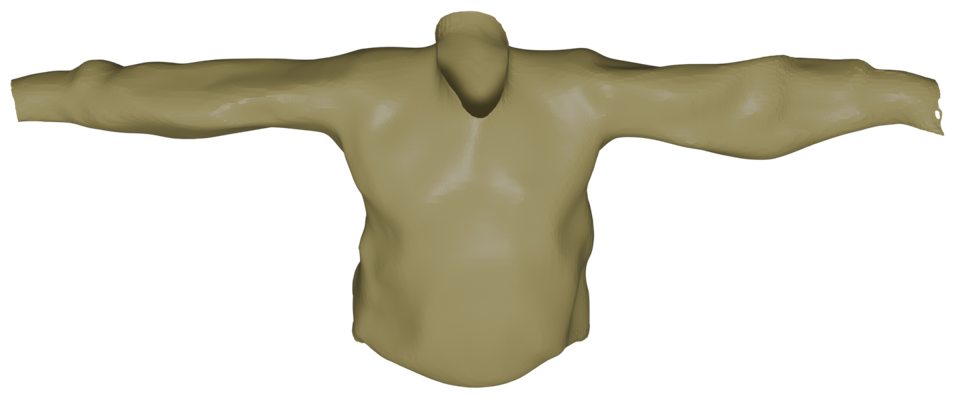}};
	\node[anchor=north east,inner sep=0, xshift=-5pt, yshift=25pt] at (nsp1.south east) {\textbf{NSP}};
	
	\node[anchor=north west, inner sep=0] (input2) 
	at (0,-1.8) {\includegraphics[width=0.15\linewidth]{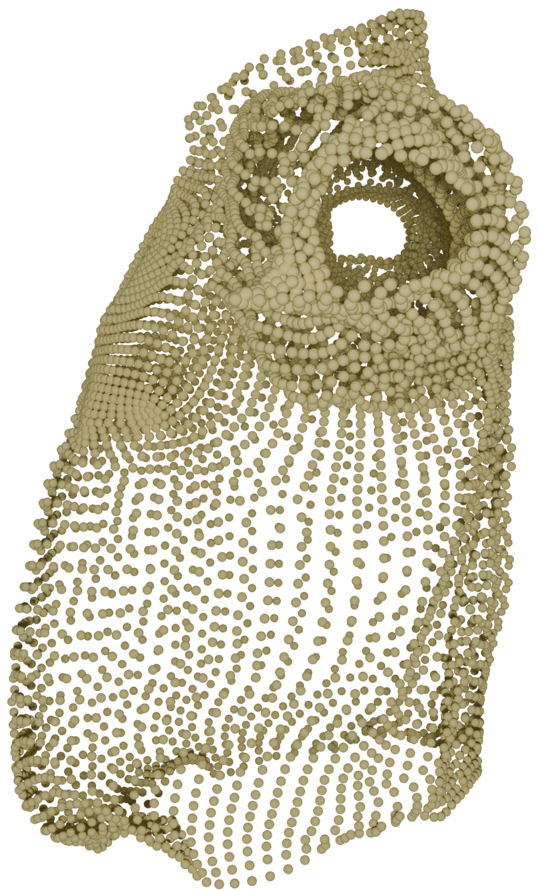}};
	\node[anchor=north east, inner sep=0, xshift=-5pt, yshift=3pt] at (input2.south east) {Input};
	\node[anchor=north west, inner sep=0] (gt2) 
	at (0.17\textwidth,-1.8) {\includegraphics[width=0.15\linewidth]{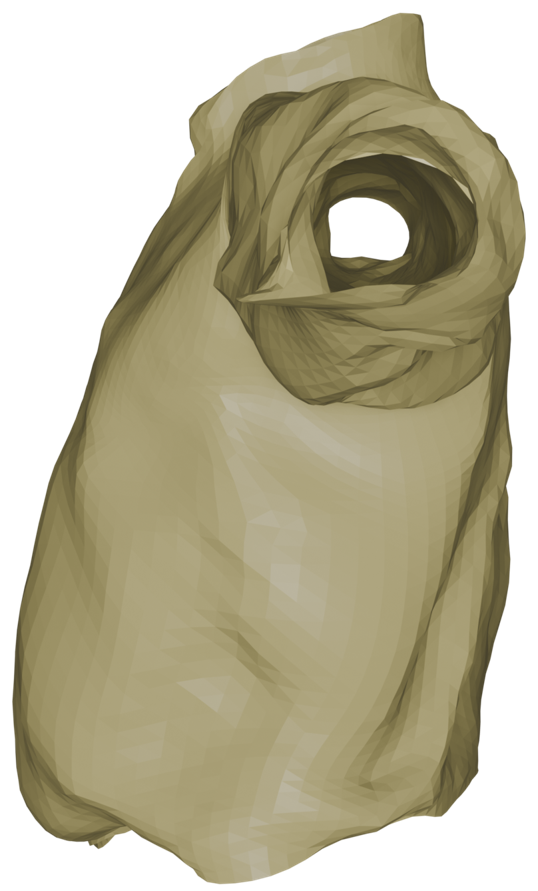}};
	\node[anchor=north east, inner sep=0, xshift=-5pt, yshift=3pt] at (gt2.south east) {GT};
	\node[anchor=north west, inner sep=0] (ndf2) 
	at (0.34\textwidth,-1.8) {\includegraphics[width=0.15\linewidth]{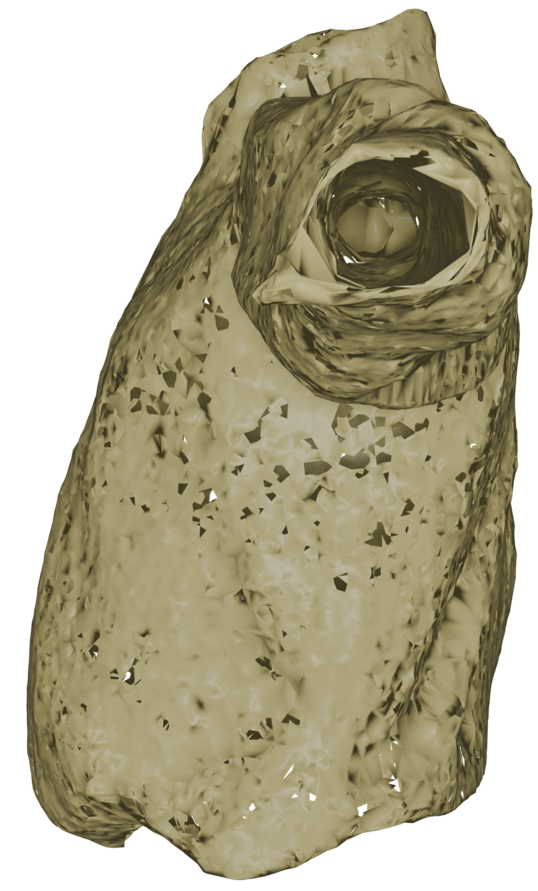}};
	\node[anchor=north east, inner sep=0, xshift=-5pt, yshift=3pt] at (ndf2.south east) {NDF};
	\node[anchor=north west, inner sep=0] (cap2) 
	at (0.51\textwidth,-1.8) {\includegraphics[width=0.15\linewidth]{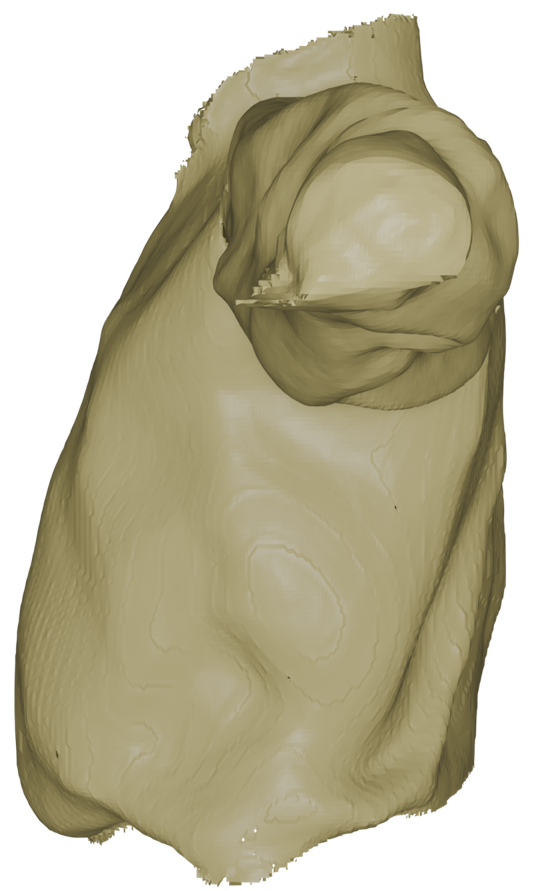}};
	\node[anchor=north east, inner sep=0, xshift=-5pt, yshift=3pt] at (cap2.south east) {CAP};
	\node[anchor=north west, inner sep=0] (csp2) 
	at (0.68\textwidth,-1.8) {\includegraphics[width=0.15\linewidth]{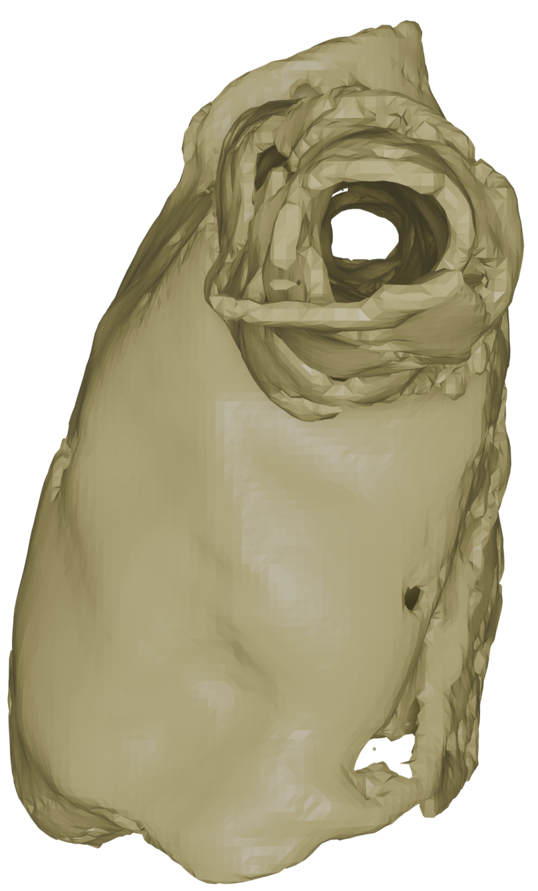}};
	\node[anchor=north east, inner sep=0, xshift=-5pt, yshift=3pt] at (csp2.south east) {CSP};
	\node[anchor=north west, inner sep=0] (nsp2) 
	at (0.85\textwidth,-1.8) {\includegraphics[width=0.15\linewidth]{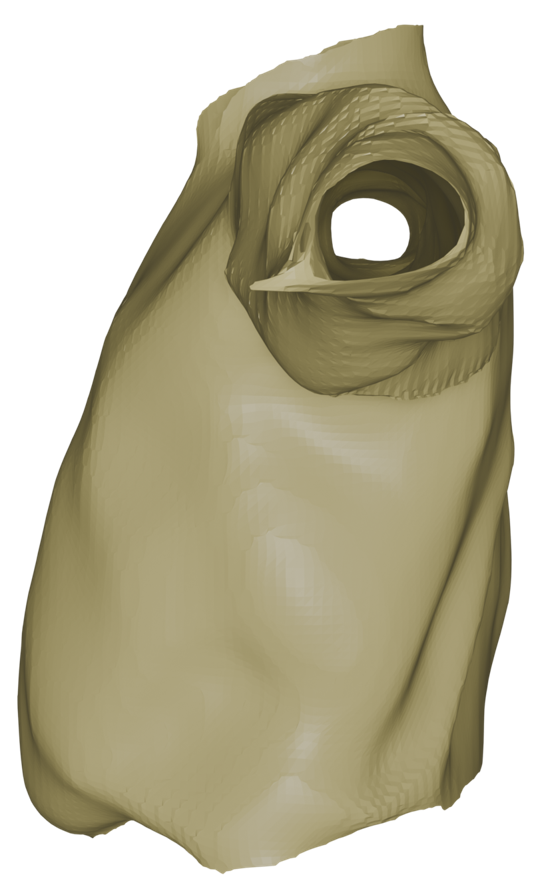}};
	\node[anchor=north east, inner sep=0, xshift=-5pt, yshift=3pt] at (nsp2.south east) {\textbf{NSP}};
	
	\node[rotate=90, anchor=north] at ($(input1.west)!0.5!(cap1.west) + (0,0.6)$) {Front};
	\node[rotate=90, anchor=north] at ($(input1.west)!0.5!(cap1.west) + (0,-1.2)$) {Front};
	\node[rotate=90, anchor=north] at ($(input2.west)!0.5!(input2.west) + (0,1)$) {Side};
\end{tikzpicture}
\caption{Reconstructed surfaces on the shirt data of MGN dataset.}
\label{fig:shirt}
\end{figure}
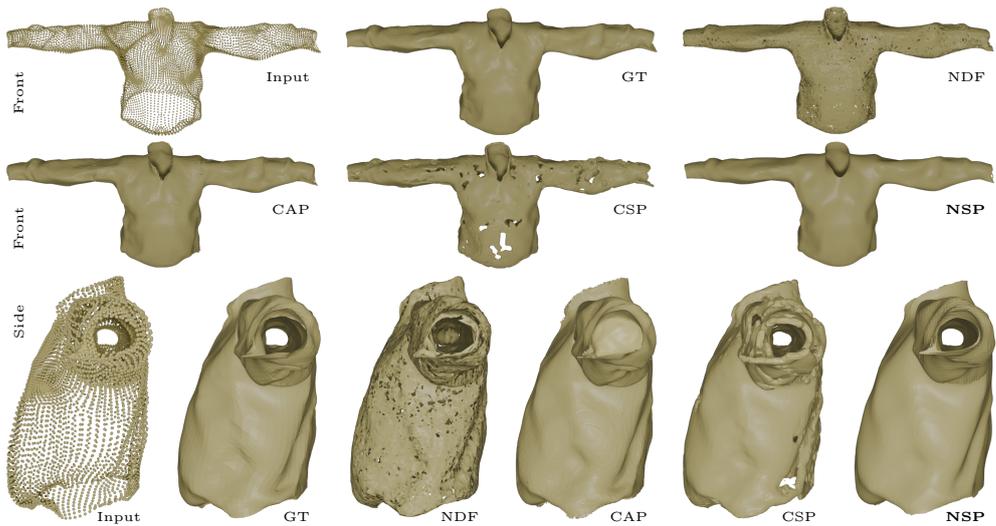

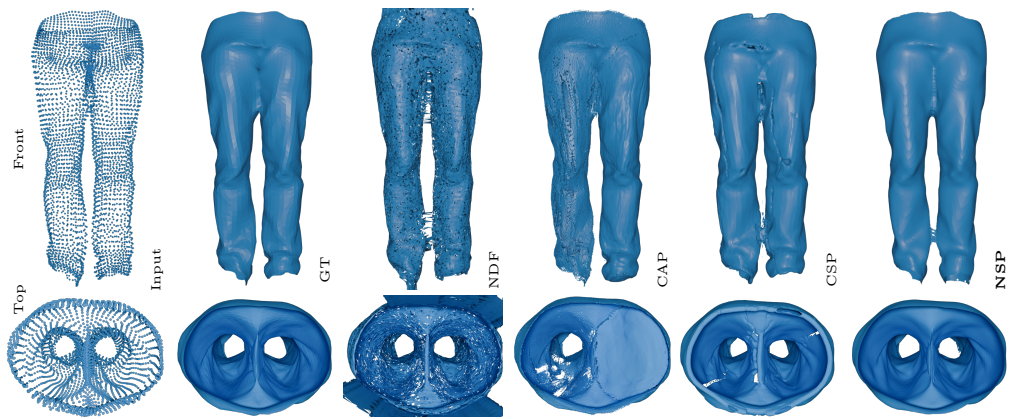
\begin{figure}
    \centering
    \begin{tikzpicture}[every node/.style={font=\tiny}]
		\node[anchor=north west, inner sep=0] (input1) 
		at (0,0) {\includegraphics[width=0.16\linewidth]{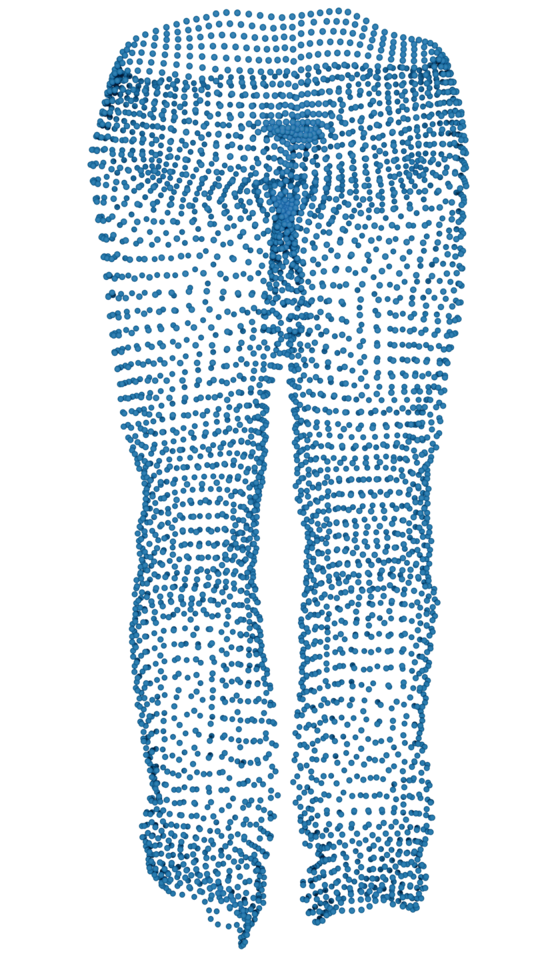}};
		\node[anchor=north west, inner sep=0] (gt1) 
		at (0.17\textwidth,0) {\includegraphics[width=0.16\linewidth]{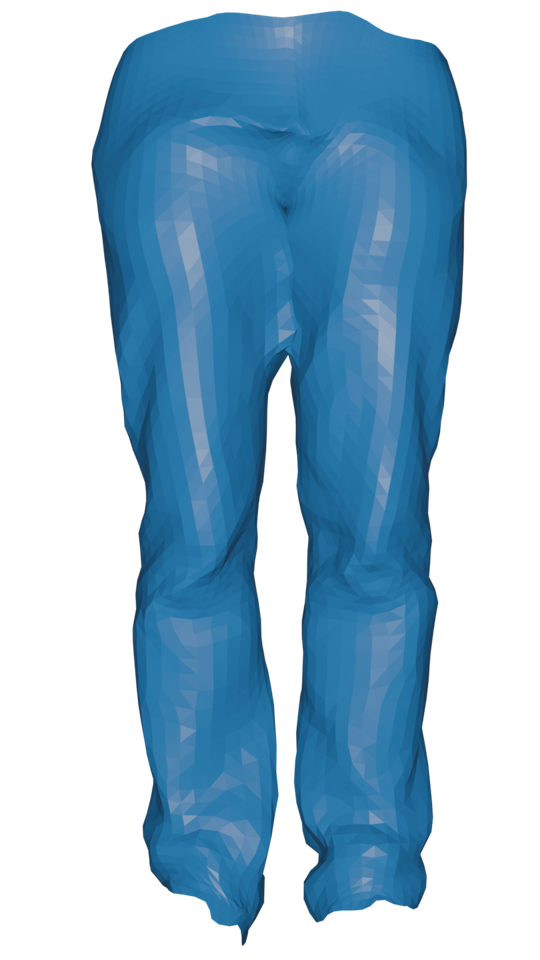}};
		\node[anchor=north west, inner sep=0] (ndf1) 
		at (0.34\textwidth,0) {\includegraphics[width=0.16\linewidth]{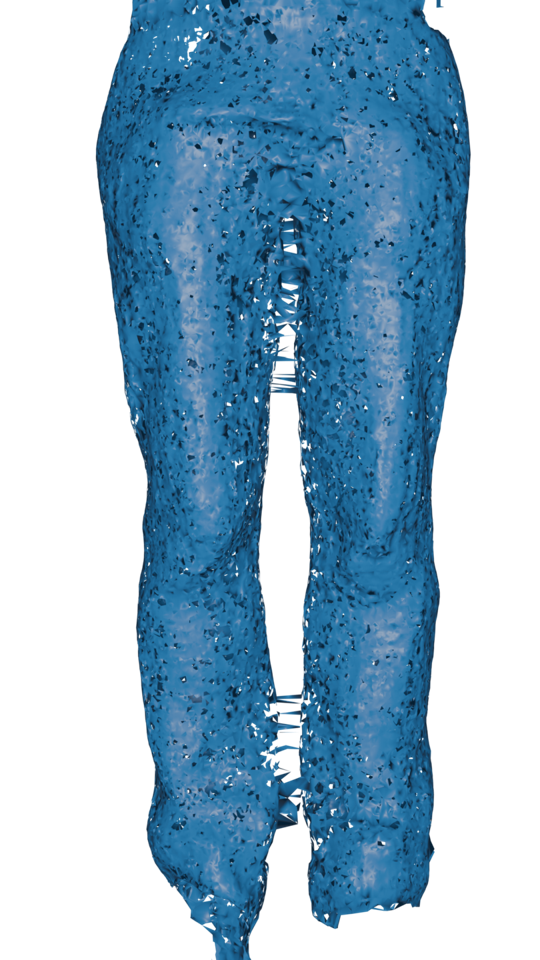}};
		\node[anchor=north west, inner sep=0] (cap1) 
		at (0.51\textwidth,0) {\includegraphics[width=0.16\linewidth]{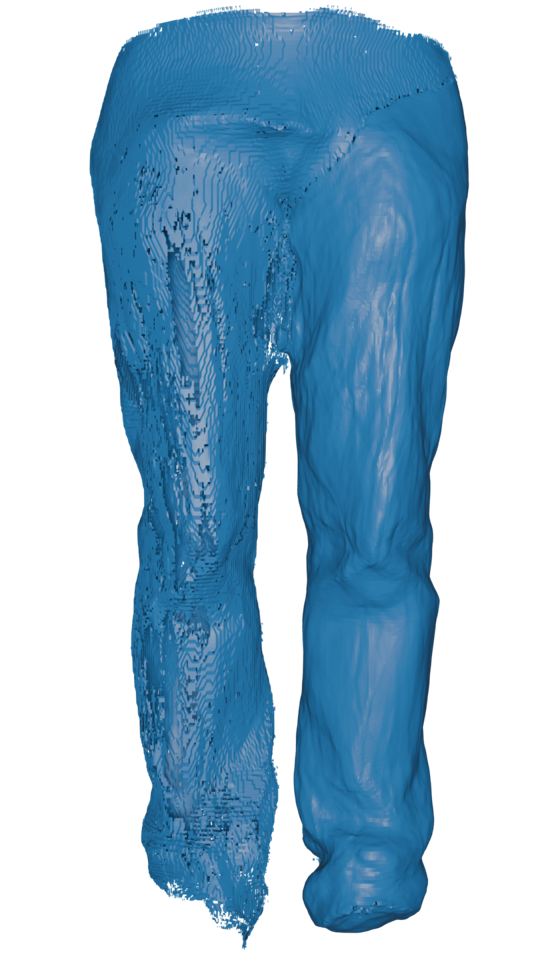}};
		\node[anchor=north west, inner sep=0] (csp1) 
		at (0.68\textwidth,0) {\includegraphics[width=0.16\linewidth]{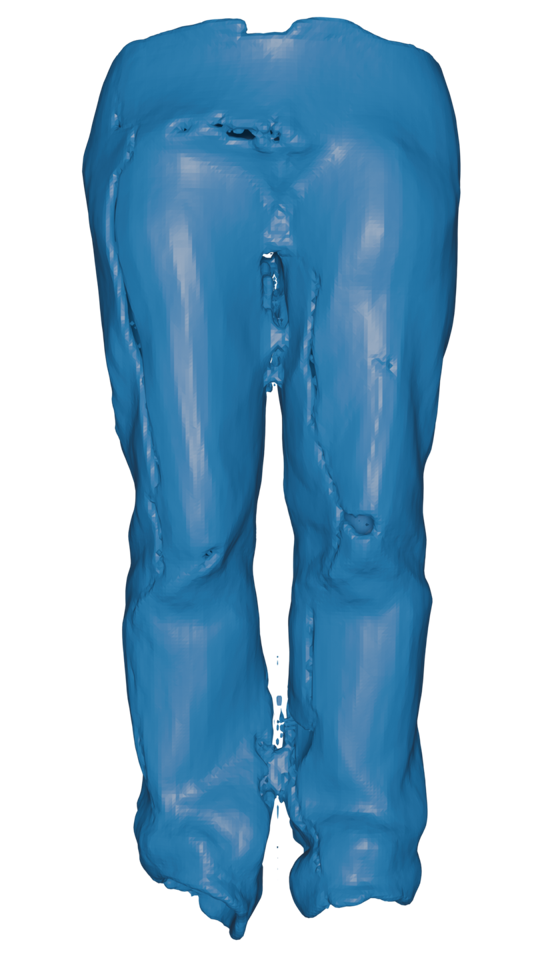}};
		\node[anchor=north west, inner sep=0] (nsp1) 
		at (0.85\textwidth,0) {\includegraphics[width=0.16\linewidth]{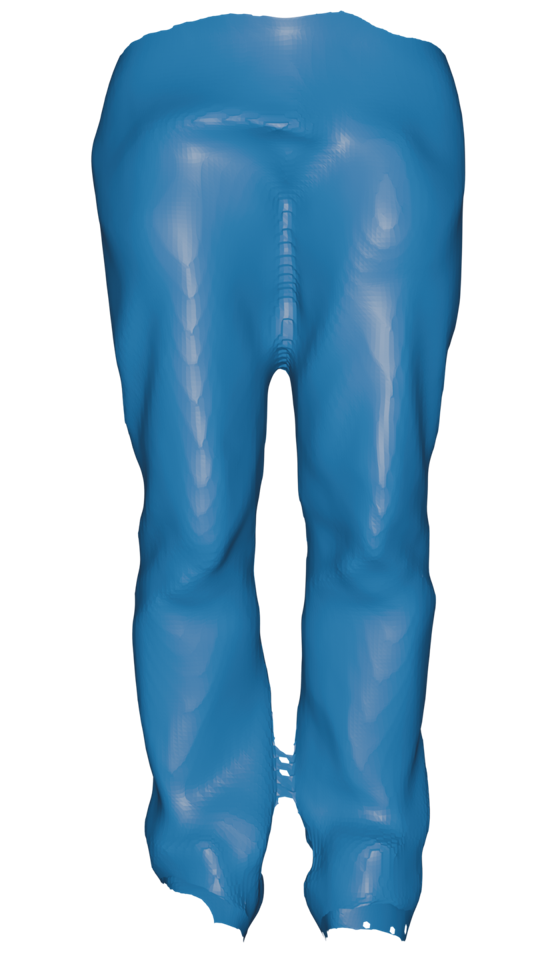}};
    	
    	\node[anchor=north west, inner sep=0] (input2) 
    	at (0,-3.8) {\includegraphics[width=0.16\linewidth]{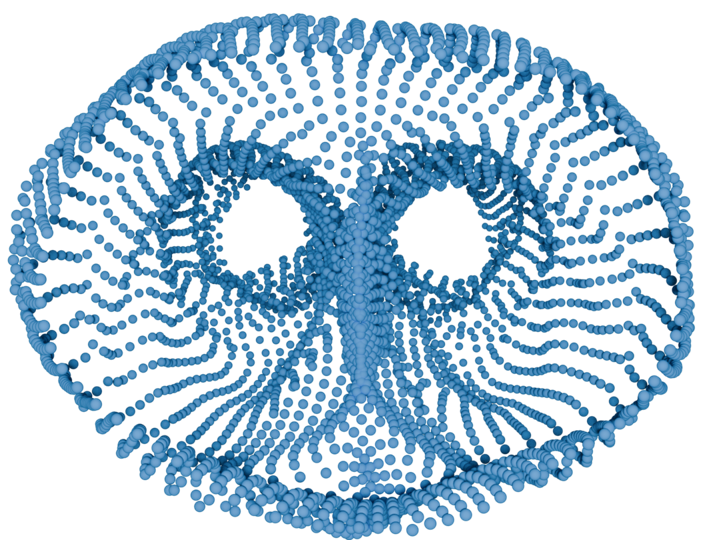}};
    	\node[anchor=north west, inner sep=0] (gt2) 
    	at (0.17\textwidth,-3.8) {\includegraphics[width=0.16\linewidth]{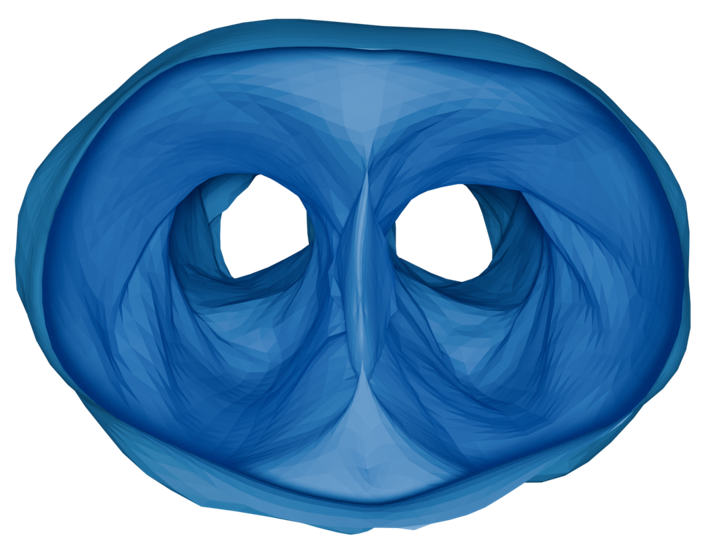}};
    	\node[anchor=north west, inner sep=0] (ndf2) 
    	at (0.34\textwidth,-3.8) {\includegraphics[width=0.16\linewidth]{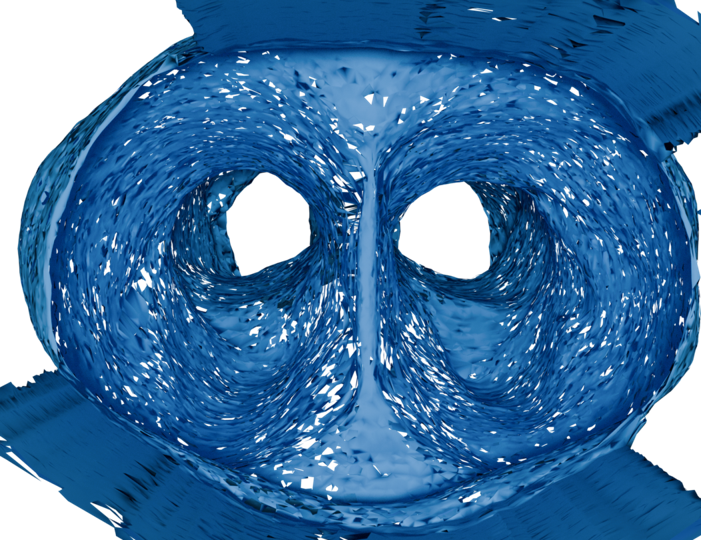}};
    	\node[anchor=north west, inner sep=0] (cap2) 
    	at (0.51\textwidth,-3.8) {\includegraphics[width=0.16\linewidth]{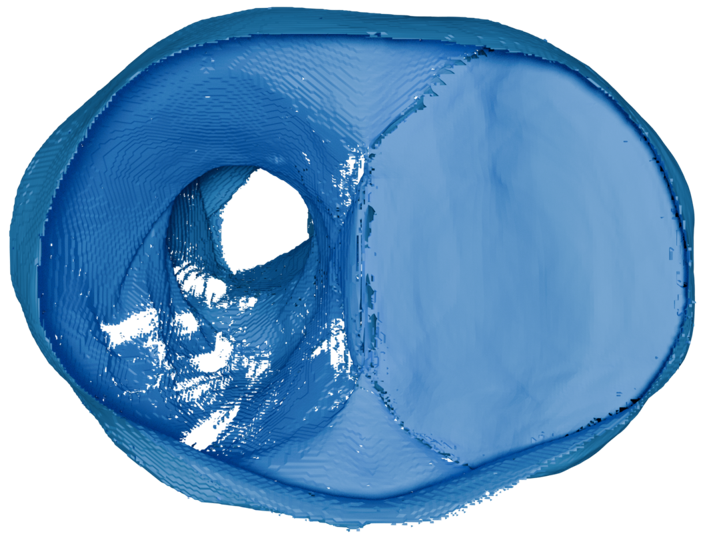}};
    	\node[anchor=north west, inner sep=0] (csp2) 
    	at (0.68\textwidth,-3.8) {\includegraphics[width=0.16\linewidth]{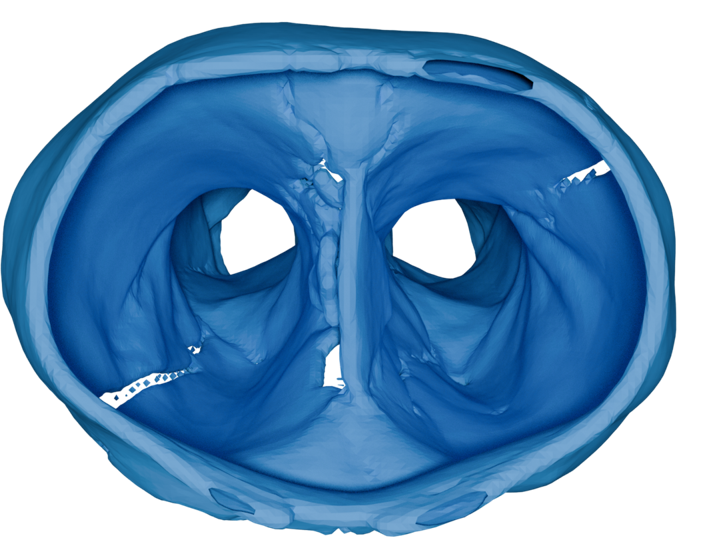}};
    	\node[anchor=north west, inner sep=0] (nsp2) 
    	at (0.85\textwidth,-3.8) {\includegraphics[width=0.16\linewidth]{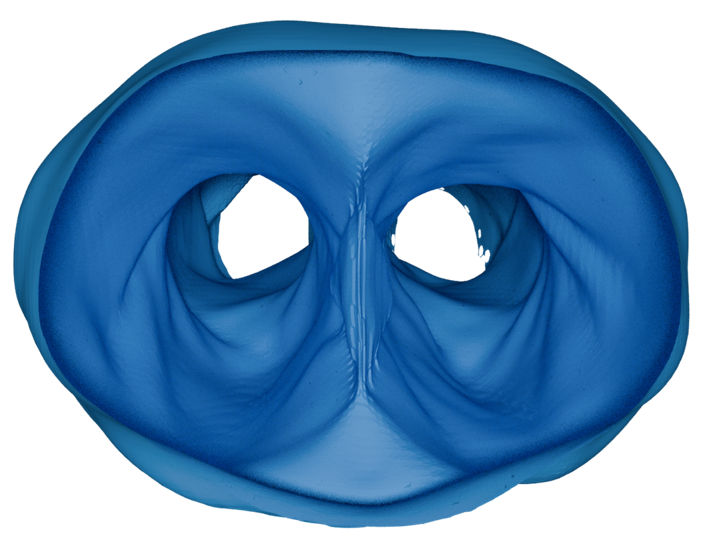}};
    	
    	\node[rotate=90, anchor=north] at ($(input1.west)!0.5!(input1.west) + (0,0)$) {Front};
    	\node[rotate=90, anchor=north] at ($(input2.west)!0.5!(input2.west) + (0,0.75)$) {Top};
    	
    	\node[rotate=90, anchor=north] at ($(input1.east)!0.5!(input1.east) + (-0.3,-1.6)$) {Input};
    	\node[rotate=90, anchor=north] at ($(gt1.east)!0.5!(gt1.east) + (-0.3,-1.6)$) {GT};
    	\node[rotate=90, anchor=north] at ($(ndf1.east)!0.5!(ndf1.east) + (-0.3,-1.6)$) {NDF};
    	\node[rotate=90, anchor=north] at ($(cap1.east)!0.5!(cap1.east) + (-0.3,-1.6)$) {CAP};
    	\node[rotate=90, anchor=north] at ($(csp1.east)!0.5!(csp1.east) + (-0.3,-1.6)$) {CSP};
    	\node[rotate=90, anchor=north] at ($(nsp1.east)!0.5!(nsp1.east) + (-0.3,-1.6)$) {\textbf{NSP}};    	
    \end{tikzpicture}
    \caption{Reconstructed surfaces on the pants data of MGN dataset.}
    \label{fig:pants}
\end{figure}

\begin{figure}
    \centering
    \begin{tikzpicture}[every node/.style={font=\tiny}]
	\node[anchor=south west, inner sep=0] (input1) 
	at (0,0) {\includegraphics[width=0.33\linewidth]{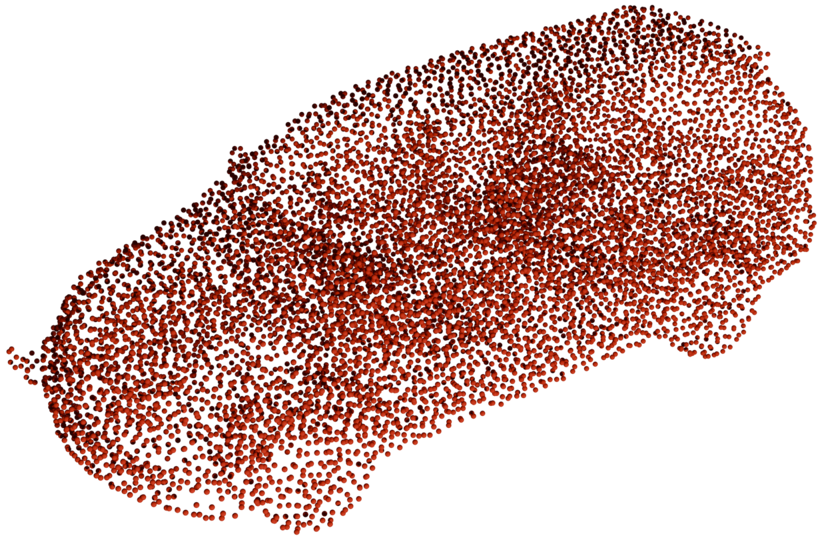}};
	\node[anchor=north east,inner sep=0, xshift=-5pt, yshift=20pt] at (input1.south east) {Input};
	\node[anchor=south west, inner sep=0] (gt1) 
	at (0.34\linewidth,0) {\includegraphics[width=0.33\linewidth]{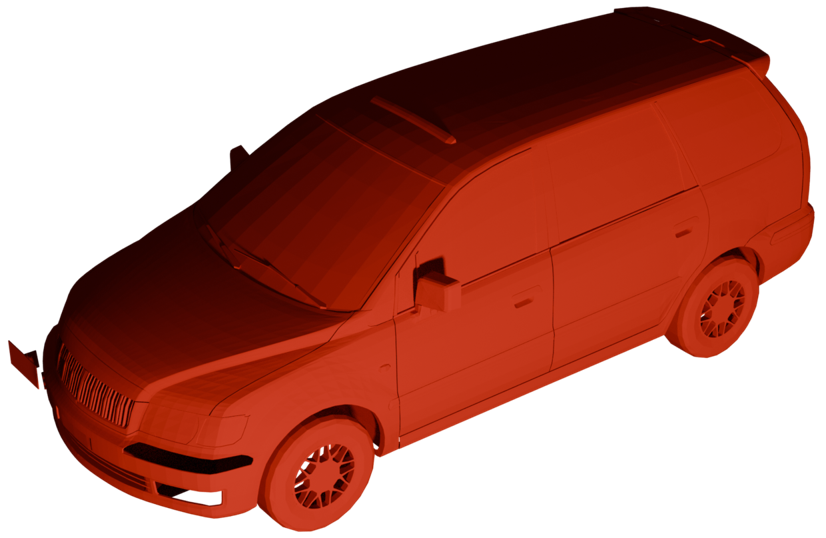}};
	\node[anchor=north east,inner sep=0, xshift=-5pt, yshift=20pt] at (gt1.south east) {GT};
	\node[anchor=south west, inner sep=0] (ndf1) 
	at (0.68\linewidth,0) {\includegraphics[width=0.33\linewidth]{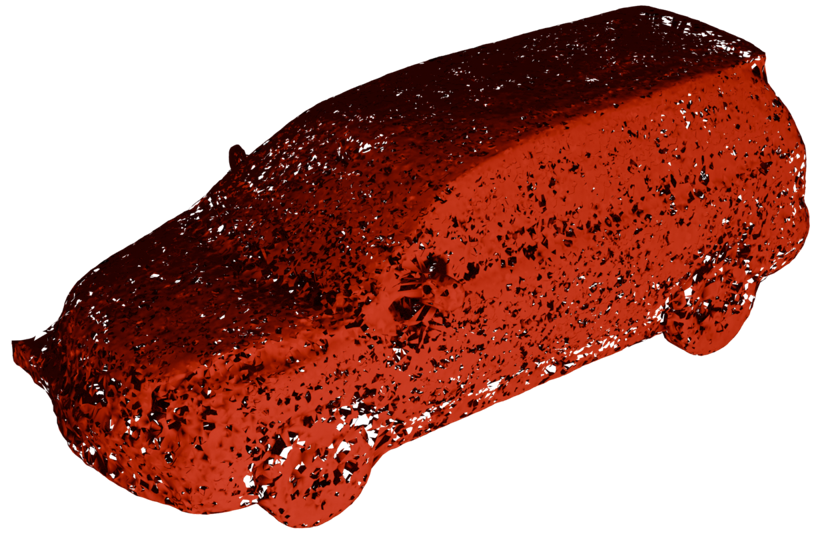}};
	\node[anchor=north east,inner sep=0, xshift=-5pt, yshift=20pt] at (ndf1.south east) {NDF};
	
	\node[anchor=north west, inner sep=0] (cap1) 
	at (0,0.3) {\includegraphics[width=0.33\linewidth]{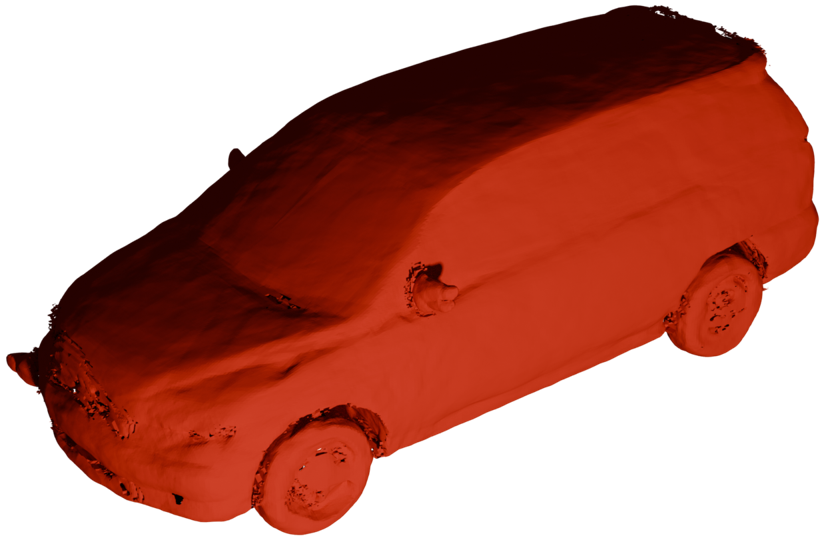}};
	\node[anchor=north east,inner sep=0, xshift=-5pt, yshift=20pt] at (cap1.south east) {CAP};
	\node[anchor=north west, inner sep=0] (csp1) 
	at (0.34\linewidth,0.3) {\includegraphics[width=0.33\linewidth]{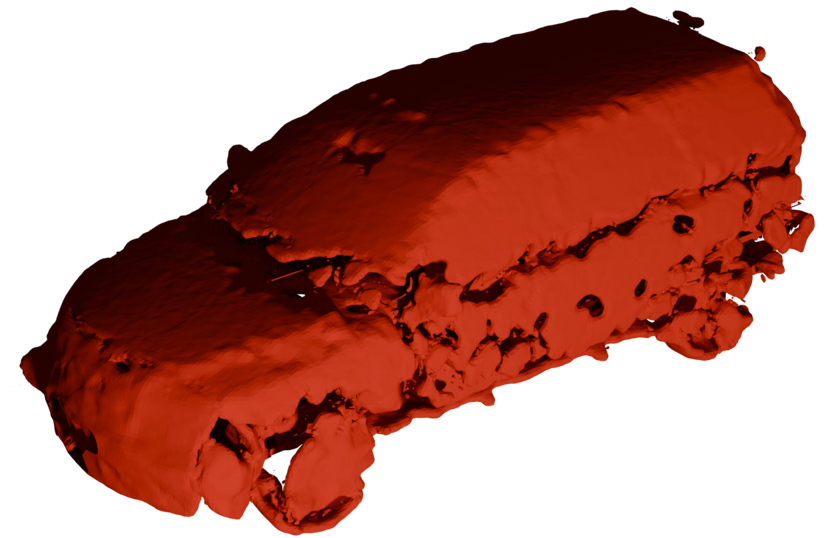}};
	\node[anchor=north east,inner sep=0, xshift=-5pt, yshift=20pt] at (csp1.south east) {CSP};
	\node[anchor=north west, inner sep=0] (nsp1) 
	at (0.68\linewidth,0.3) {\includegraphics[width=0.33\linewidth]{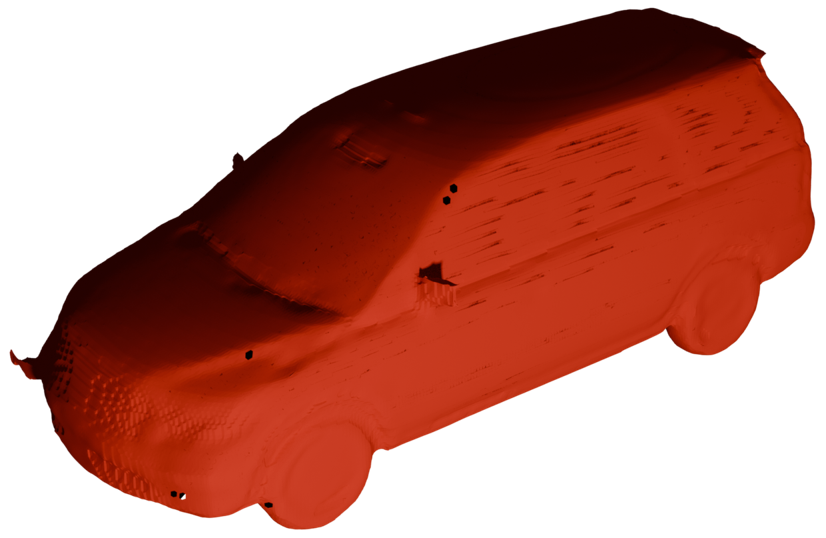}};
	\node[anchor=north east,inner sep=0, xshift=-5pt, yshift=20pt] at (nsp1.south east) {\textbf{NSP}};
	
	\node[anchor=north west, inner sep=0] (input2) 
	at (0,-2.8) {\includegraphics[width=0.33\linewidth]{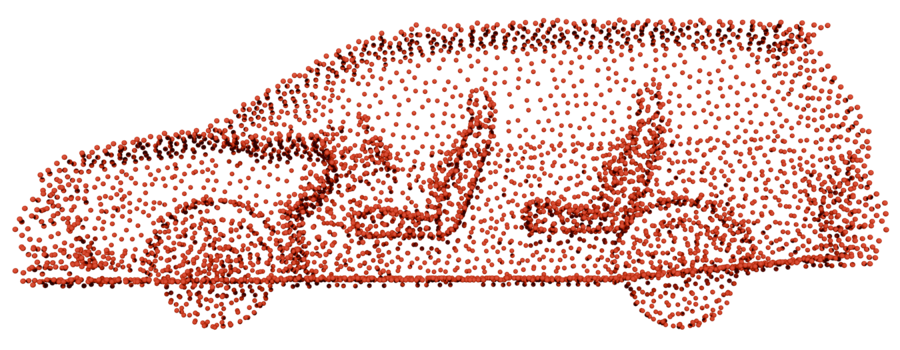}};
	\node[anchor=north west,inner sep=0, xshift=15pt, yshift=-8pt] at (input2.north west) {Input};
	\node[anchor=north west, inner sep=0] (gt2) 
	at (0.34\linewidth,-2.8) {\includegraphics[width=0.33\linewidth]{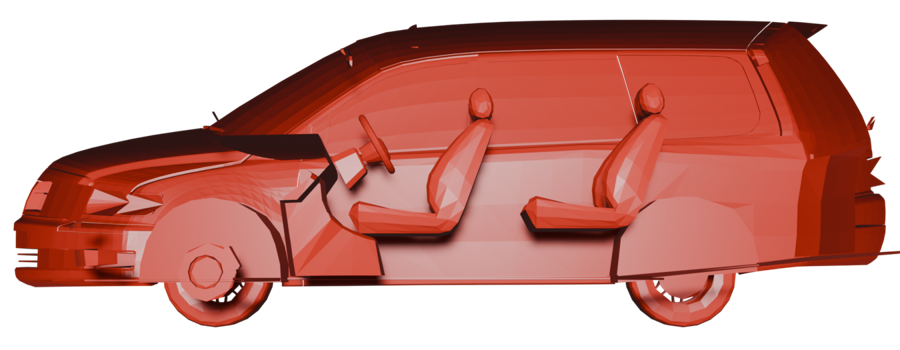}};
	\node[anchor=north west,inner sep=0, xshift=15pt, yshift=-8pt] at (gt2.north west) {GT};
	\node[anchor=north west, inner sep=0] (ndf2) 
	at (0.68\linewidth,-2.8) {\includegraphics[width=0.33\linewidth]{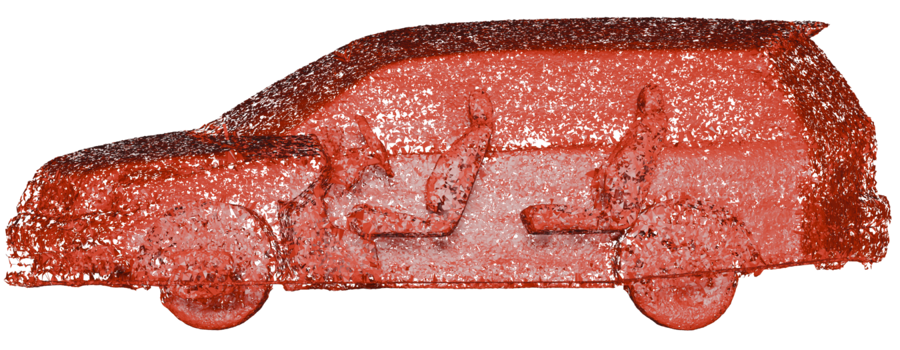}};
	\node[anchor=north west,inner sep=0, xshift=15pt, yshift=-8pt] at (ndf2.north west) {NDF};
	
	\node[anchor=north west, inner sep=0] (cap2) 
	at (0,-4.5) {\includegraphics[width=0.33\linewidth]{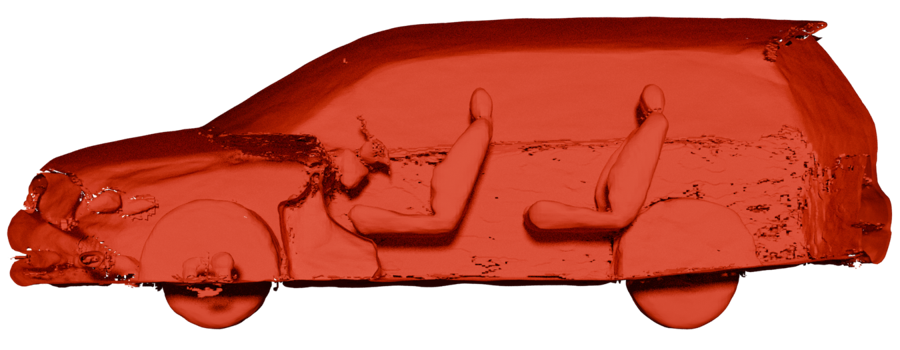}};
	\node[anchor=north west,inner sep=0, xshift=15pt, yshift=-8pt] at (cap2.north west) {CAP};
	\node[anchor=north west, inner sep=0] (csp2) 
	at (0.34\linewidth,-4.5) {\includegraphics[width=0.33\linewidth]{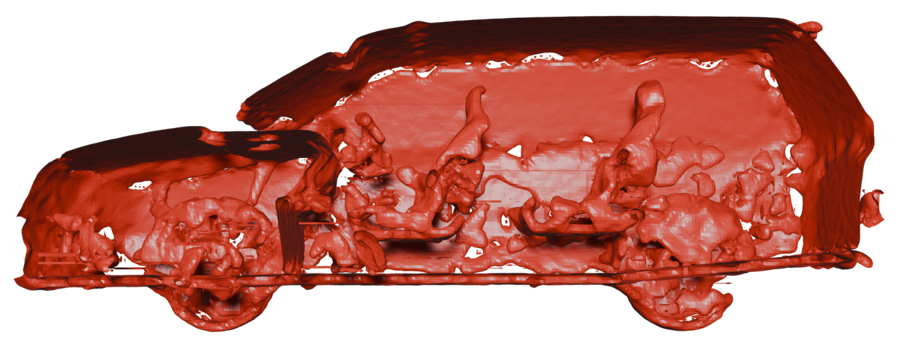}};
	\node[anchor=north west,inner sep=0, xshift=15pt, yshift=-8pt] at (csp2.north west) {CSP};
	\node[anchor=north west, inner sep=0] (nsp2) 
	at (0.68\linewidth,-4.5) {\includegraphics[width=0.33\linewidth]{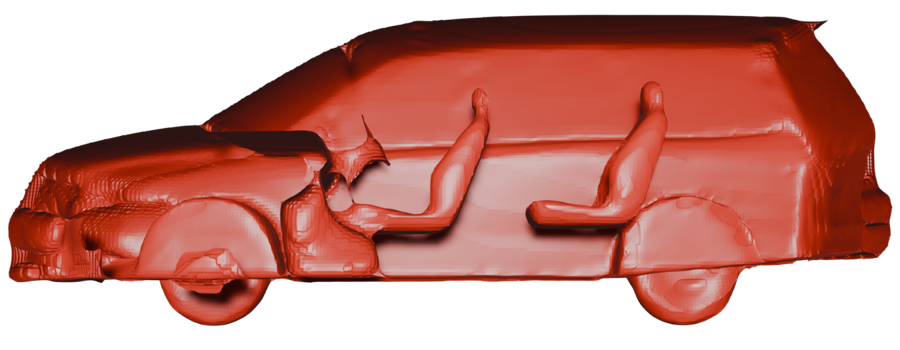}};
	\node[anchor=north west,inner sep=0, xshift=15pt, yshift=-8pt] at (nsp2.north west) {\textbf{NSP}};
		
	\node[rotate=90, anchor=north] at ($(input1.west)!0.5!(cap1.west) + (0,-0.5)$) {External};
	\node[rotate=90, anchor=north] at ($(input2.west)!0.5!(cap2.west) + (0,-0.25)$) {Internal};
    \end{tikzpicture}
    \caption{Reconstructed surfaces on ShapeNet car data.}
    \label{fig:car1}
\end{figure}
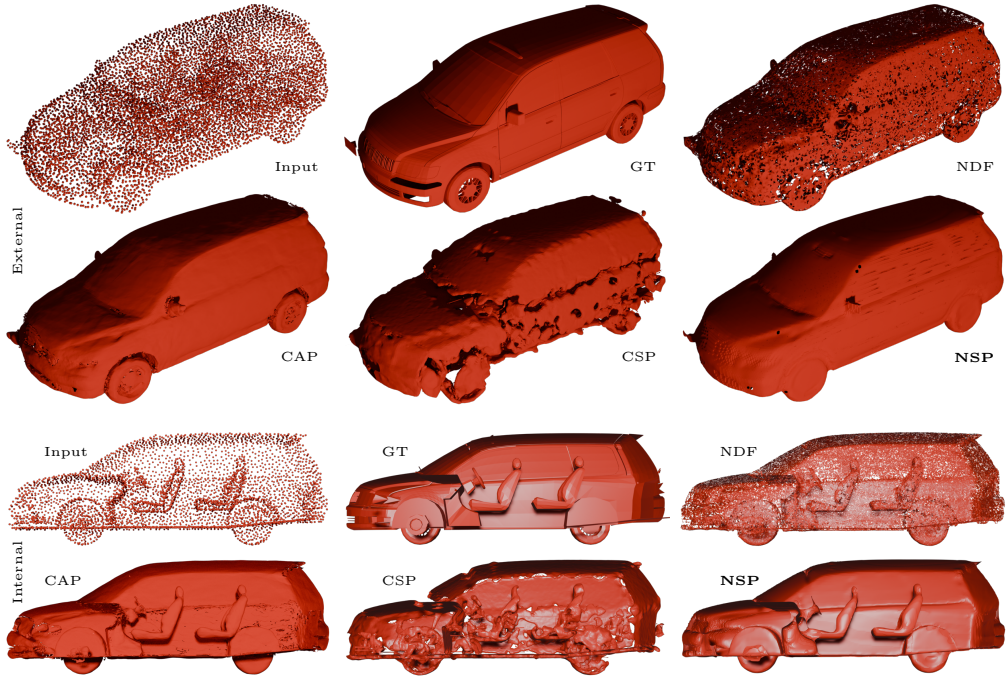

\begin{figure}
    \centering
        \begin{tikzpicture}[every node/.style={font=\tiny}]
    	\node[anchor=south west, inner sep=0] (input1) 
    	at (0,0) {\includegraphics[width=0.33\linewidth]{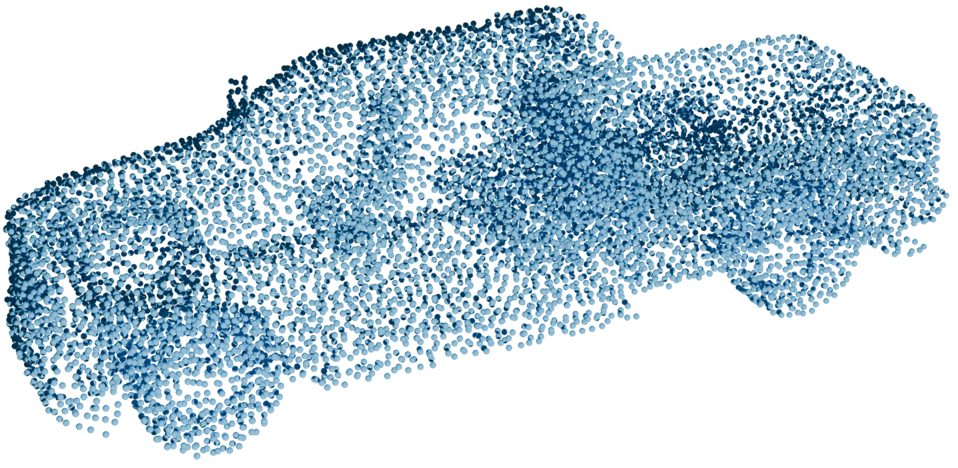}};
    	\node[anchor=north east,inner sep=0, xshift=-5pt, yshift=15pt] at (input1.south east) {Input};
    	\node[anchor=south west, inner sep=0] (gt1) 
    	at (0.34\linewidth,0) {\includegraphics[width=0.33\linewidth]{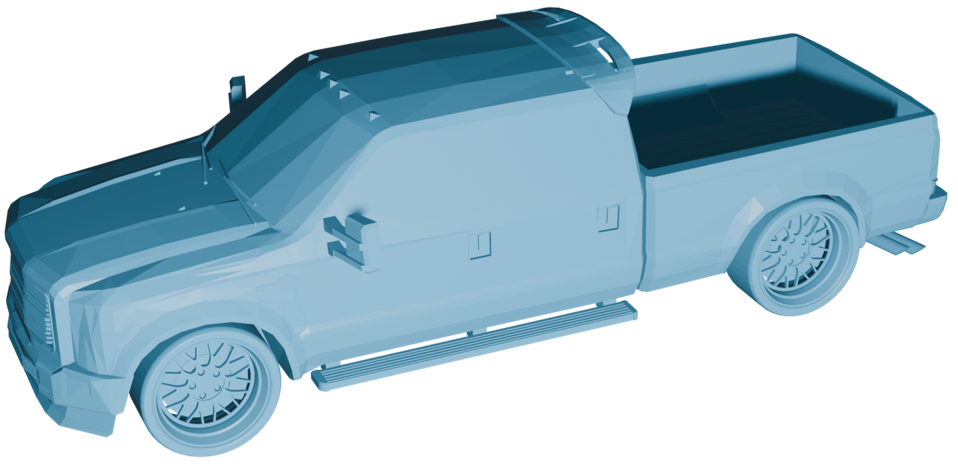}};
    	\node[anchor=north east,inner sep=0, xshift=-5pt, yshift=15pt] at (gt1.south east) {GT};
    	\node[anchor=south west, inner sep=0] (ndf1) 
    	at (0.68\linewidth,0) {\includegraphics[width=0.33\linewidth]{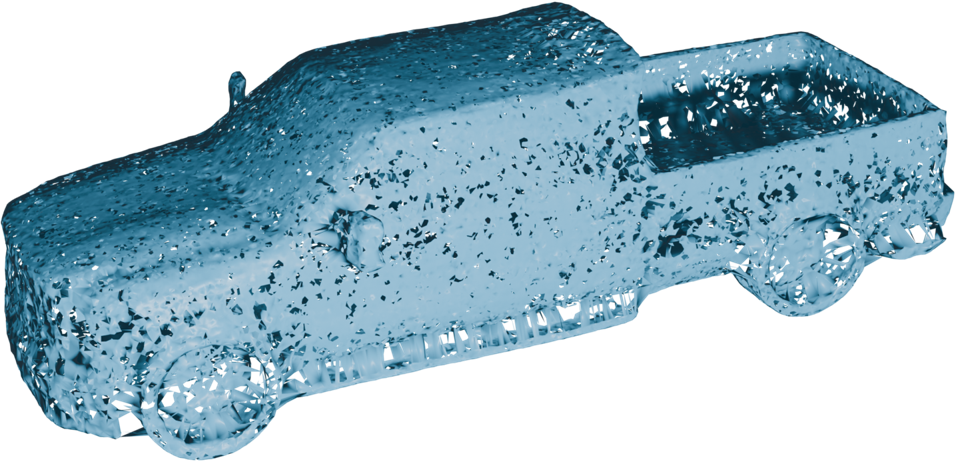}};
    	\node[anchor=north east,inner sep=0, xshift=-5pt, yshift=15pt] at (ndf1.south east) {NDF};
    	
    	\node[anchor=north west, inner sep=0] (cap1) 
    	at (0,0.3) {\includegraphics[width=0.33\linewidth]{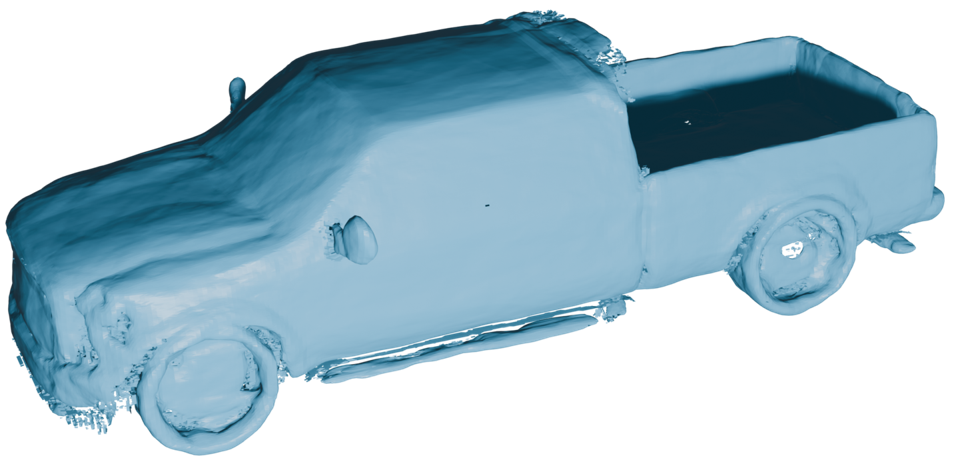}};
    	\node[anchor=north east,inner sep=0, xshift=-5pt, yshift=15pt] at (cap1.south east) {CAP};
    	\node[anchor=north west, inner sep=0] (csp1) 
    	at (0.34\linewidth,0.3) {\includegraphics[width=0.33\linewidth]{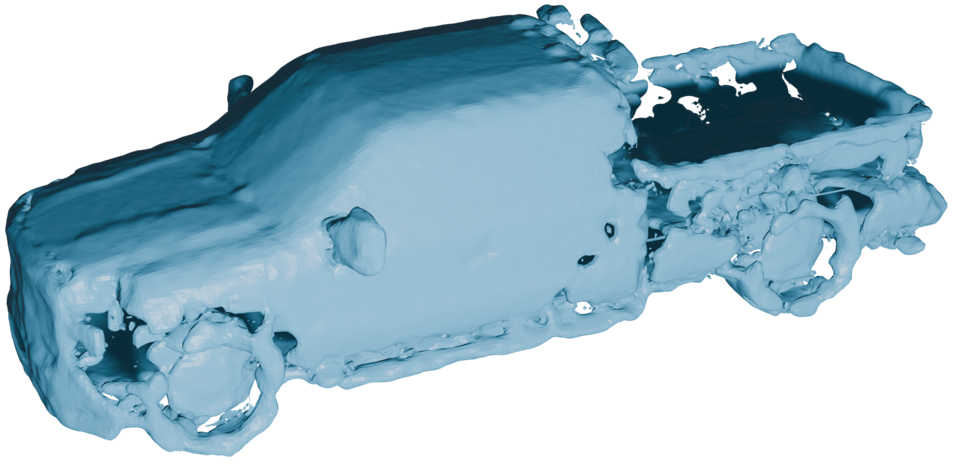}};
    	\node[anchor=north east,inner sep=0, xshift=-5pt, yshift=15pt] at (csp1.south east) {CSP};
    	\node[anchor=north west, inner sep=0] (nsp1) 
    	at (0.68\linewidth,0.3) {\includegraphics[width=0.33\linewidth]{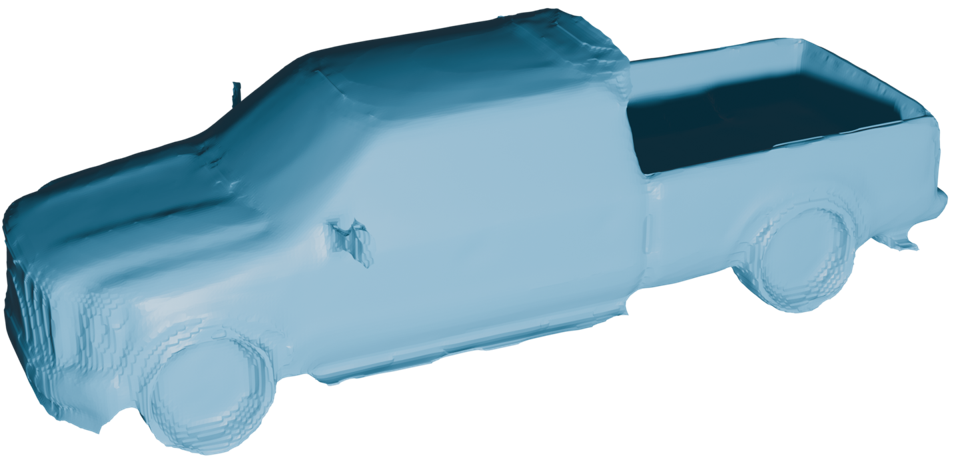}};
    	\node[anchor=north east,inner sep=0, xshift=-5pt, yshift=15pt] at (nsp1.south east) {\textbf{NSP}};
    	
    	\node[anchor=north west, inner sep=0] (input2) 
    	at (0,-2.0) {\includegraphics[width=0.33\linewidth]{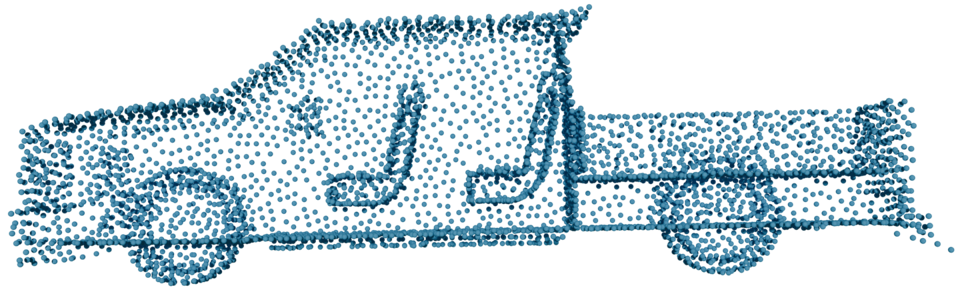}};
    	\node[anchor=north east,inner sep=0, xshift=-8pt, yshift=-7pt] at (input2.north east) {Input};
    	\node[anchor=north west, inner sep=0] (gt2) 
    	at (0.34\linewidth,-2.0) {\includegraphics[width=0.33\linewidth]{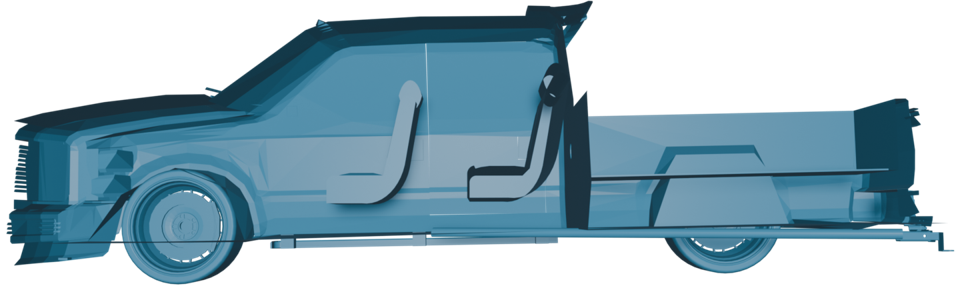}};
    	\node[anchor=north east,inner sep=0, xshift=-8pt, yshift=-7pt] at (gt2.north east) {GT};
    	\node[anchor=north west, inner sep=0] (ndf2) 
    	at (0.68\linewidth,-2.0) {\includegraphics[width=0.33\linewidth]{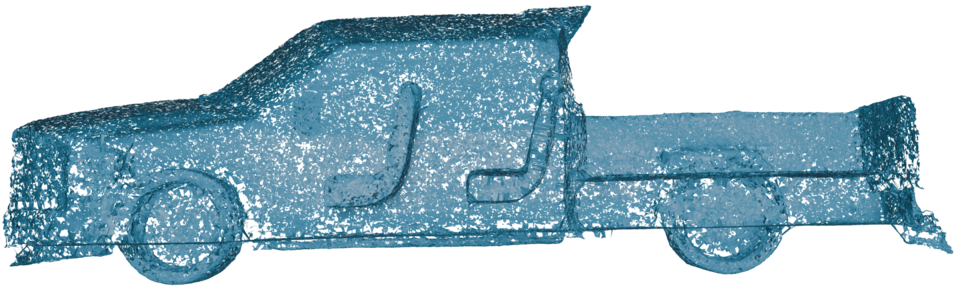}};
    	\node[anchor=north east,inner sep=0, xshift=-8pt, yshift=-7pt] at (ndf2.north east) {NDF};
    	
    	\node[anchor=north west, inner sep=0] (cap2) 
    	at (0,-3.5) {\includegraphics[width=0.33\linewidth]{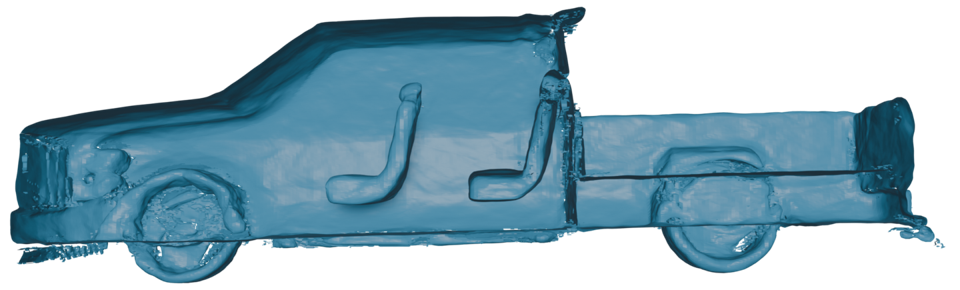}};
    	\node[anchor=north east,inner sep=0, xshift=-8pt, yshift=-7pt] at (cap2.north east) {CAP};
    	\node[anchor=north west, inner sep=0] (csp2) 
    	at (0.34\linewidth,-3.5) {\includegraphics[width=0.33\linewidth]{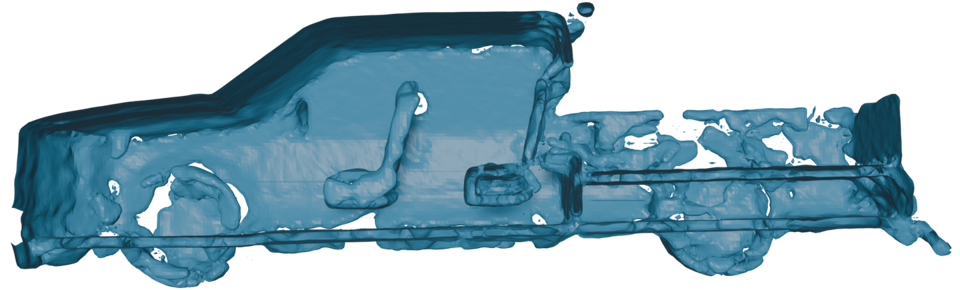}};
    	\node[anchor=north east,inner sep=0, xshift=-8pt, yshift=-7pt] at (csp2.north east) {CSP};
    	\node[anchor=north west, inner sep=0] (nsp2) 
    	at (0.68\linewidth,-3.5) {\includegraphics[width=0.33\linewidth]{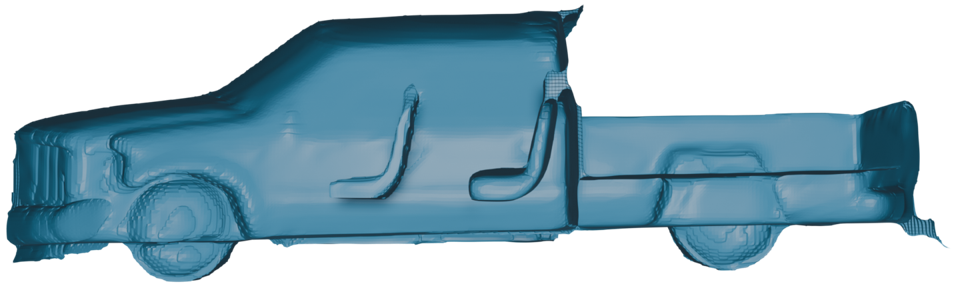}};
    	\node[anchor=north east,inner sep=0, xshift=-8pt, yshift=-7pt] at (nsp2.north east) {\textbf{NSP}};
    	
    	\node[rotate=90, anchor=north] at ($(input1.west)!0.5!(cap1.west) + (-0.27,-0.2)$) {External};
    	\node[rotate=90, anchor=north] at ($(input2.west)!0.5!(cap2.west) + (-0.27,-0.2)$) {Internal};
    \end{tikzpicture}
    \caption{Reconstructed surfaces on ShapeNet car data.}
    \label{fig:car2}
\end{figure}
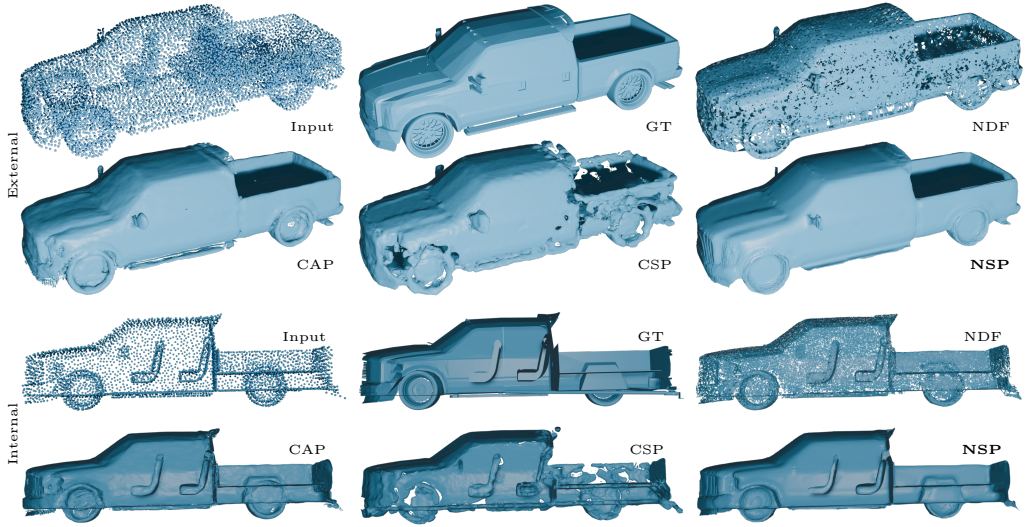

The mean and median of $d_C$ and $d_H$ of the reconstructed surfaces for each dataset are presented in Tables \ref{tab:mgn} and \ref{tab:cars}, respectively.
The results demonstrate that the proposed NSP method achieves performance that is on par with or superior to the baseline models. While Hausdorff and Chamfer distances are standard metrics for assessing the quality of reconstructed surfaces, they do not capture all surface characteristics. For a more comprehensive evaluation, the reconstructed surfaces are presented in Figures \ref{fig:shirt}, \ref{fig:pants}, \ref{fig:car1}, and \ref{fig:car2} with the input point cloud and the GT surfaces.
It is clear that methods such as NDF and CSP produce surfaces with noticeable gaps and lack smoothness in certain areas or loss of fine details. The CAP method fails to accurately reproduce the holes in the shirt sleeves and pants in the MGN dataset, while also generating irregular surfaces in the pants and ShapeNet car dataset. In contrast, the proposed NSP consistently produces much smoother surfaces compared to the other models. It also successfully reconstructs the holes in the shirt and pants of the MGN data and accurately captures the interior of the cars of the ShapeNet dataset. These results demonstrate that the NSP approach, which accurately learns both the distance function and the gradient, is effective in reconstructing scan data of complex shapes.

\subsection{Surface reconstruction for indoor scenes}\label{sec:exp_indoor}
To evaluate the performance of the proposed method for surface reconstruction from real 3D scans, we conducted experiments on a challenging 3D indoor scene dataset provided from \citep{zhou2013dense}. The dataset comprises point clouds with intricate geometries and noisy open surfaces, making it a suitable benchmark for testing the performance of the model. The reconstruction of 3D indoor scenes represents a valuable task within the field of computer vision, with numerous applications in areas such as computer graphics, virtual reality, and robotics.
\begin{table}
    \centering
    \caption{Results of surface reconstruction on 3D scene dataset.} \label{tab:indoor}
    \begin{tabular}{lcccccccc}
   	\toprule 
   	& \multicolumn{4}{c}{Mesh} & \multicolumn{4}{c}{Points}\\
   	\cmidrule(lr){2-5} \cmidrule(lr){6-9}
   	& \multicolumn{2}{c}{$d_C$} & \multicolumn{2}{c}{$d_H$} & \multicolumn{2}{c}{$d_C$} & \multicolumn{2}{c}{$d_H$}\\
   	\cmidrule(lr){2-3} \cmidrule(lr){4-5} \cmidrule(lr){6-7} \cmidrule(lr){8-9}
   	Model & Mean & Median & Mean & Median & Mean & Median & Mean & Median\\
   	\cmidrule(lr){1-1} \cmidrule(lr){2-3} \cmidrule(lr){4-5} \cmidrule(lr){6-7} \cmidrule(lr){8-9}
     NDF    & 2.0393 & 2.2410 & 5.3444 & 5.8414 & 0.1395 & 0.1483 & 0.3860 & 0.3922\\
     CAP    &  2.3404 & 2.3003 & 6.3647 & 6.1584 & 0.0026 & 0.0014 & 0.0359 & 0.0116\\
     CSP    &  2.1160 & 2.3024 & 5.2513 & 5.7309 & 0.0052 & 0.0052 & 0.1724 & 0.0116\\
     \textbf{NSP} & 2.3484 & 2.3094 & 6.3791 & 6.1850 & 0.0022 & 0.0024 & 0.0470 & 0.2330\\
     \bottomrule
    \end{tabular}
\end{table}

\begin{figure}
    \centering
	\begin{tikzpicture}[every node/.style={font=\tiny}]
	\node[anchor=south west, inner sep=0] (input1) 
	at (0,0) {\includegraphics[width=0.33\linewidth]{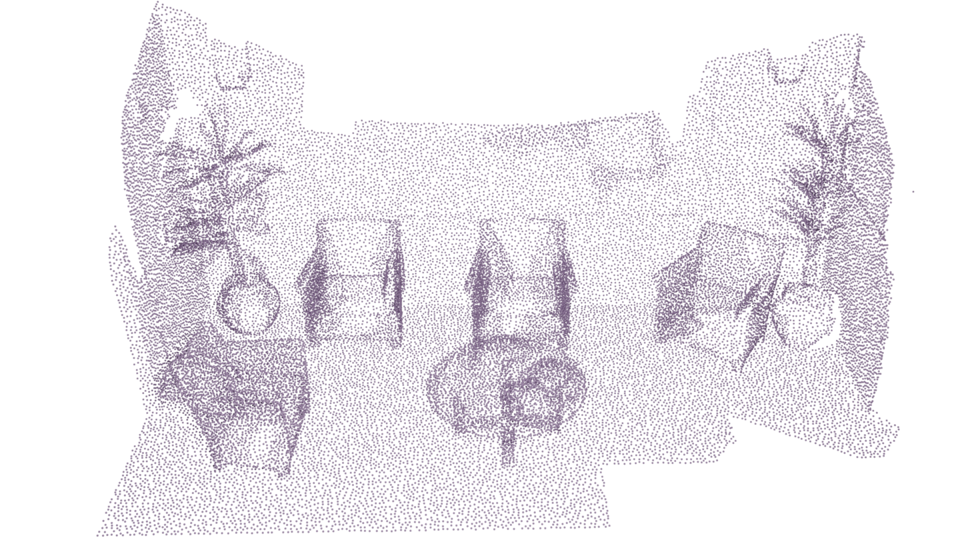}};
	\node[anchor=north,inner sep=0, xshift=0pt, yshift=-5pt] at (input1.north) {Input};
	\node[anchor=south west, inner sep=0] (gt1) 
	at (0.34\linewidth,0) {\includegraphics[width=0.33\linewidth]{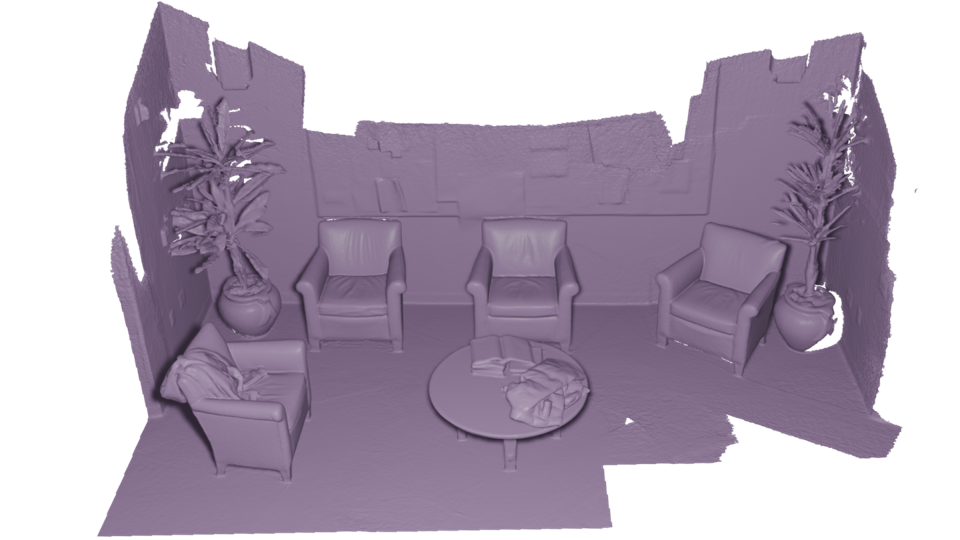}};
	\node[anchor=north,inner sep=0, xshift=0pt, yshift=-5pt] at (gt1.north) {GT};
	\node[anchor=south west, inner sep=0] (ndf1) 
	at (0.68\linewidth,0) {\includegraphics[width=0.33\linewidth]{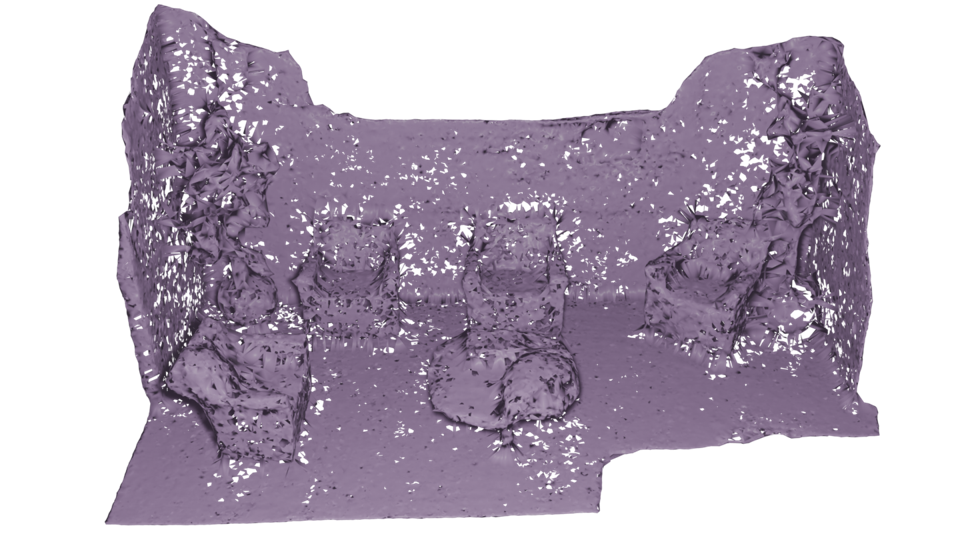}};
	\node[anchor=north,inner sep=0, xshift=-5pt, yshift=-5pt] at (ndf1.north) {NDF};
	
	\node[anchor=north west, inner sep=0] (cap1) 
	at (0,-0.15) {\includegraphics[width=0.33\linewidth]{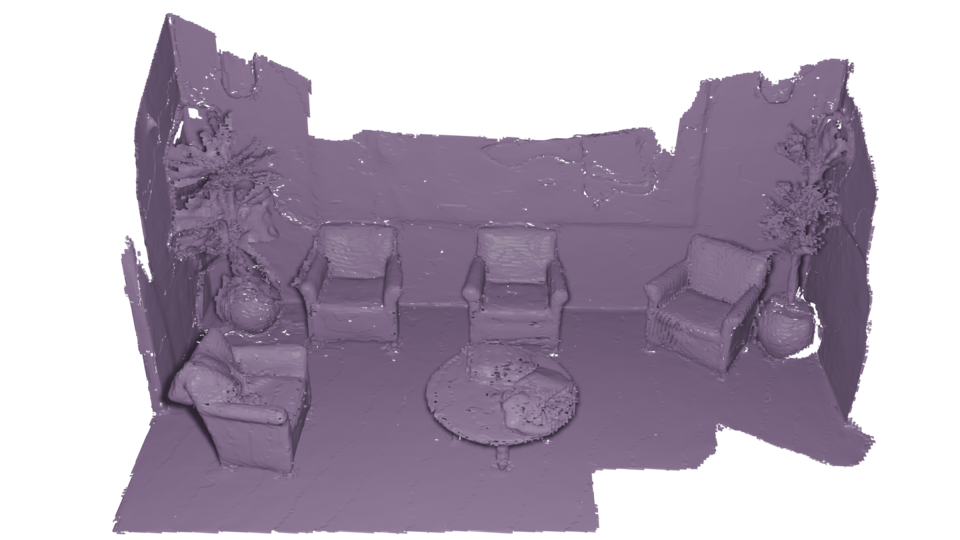}};
	\node[anchor=north,inner sep=0, xshift=0pt, yshift=-5pt] at (cap1.north) {CAP};
	\node[anchor=north west, inner sep=0] (csp1) 
	at (0.34\linewidth,-0.15) {\includegraphics[width=0.33\linewidth]{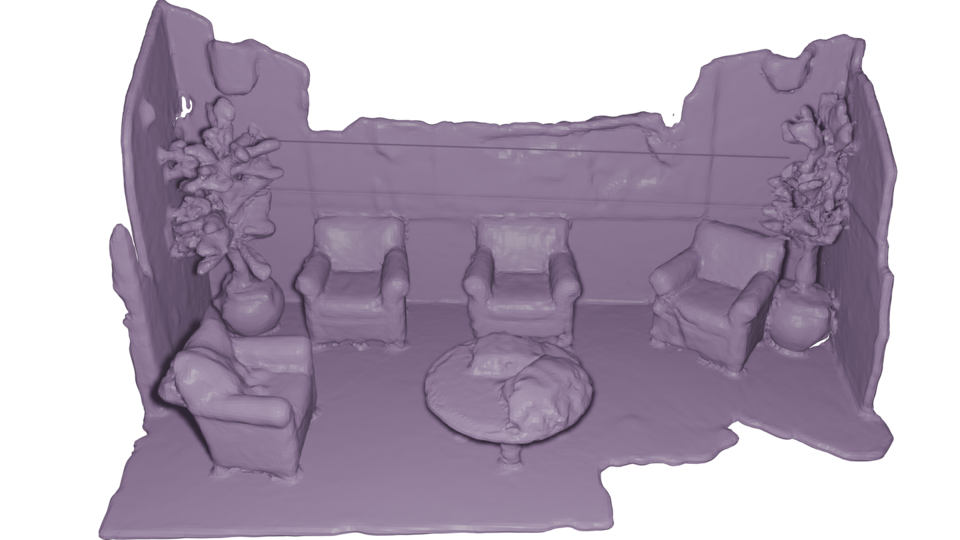}};
	\node[anchor=north,inner sep=0, xshift=0pt, yshift=-5pt] at (csp1.north) {CSP};
	\node[anchor=north west, inner sep=0] (nsp1) 
	at (0.68\linewidth,-0.15) {\includegraphics[width=0.33\linewidth]{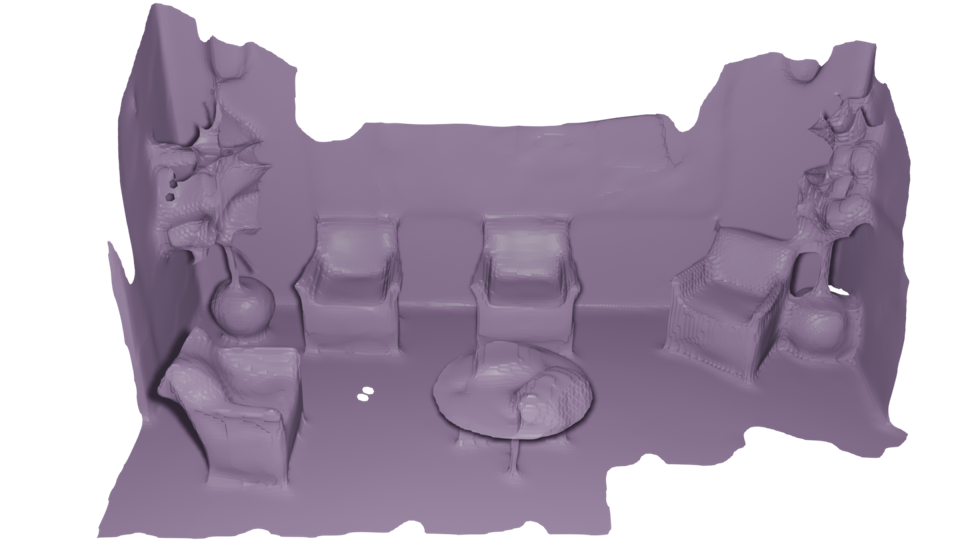}};
	\node[anchor=north,inner sep=0, xshift=0pt, yshift=-5pt] at (nsp1.north) {\textbf{NSP}};
	
	\node[anchor=north west, inner sep=0] (input2) 
	at (0,-2.9) {\includegraphics[width=0.33\linewidth]{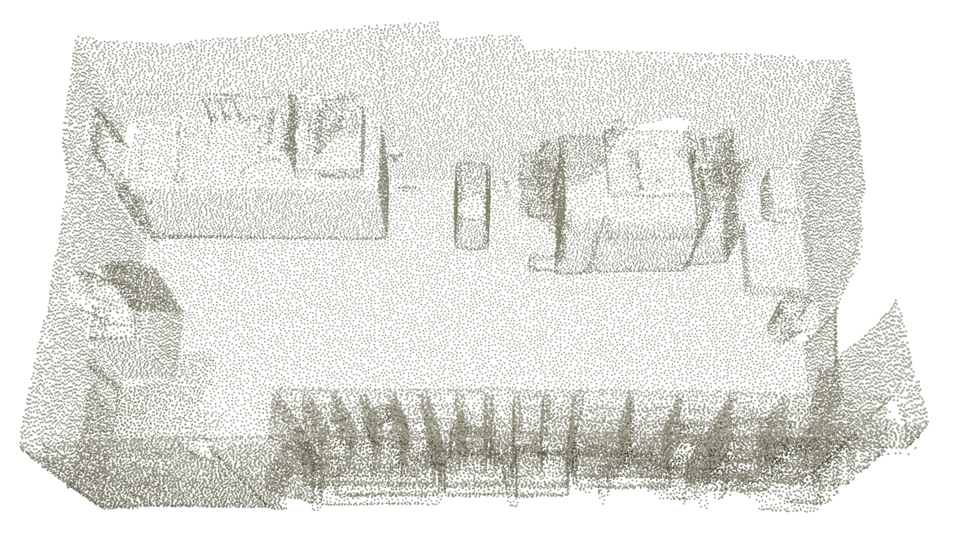}};
	\node[anchor=north east,inner sep=0, xshift=-12pt, yshift=-1pt] at (input2.north east) {Input};
	\node[anchor=north west, inner sep=0] (gt2) 
	at (0.34\linewidth,-2.9) {\includegraphics[width=0.33\linewidth]{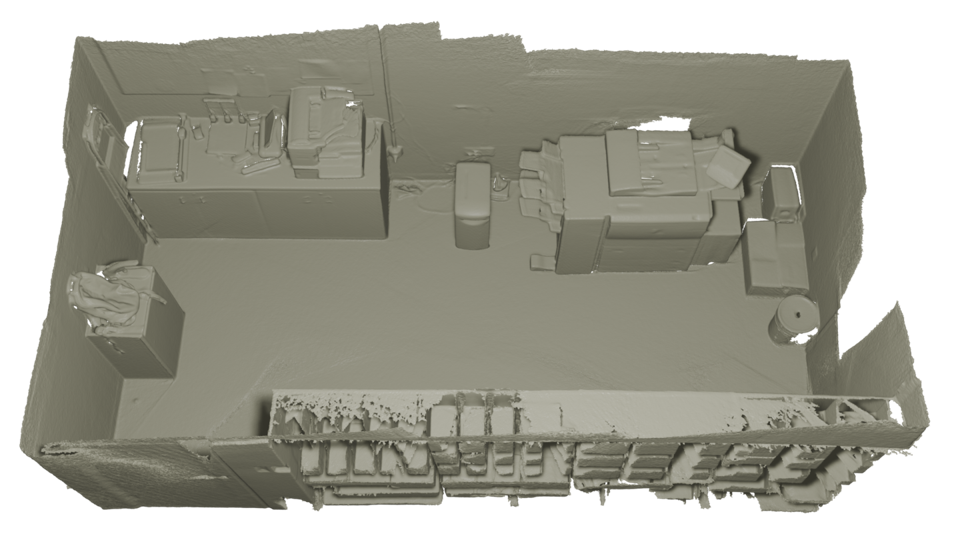}};
	\node[anchor=north east,inner sep=0, xshift=-12pt, yshift=-1pt] at (gt2.north east) {GT};
	\node[anchor=north west, inner sep=0] (ndf2) 
	at (0.68\linewidth,-2.9) {\includegraphics[width=0.33\linewidth]{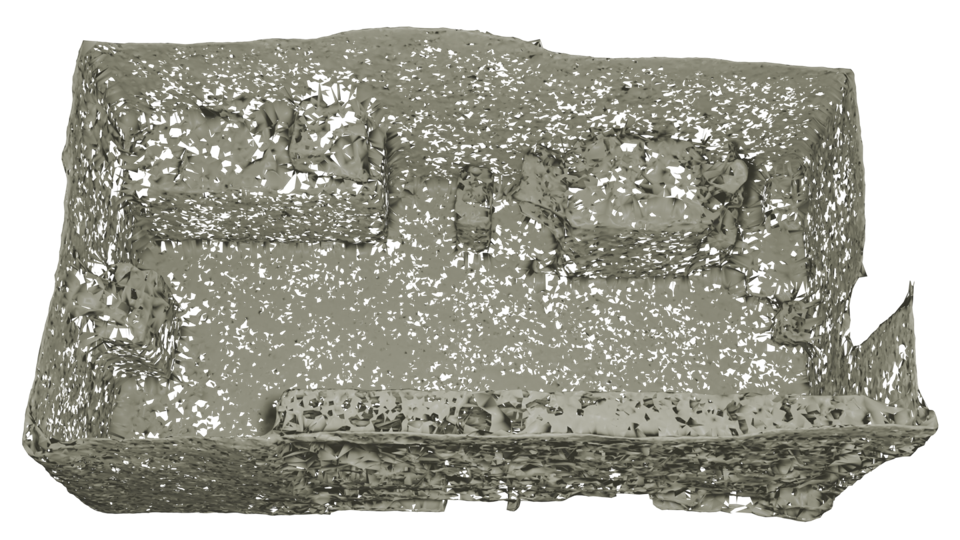}};
	\node[anchor=north east,inner sep=0, xshift=-12pt, yshift=-1pt] at (ndf2.north east) {NDF};
	
	\node[anchor=north west, inner sep=0] (cap2) 
	at (0,-5.3) {\includegraphics[width=0.33\linewidth]{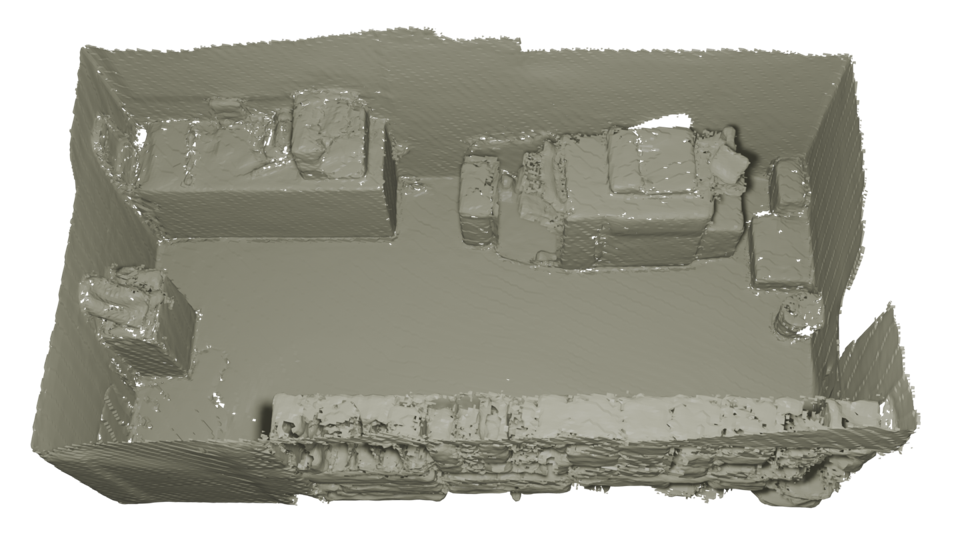}};
	\node[anchor=north east,inner sep=0, xshift=-12pt, yshift=-1pt] at (cap2.north east) {CAP};
	\node[anchor=north west, inner sep=0] (csp2) 
	at (0.34\linewidth,-5.3) {\includegraphics[width=0.33\linewidth]{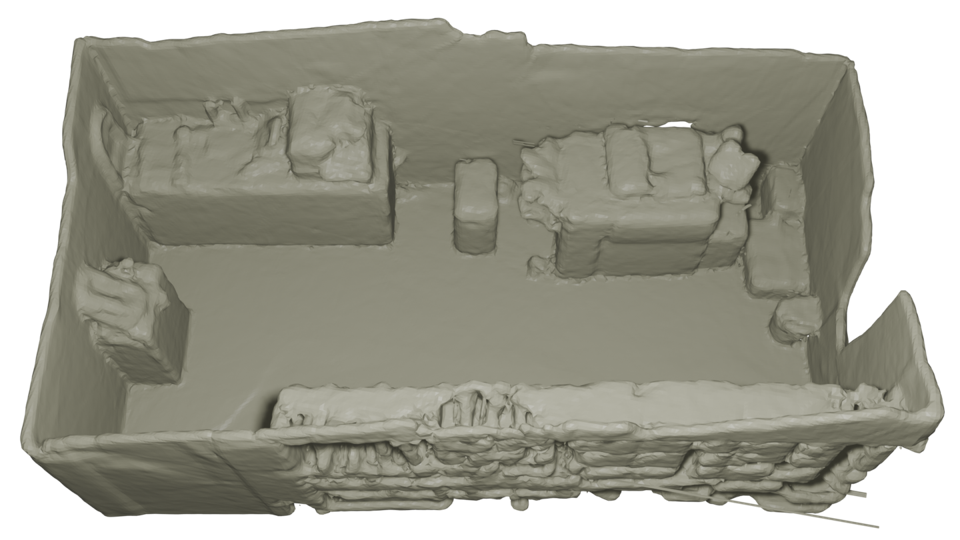}};
	\node[anchor=north east,inner sep=0, xshift=-12pt, yshift=-1pt] at (csp2.north east) {CSP};
	\node[anchor=north west, inner sep=0] (nsp2) 
	at (0.68\linewidth,-5.3) {\includegraphics[width=0.33\linewidth]{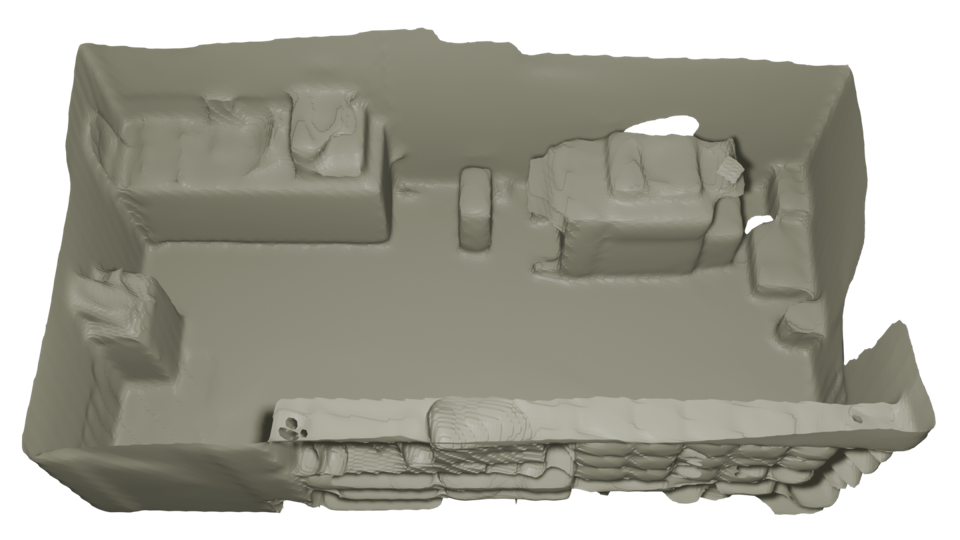}};
	\node[anchor=north east,inner sep=0, xshift=-12pt, yshift=-1pt] at (nsp2.north east) {\textbf{NSP}};
	
	\node[anchor=north west, inner sep=0] (input3) 
	at (0,-7.9) {\includegraphics[width=0.33\linewidth]{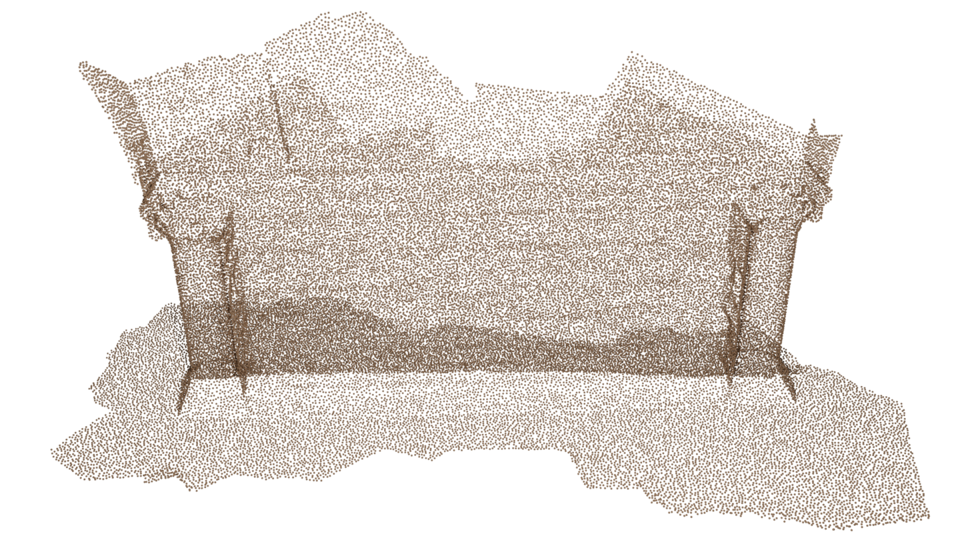}};
	\node[anchor=north east,inner sep=0, xshift=-15pt, yshift=-6pt] at (input3.north east) {Input};
	\node[anchor=north west, inner sep=0] (gt3) 
	at (0.34\linewidth,-7.9) {\includegraphics[width=0.33\linewidth]{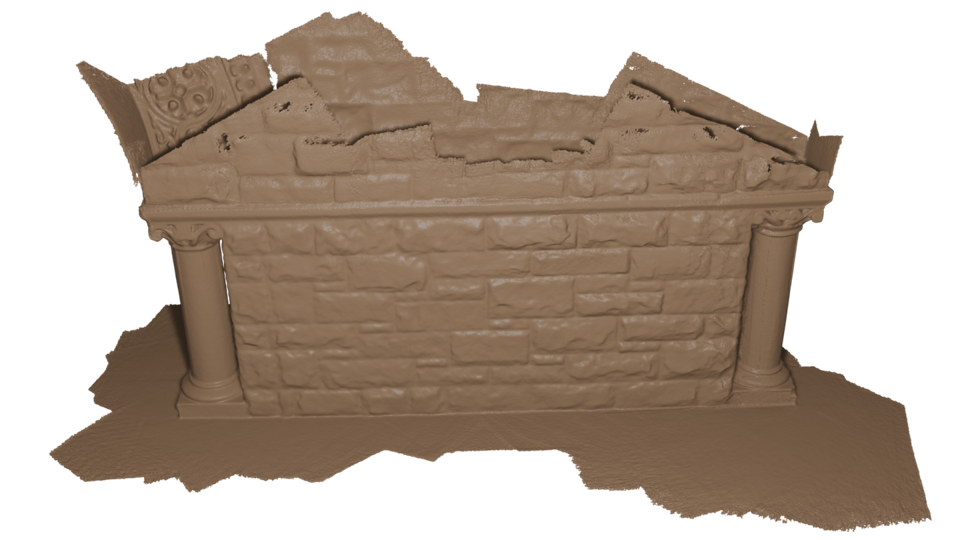}};
	\node[anchor=north east,inner sep=0, xshift=-15pt, yshift=-6pt] at (gt3.north east) {GT};
	\node[anchor=north west, inner sep=0] (ndf3) 
	at (0.68\linewidth,-7.9) {\includegraphics[width=0.33\linewidth]{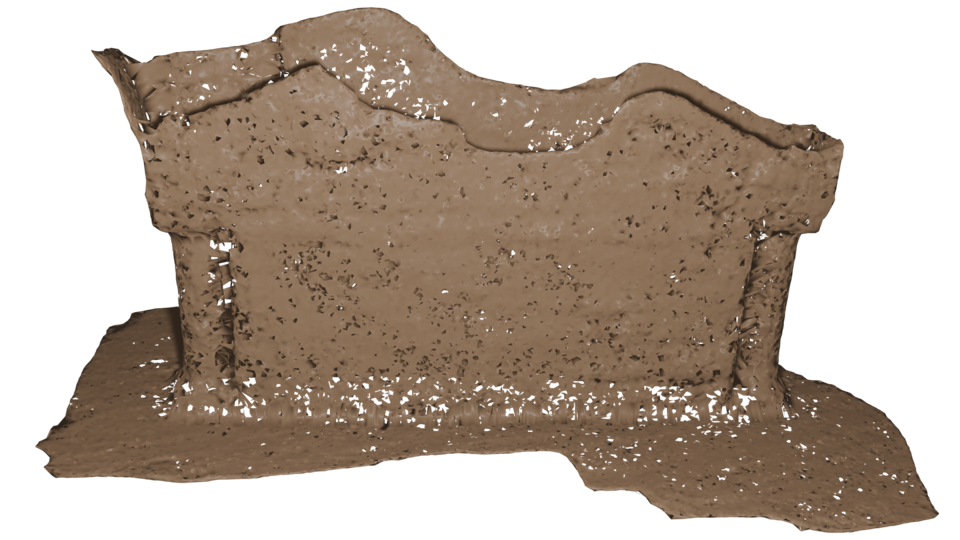}};
	\node[anchor=north east,inner sep=0, xshift=-15pt, yshift=-6pt] at (ndf3.north east) {NDF};
	
	\node[anchor=north west, inner sep=0] (cap3) 
	at (0,-10.2) {\includegraphics[width=0.33\linewidth]{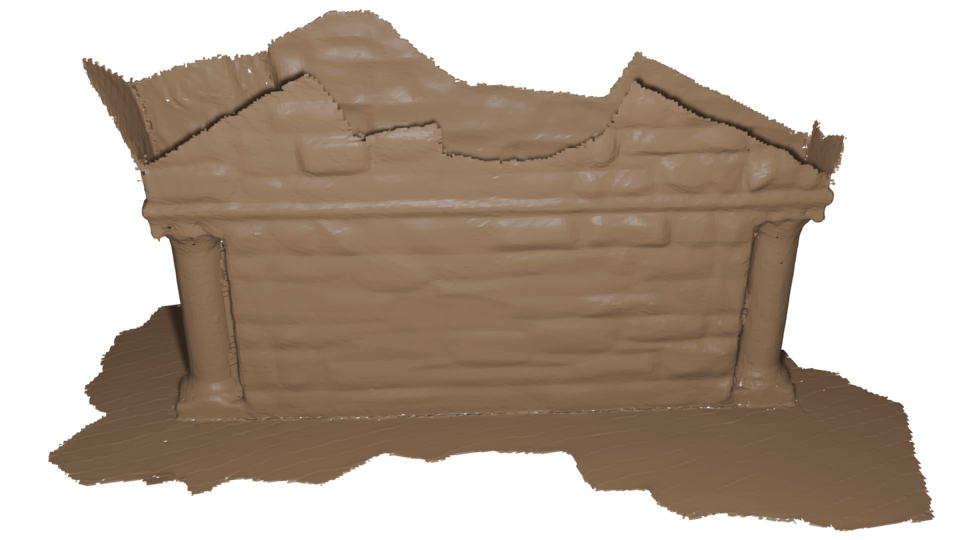}};
	\node[anchor=north east,inner sep=0, xshift=-15pt, yshift=-6pt] at (cap3.north east) {CAP};
	\node[anchor=north west, inner sep=0] (csp3) 
	at (0.34\linewidth,-10.2) {\includegraphics[width=0.33\linewidth]{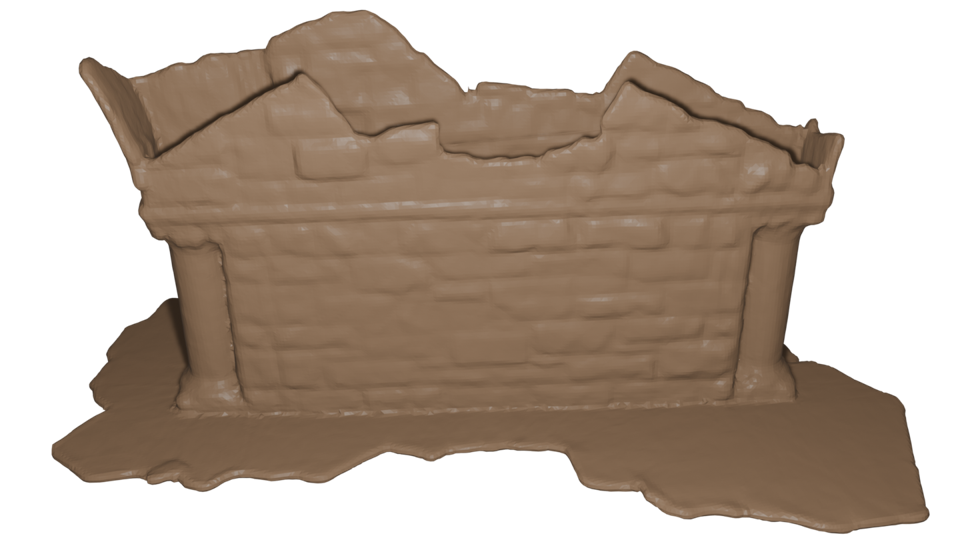}};
	\node[anchor=north east,inner sep=0, xshift=-15pt, yshift=-6pt] at (csp3.north east) {CSP};
	\node[anchor=north west, inner sep=0] (nsp3) 
	at (0.68\linewidth,-10.2) {\includegraphics[width=0.33\linewidth]{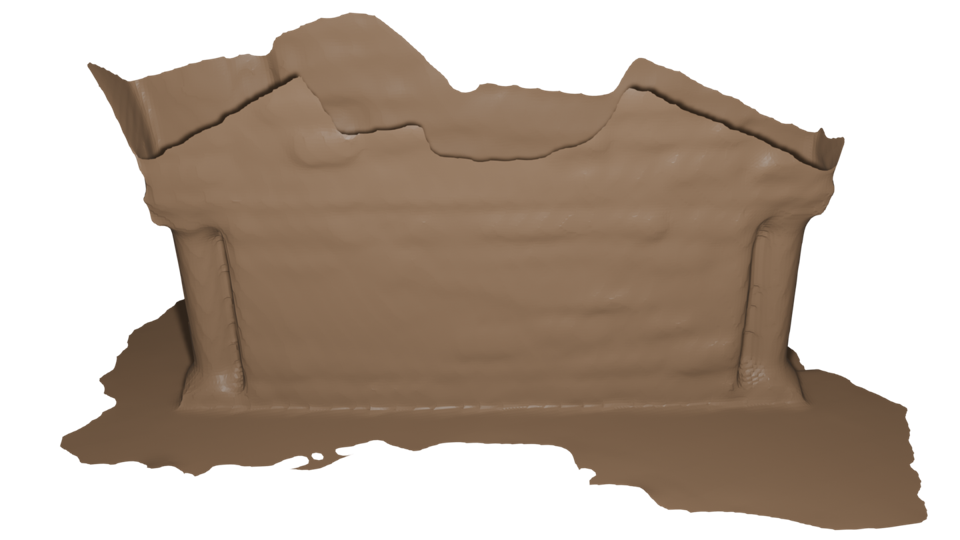}};
	\node[anchor=north east,inner sep=0, xshift=-15pt, yshift=-6pt] at (nsp3.north east) {\textbf{NSP}};
	\end{tikzpicture}
    \caption{Reconstructed surfaces on indoor scene data.}
    \label{fig:indoor}
\end{figure}
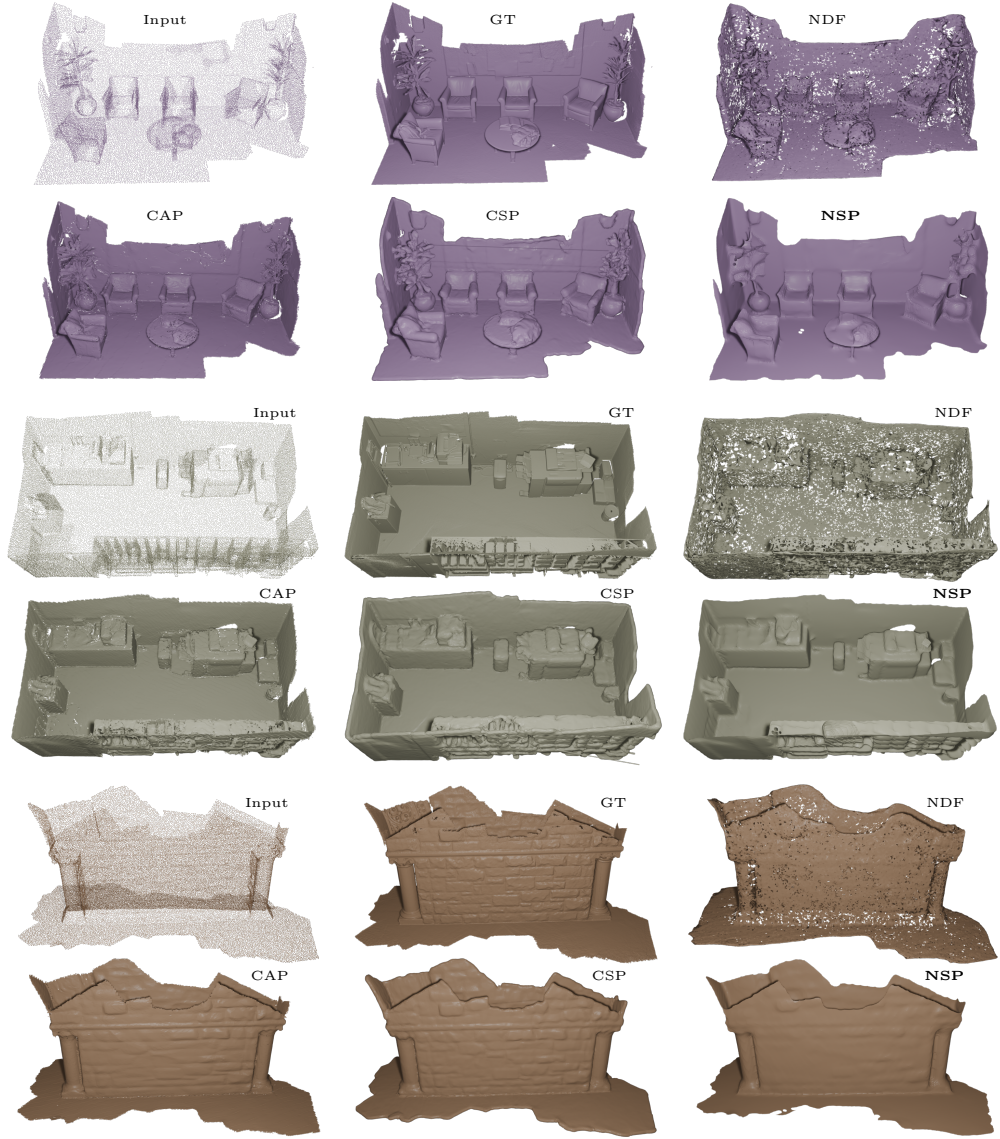

A comparison is made with the same baseline models considered in the previous section. The Chamfer and Hausdorff distances are employed as evaluation metrics. 
A summary of the results for these metrics is presented in Table \ref{tab:indoor}.
As evidenced in the table, the quantitative results of the proposed NSP are comparable to those of the competing methods, demonstrating its effectiveness. Reconstructed surfaces are presented in Figure \ref{fig:indoor}.
The results show that the baseline models tend to produce rough surfaces with holes.
While the proposed method may not capture all fine details, NSP effectively preserves the overall geometric features and reconstructs less irregular surfaces than baselines.
The reconstructed surfaces are clearly more refined, with better preservation of details and fewer artifacts, highlighting the superiority of our method in producing high-quality surface reconstructions from noisy and incomplete real scan point cloud data.

\subsection{Surface reconstruction for corrupted data}\label{sec:exp_corruption}
This section presents an evaluation of the robustness of the models against corruption of the point cloud.
Most existing distance-based INR models utilize metrics that quantify the proximity of an approximate surface to a given point cloud as their loss functions \citep{chibane2020neural, zhou2024cap}. Some approaches \citep{venkatesh2021deep} find the nearest points for sampled queries and use this information as a reference for the shortest path. Consequently, these methodologies heavily depend on the quality of the given point cloud. When the point cloud accurately captures the surface features without corruption, these methods can yield precise distance functions. However, in practical applications, point clouds are often incomplete due to noise during the scanning or extraction processes, resulting in missing parts and non-uniform distributions of points. Existing INR models typically treat these corrupted point clouds as clean during training, raising concerns about potential performance degradation.

In contrast, the proposed loss function \eqref{eq:loss_final} significantly reduces this reliance on the given point cloud. Aside from the manifold loss \eqref{eq:mnfld_loss} that enforces minimizing the distance function at the given point cloud, the point cloud is not further incorporated into the loss function. Instead of calculating metrics directly or finding the nearest points to the point cloud, we have devised a novel SP loss \eqref{eq:SP_loss} that encourages points pulled by the shortest path to lie within the zero level set of the learned distance function. This approach is expected to be more robust against corrupted point clouds compared to previous methods.

\begin{figure}
    \centering
	\begin{tikzpicture}[every node/.style={font=\tiny}]
	\node[anchor=south west, inner sep=0] (input1) 
	at (0,0) {\includegraphics[width=0.19\linewidth]{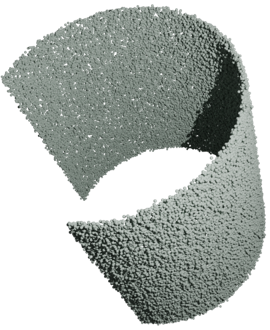}};
	\node[anchor=south west, inner sep=0] (ndf1) 
	at (0.20\linewidth,0) {\includegraphics[width=0.19\linewidth]{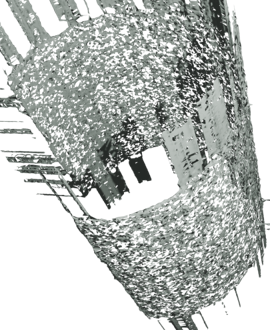}};
	\node[anchor=south west, inner sep=0] (cap1) 
	at (0.40\linewidth,0) {\includegraphics[width=0.19\linewidth]{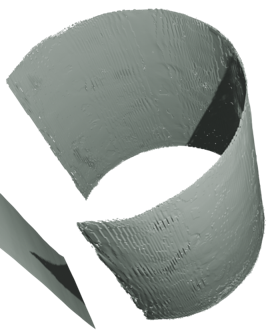}};
	\node[anchor=south west, inner sep=0] (csp1) 
	at (0.60\linewidth,0) {\includegraphics[width=0.19\linewidth]{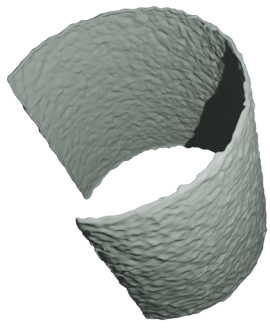}};
	\node[anchor=south west, inner sep=0] (nsp1) 
	at (0.80\linewidth,0) {\includegraphics[width=0.19\linewidth]{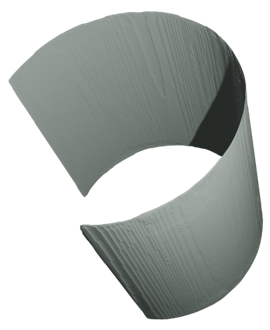}};
	
	\node[anchor=south west, inner sep=0] (input2) 
	at (0,-3.1) {\includegraphics[width=0.19\linewidth]{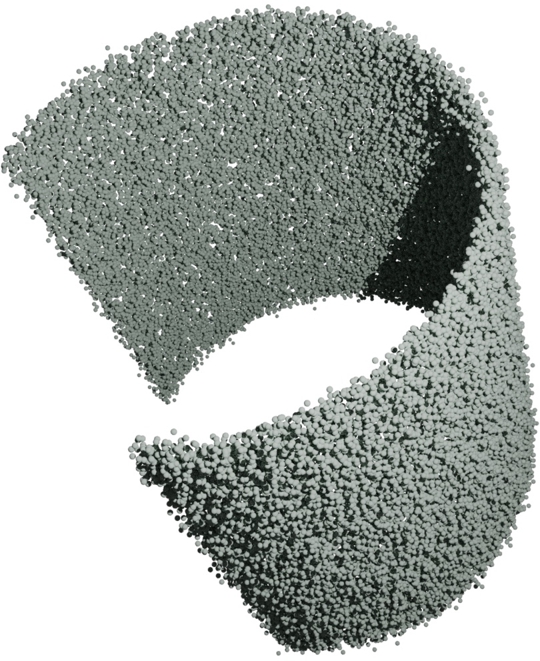}};
	\node[anchor=south west, inner sep=0] (ndf2) 
	at (0.20\linewidth,-3.1) {\includegraphics[width=0.19\linewidth]{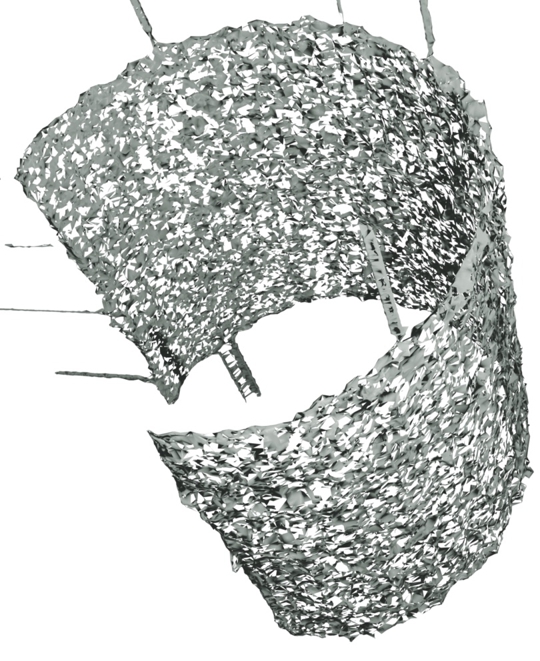}};
	\node[anchor=south west, inner sep=0] (cap2) 
	at (0.40\linewidth,-3.1) {\includegraphics[width=0.19\linewidth]{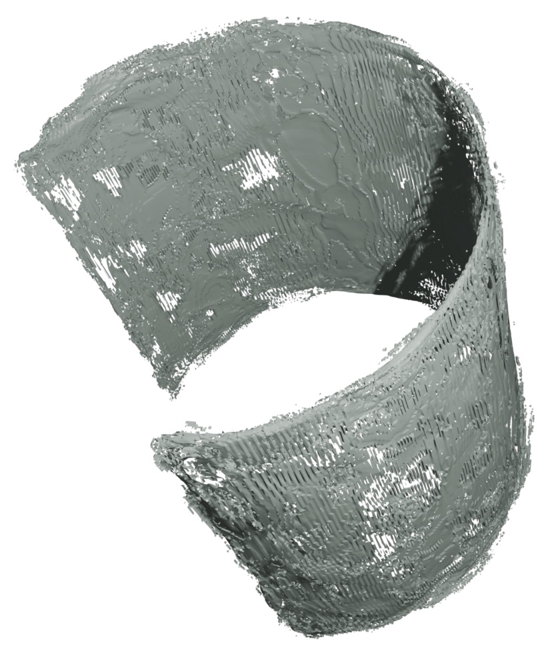}};
	\node[anchor=south west, inner sep=0] (csp2) 
	at (0.60\linewidth,-3.1) {\includegraphics[width=0.19\linewidth]{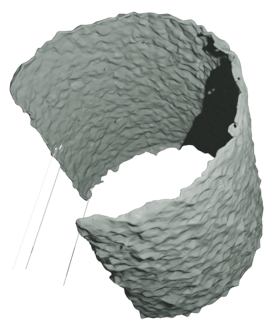}};
	\node[anchor=south west, inner sep=0] (nsp2) 
	at (0.80\linewidth,-3.1) {\includegraphics[width=0.19\linewidth]{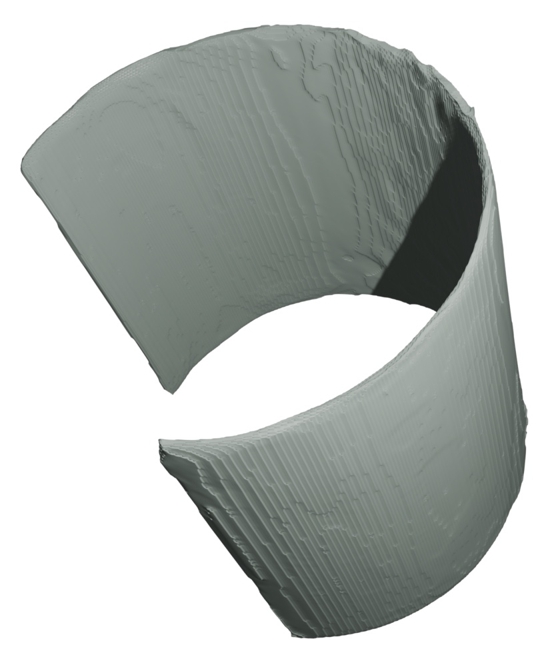}};
	
	\node[anchor=south west, inner sep=0] (input3) 
	at (0,-6.2) {\includegraphics[width=0.19\linewidth]{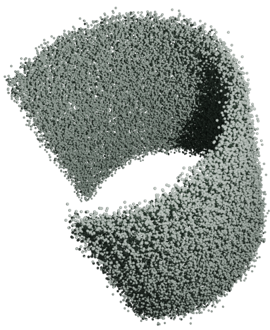}};
	\node[anchor=south west, inner sep=0] (ndf3) 
	at (0.20\linewidth,-6.2) {\includegraphics[width=0.19\linewidth]{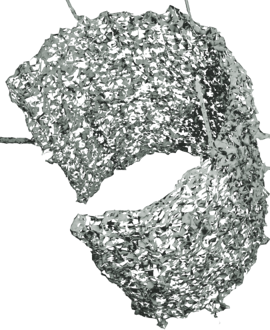}};
	\node[anchor=south west, inner sep=0] (cap3) 
	at (0.40\linewidth,-6.2) {\includegraphics[width=0.19\linewidth]{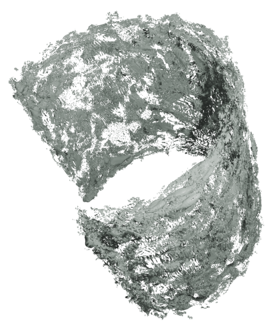}};
	\node[anchor=south west, inner sep=0] (csp3) 
	at (0.60\linewidth,-6.2) {\includegraphics[width=0.19\linewidth]{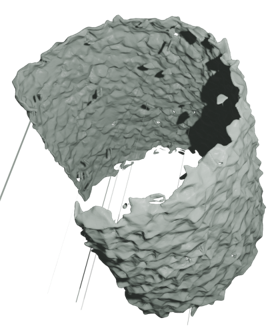}};
	\node[anchor=south west, inner sep=0] (nsp3) 
	at (0.80\linewidth,-6.2) {\includegraphics[width=0.19\linewidth]{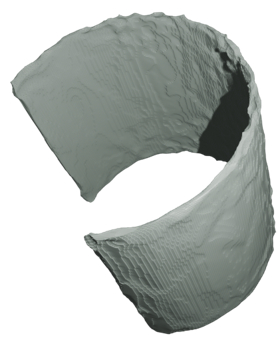}};

	\node[rotate=90, anchor=north] at ($(input1.west)!0.5!(input1.west) + (0,-0.5)$) {$\sigma=0.005$};
	\node[rotate=90, anchor=north] at ($(input2.west)!0.5!(input2.west) + (0,-0.6)$) {$\sigma=0.01$};
	\node[rotate=90, anchor=north] at ($(input3.west)!0.5!(input3.west) + (0,-0.7)$) {$\sigma=0.02$};
	
	\node[anchor=north,inner sep=0pt, yshift=-2pt] at (input3.south) {Input};
	\node[anchor=north,inner sep=0pt, yshift=-2pt] at (ndf3.south) {NDF};
	\node[anchor=north,inner sep=0pt, yshift=-2pt] at (cap3.south) {CAP};
	\node[anchor=north,inner sep=0pt, yshift=-2pt] at (csp3.south) {CSP};
	\node[anchor=north,inner sep=0pt, yshift=-2pt] at (nsp3.south) {\textbf{NSP}};
	\end{tikzpicture} 
    \caption{Reconstructed surfaces from noisy point cloud (shown in the leftmost column) on a partial cylinder. Three levels of additive Gaussian noise with standard deviations $\sigma=0.005$, $0.01$, and $0.02$ are considered.}
    \label{fig:noise_cylinder}
\end{figure}
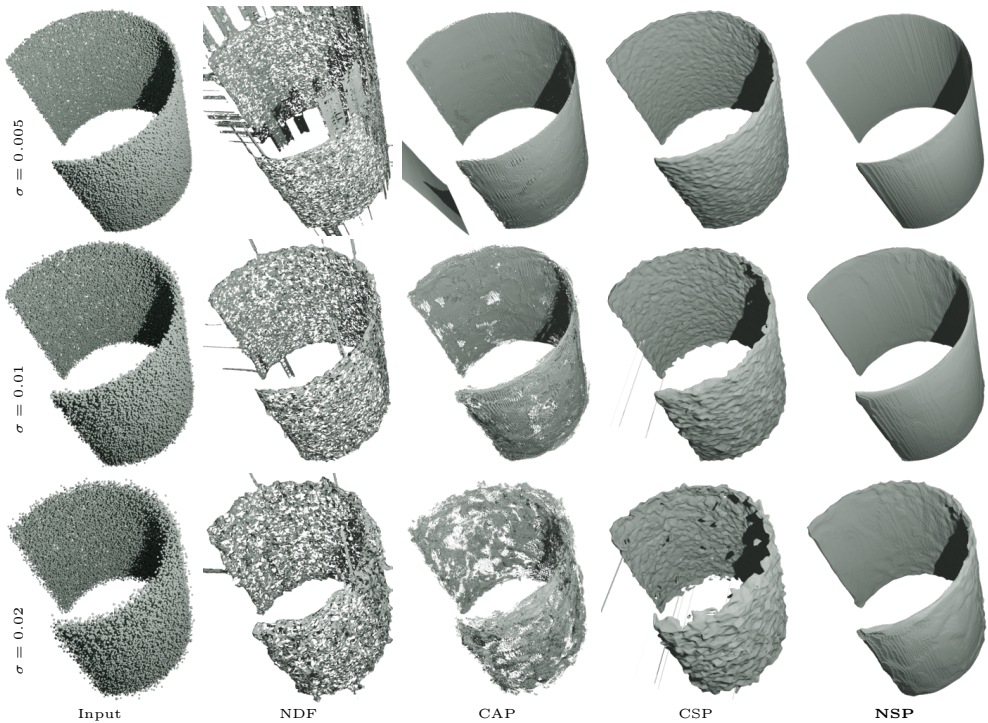

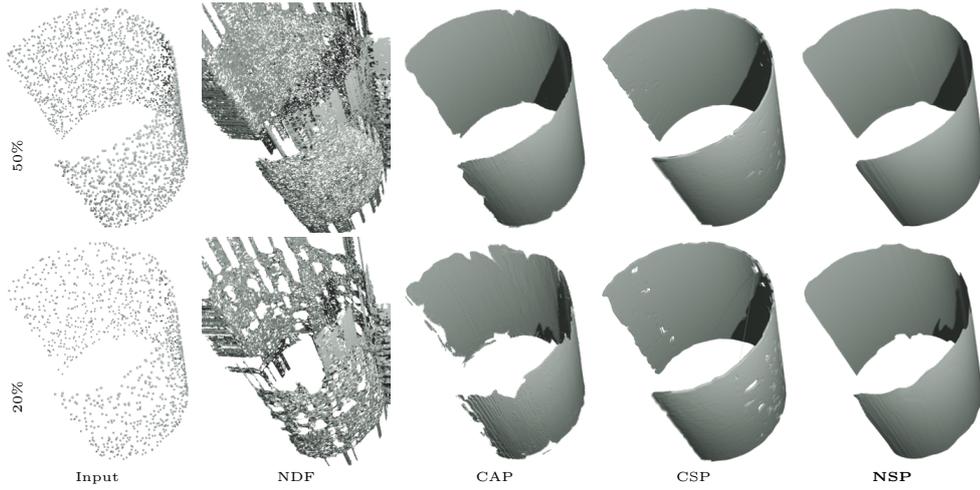
\begin{figure}
    \centering 
	\begin{tikzpicture}[every node/.style={font=\tiny}]
	\node[anchor=south west, inner sep=0] (input1) 
	at (0,0) {\includegraphics[width=0.19\linewidth]{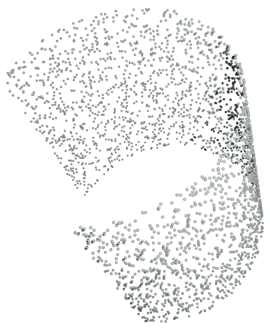}};
	\node[anchor=south west, inner sep=0] (ndf1) 
	at (0.20\linewidth,0) {\includegraphics[width=0.19\linewidth]{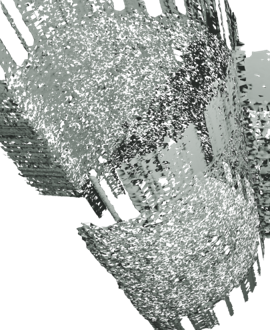}};
	\node[anchor=south west, inner sep=0] (cap1) 
	at (0.40\linewidth,0) {\includegraphics[width=0.19\linewidth]{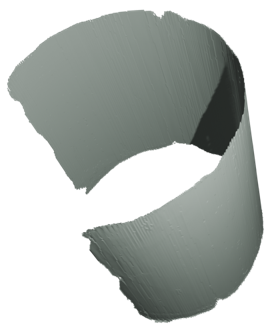}};
	\node[anchor=south west, inner sep=0] (csp1) 
	at (0.60\linewidth,0) {\includegraphics[width=0.19\linewidth]{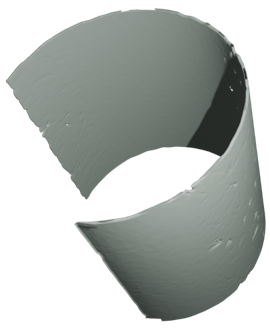}};
	\node[anchor=south west, inner sep=0] (nsp1) 
	at (0.80\linewidth,0) {\includegraphics[width=0.19\linewidth]{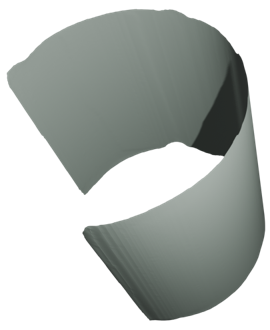}};
	
	\node[anchor=south west, inner sep=0] (input2) 
	at (0,-3.1) {\includegraphics[width=0.19\linewidth]{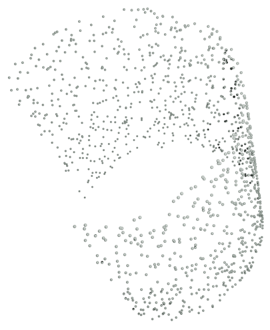}};
	\node[anchor=south west, inner sep=0] (ndf2) 
	at (0.20\linewidth,-3.1) {\includegraphics[width=0.19\linewidth]{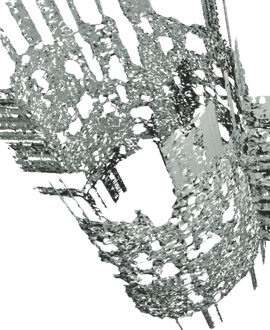}};
	\node[anchor=south west, inner sep=0] (cap2) 
	at (0.40\linewidth,-3.1) {\includegraphics[width=0.19\linewidth]{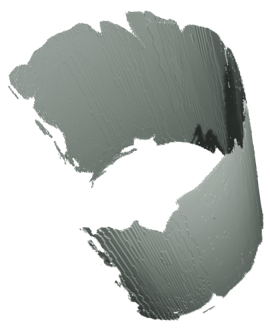}};
	\node[anchor=south west, inner sep=0] (csp2) 
	at (0.60\linewidth,-3.1) {\includegraphics[width=0.19\linewidth]{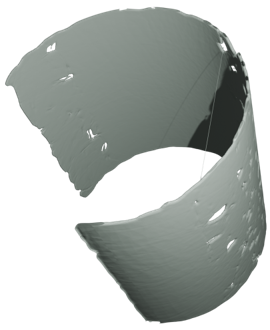}};
	\node[anchor=south west, inner sep=0] (nsp2) 
	at (0.80\linewidth,-3.1) {\includegraphics[width=0.19\linewidth]{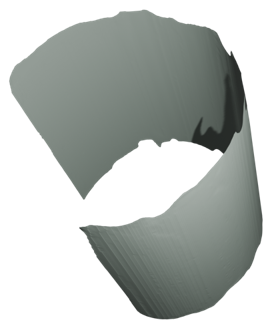}};
	
	\node[rotate=90, anchor=north] at ($(input1.west)!0.5!(input1.west) + (0,-0.5)$) {$50\%$};
	\node[rotate=90, anchor=north] at ($(input2.west)!0.5!(input2.west) + (0,-0.6)$) {$20\%$};
	
	\node[anchor=north,inner sep=0pt, yshift=-2pt] at (input2.south) {Input};
	\node[anchor=north,inner sep=0pt, yshift=-2pt] at (ndf2.south) {NDF};
	\node[anchor=north,inner sep=0pt, yshift=-2pt] at (cap2.south) {CAP};
	\node[anchor=north,inner sep=0pt, yshift=-2pt] at (csp2.south) {CSP};
	\node[anchor=north,inner sep=0pt, yshift=-2pt] at (nsp2.south) {\textbf{NSP}};
	\end{tikzpicture}     
    \caption{Reconstructed surfaces from sparse observations on a partial cylinder with sparsities of $50\%$ and $20\%$. Figures in the leftmost column illustrate the input sparse point clouds.}
    \label{fig:sparse_cylinder}
\end{figure}

\begin{figure}
    \centering 
	\begin{tikzpicture}[every node/.style={font=\tiny}]
	\node[anchor=south west, inner sep=0] (input1) 
	at (0,0) {\includegraphics[width=0.19\linewidth]{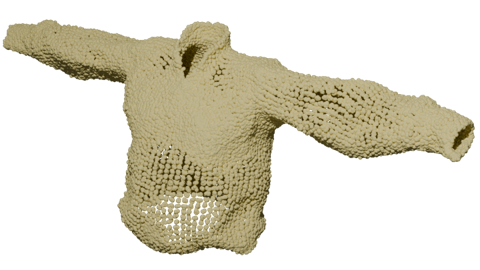}};
	\node[anchor=south west, inner sep=0] (ndf1) 
	at (0.20\linewidth,0) {\includegraphics[width=0.19\linewidth]{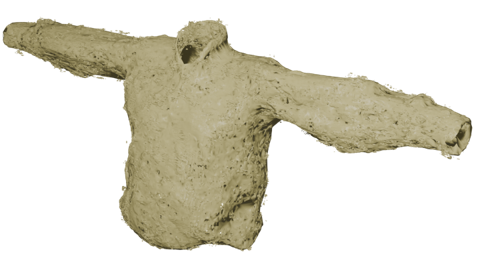}};
	\node[anchor=south west, inner sep=0] (cap1) 
	at (0.40\linewidth,0) {\includegraphics[width=0.19\linewidth]{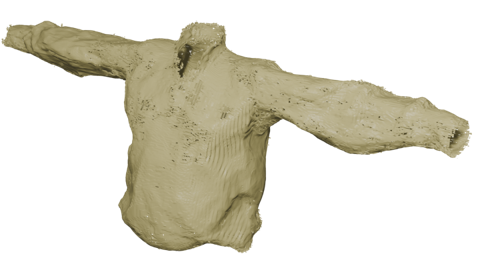}};
	\node[anchor=south west, inner sep=0] (csp1) 
	at (0.60\linewidth,0) {\includegraphics[width=0.19\linewidth]{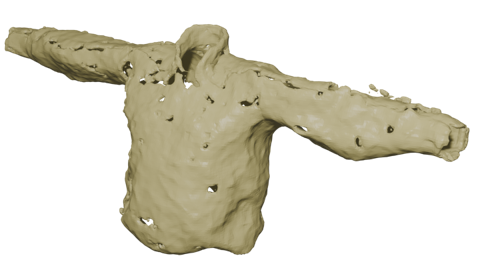}};
	\node[anchor=south west, inner sep=0] (nsp1) 
	at (0.80\linewidth,0) {\includegraphics[width=0.19\linewidth]{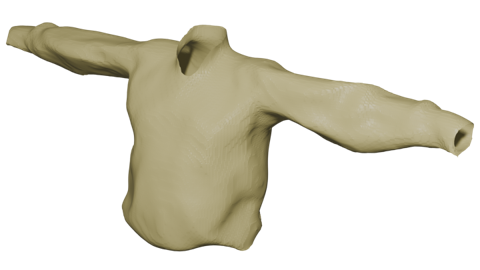}};
	
	\node[anchor=south west, inner sep=0] (input2) 
	at (0,-1.5) {\includegraphics[width=0.19\linewidth]{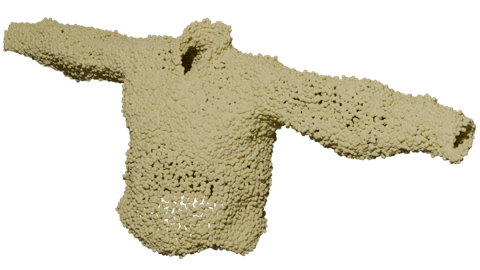}};
	\node[anchor=south west, inner sep=0] (ndf2) 
	at (0.20\linewidth,-1.5) {\includegraphics[width=0.19\linewidth]{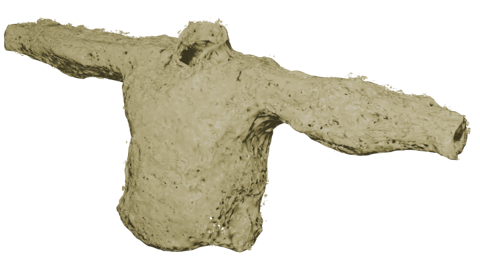}};
	\node[anchor=south west, inner sep=0] (cap2) 
	at (0.40\linewidth,-1.5) {\includegraphics[width=0.19\linewidth]{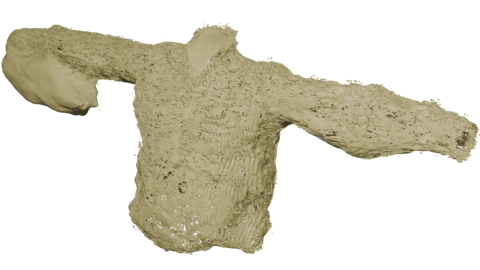}};
	\node[anchor=south west, inner sep=0] (csp2) 
	at (0.60\linewidth,-1.5) {\includegraphics[width=0.19\linewidth]{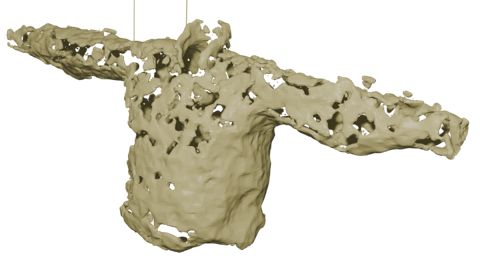}};
	\node[anchor=south west, inner sep=0] (nsp2) 
	at (0.80\linewidth,-1.5) {\includegraphics[width=0.19\linewidth]{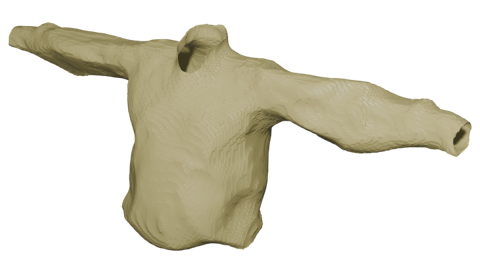}};
	
	\node[anchor=south west, inner sep=0] (input3) 
	at (0,-3) {\includegraphics[width=0.19\linewidth]{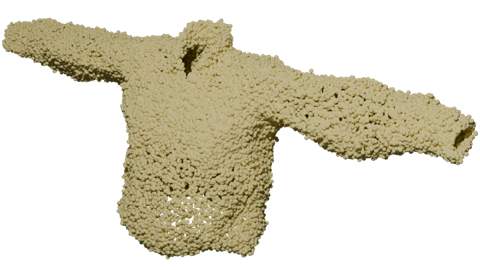}};
	\node[anchor=south west, inner sep=0] (ndf3) 
	at (0.20\linewidth,-3) {\includegraphics[width=0.19\linewidth]{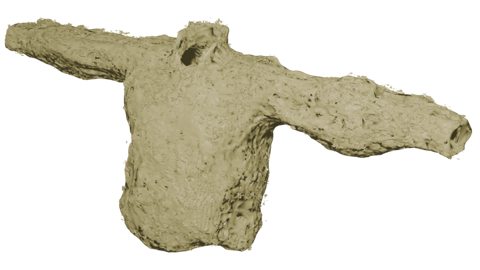}};
	\node[anchor=south west, inner sep=0] (cap3) 
	at (0.40\linewidth,-3) {\includegraphics[width=0.19\linewidth]{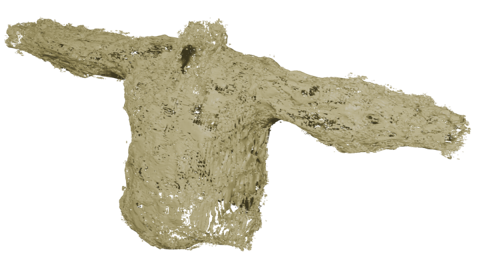}};
	\node[anchor=south west, inner sep=0] (csp3) 
	at (0.60\linewidth,-3) {\includegraphics[width=0.19\linewidth]{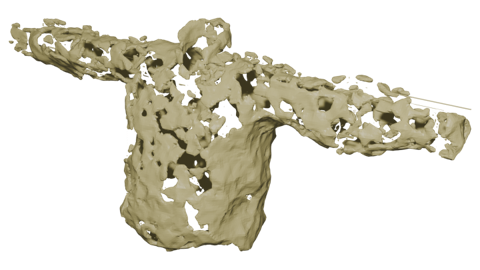}};
	\node[anchor=south west, inner sep=0] (nsp3) 
	at (0.80\linewidth,-3) {\includegraphics[width=0.19\linewidth]{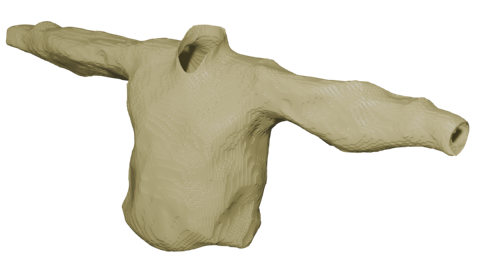}};
	
	\node[rotate=-15, anchor=north, xshift=-13pt, yshift=-7pt] at (input1.north east) {$\sigma=0.002$};
	\node[rotate=-15, anchor=north, xshift=-13pt, yshift=-7pt] at (input2.north east) {$\sigma=0.004$};
	\node[rotate=-15, anchor=north, xshift=-13pt, yshift=-7pt] at (input3.north east) {$\sigma=0.005$};
	
	\node[rotate=-15, anchor=north, xshift=-8pt, yshift=-27pt] at (input3.north east) {Input};
	\node[rotate=-15, anchor=north, xshift=-8pt, yshift=-27pt] at (ndf3.north east) {NDF};
	\node[rotate=-15, anchor=north, xshift=-8pt, yshift=-27pt] at (cap3.north east) {CAP};
	\node[rotate=-15, anchor=north, xshift=-8pt, yshift=-27pt] at (csp3.north east) {CSP};
	\node[rotate=-15, anchor=north, xshift=-8pt, yshift=-27pt] at (nsp3.north east) {\textbf{NSP}};
	\end{tikzpicture} 
    \caption{Reconstructed surfaces from noisy point cloud (shown in the leftmost column) on a shirt data. Three levels of additive Gaussian noise with standard deviations $\sigma=0.002$, $0.004$, and $0.005$ are considered.}
    \label{fig:noise_shirt}
\end{figure}
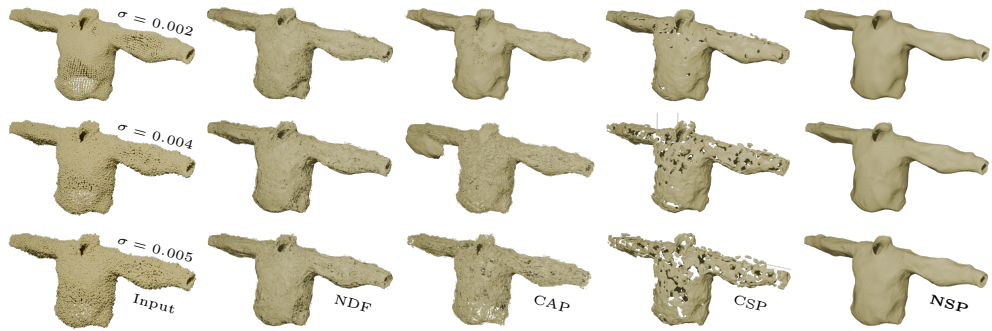

\begin{figure}
    \centering
	\begin{tikzpicture}[every node/.style={font=\tiny}]
	\node[anchor=south west, inner sep=0] (input1) 
	at (0,0) {\includegraphics[width=0.19\linewidth]{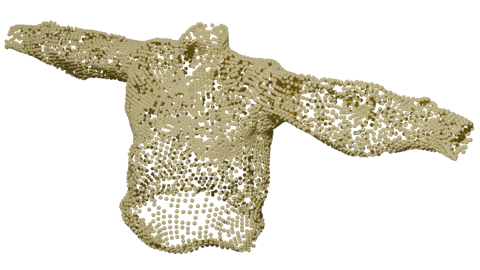}};
	\node[anchor=south west, inner sep=0] (ndf1) 
	at (0.20\linewidth,0) {\includegraphics[width=0.19\linewidth]{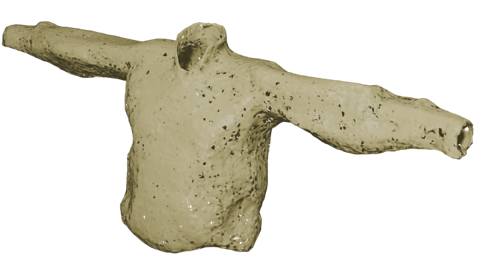}};
	\node[anchor=south west, inner sep=0] (cap1) 
	at (0.40\linewidth,0) {\includegraphics[width=0.19\linewidth]{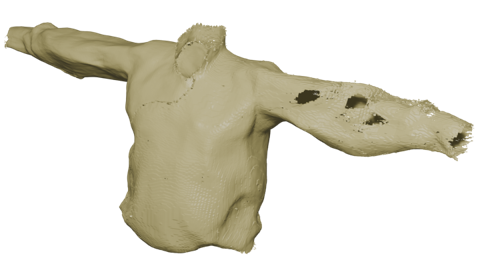}};
	\node[anchor=south west, inner sep=0] (csp1) 
	at (0.60\linewidth,0) {\includegraphics[width=0.19\linewidth]{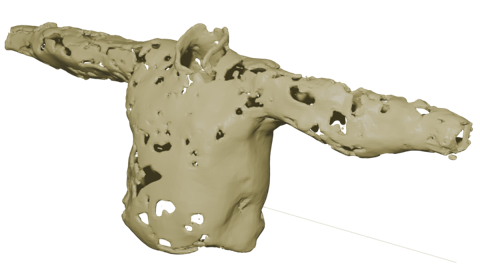}};
	\node[anchor=south west, inner sep=0] (nsp1) 
	at (0.80\linewidth,0) {\includegraphics[width=0.19\linewidth]{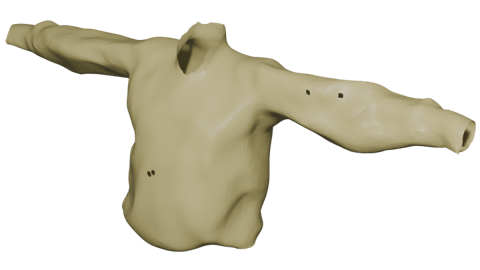}};
	
	\node[anchor=south west, inner sep=0] (input2) 
	at (0,-1.5) {\includegraphics[width=0.19\linewidth]{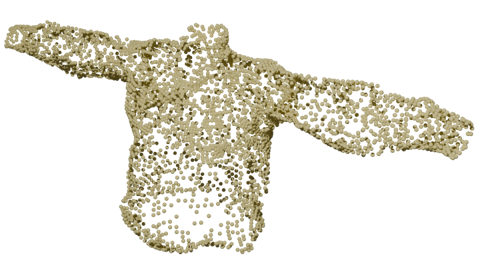}};
	\node[anchor=south west, inner sep=0] (ndf2) 
	at (0.20\linewidth,-1.5) {\includegraphics[width=0.19\linewidth]{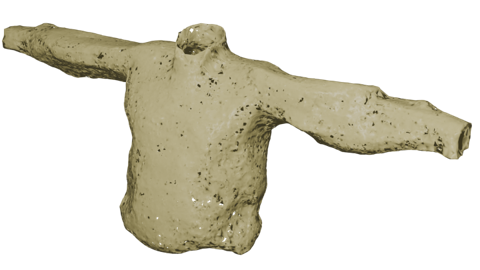}};
	\node[anchor=south west, inner sep=0] (cap2) 
	at (0.40\linewidth,-1.5) {\includegraphics[width=0.19\linewidth]{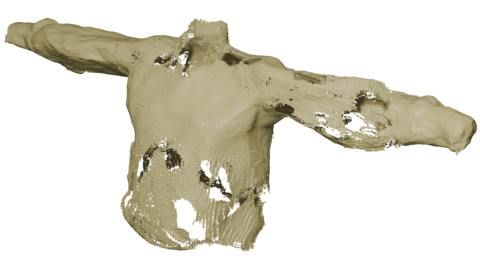}};
	\node[anchor=south west, inner sep=0] (csp2) 
	at (0.60\linewidth,-1.5) {\includegraphics[width=0.19\linewidth]{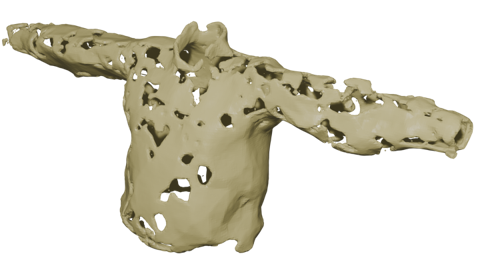}};
	\node[anchor=south west, inner sep=0] (nsp2) 
	at (0.80\linewidth,-1.5) {\includegraphics[width=0.19\linewidth]{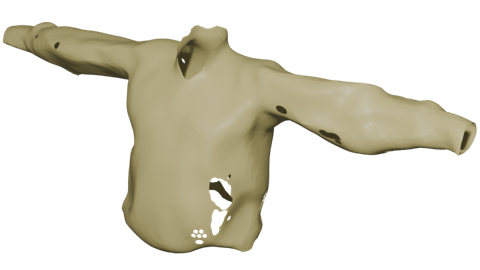}};
		
	\node[rotate=-15, anchor=north, xshift=-13pt, yshift=-7pt] at (input1.north east) {$20\%$};
	\node[rotate=-15, anchor=north, xshift=-13pt, yshift=-7pt] at (input2.north east) {$10\%$};
	
	\node[rotate=-15, anchor=north, xshift=-8pt, yshift=-27pt] at (input2.north east) {Input};
	\node[rotate=-15, anchor=north, xshift=-8pt, yshift=-27pt] at (ndf2.north east) {NDF};
	\node[rotate=-15, anchor=north, xshift=-8pt, yshift=-27pt] at (cap2.north east) {CAP};
	\node[rotate=-15, anchor=north, xshift=-8pt, yshift=-27pt] at (csp2.north east) {CSP};
	\node[rotate=-15, anchor=north, xshift=-8pt, yshift=-27pt] at (nsp2.north east) {\textbf{NSP}};
	\end{tikzpicture} 
    \caption{Reconstructed surfaces from sparse observations on a shirt data with sparsities of $20\%$ and $10\%$. Figures in the leftmost column illustrate the input sparse point clouds.}
    \label{fig:sparse_shirt}
\end{figure}
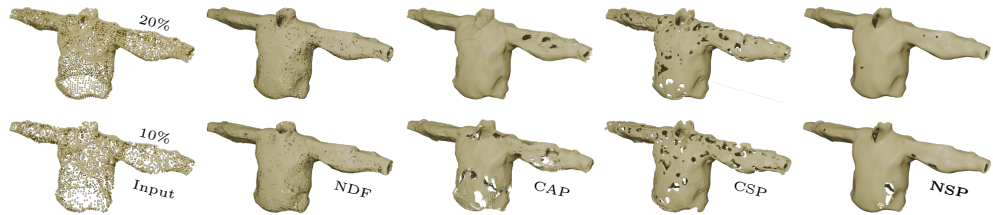

To validate our approach, we conducted experiments using noisy and sparse data, comparing our model against others, including NDF, CAP, and CSP. To effectively assess the impact of corruption, we selected data in which the models exhibited comparable reconstructions without introducing additional corruption. We chose simple datasets to clearly evaluate the influence of data corruption, specifically conducting experiments on the partial cylinder dataset discussed in Section \ref{sec:exp_toy} and the shirt data from the MGN dataset in Section \ref{sec:exp_synthetic}. 
To investigate the robustness to noise, we perturb the partial cylinder data with additive Gaussian noise with a mean of zero and three standard deviations of 0.005, 0.01, and 0.02, as well as deviations of $0.002$, $0.004$, and $0.005$ to the shirt data.
For sparsity, we uniform randomly sampled $50\%$ and $20\%$ of the points from the partial cylinder data, and $20\%$ and $10\%$ of the points from the shirt data.
The results are summarized in Figures \ref{fig:noise_cylinder}, \ref{fig:sparse_cylinder}, \ref{fig:noise_shirt}, and \ref{fig:sparse_shirt}.

The results demonstrate that the proposed method exhibits better robustness to noise and sparsity in comparison to the baseline models. The comparison models are easily affected by even small amounts of noise, tending to directly learn noise from the noisy data and incorporate it into the reconstructed surface. This behavior could be attributed to their loss functions, which directly employ the provided point clouds to learn the distance fields. Consequently, the corruption of the point cloud data, including any noise, has a strong influence on the learning process, leading to suboptimal reconstructions in the presence of noise. Similarly, the baselines show similar challenges when confronted with sparse data. They struggle to provide accurate reconstructions when the point clouds are incomplete or lack density, with noticeable degradation in the quality of the reconstructed surfaces.

In contrast, the proposed method demonstrates significantly more stable and smoother reconstructions, even in the presence of noise or sparse data. This improvement highlights the effectiveness of the novel loss function, which does not directly rely on the given point cloud but instead learns the distance function from the ESP property. By decoupling the reconstruction from the noisy or sparse point cloud, the proposed method is able to produce more accurate and robust surface reconstructions, thereby showing its superior performance in handling challenging data that arise in real-world situations.

\section{Conclusion}
We introduced the neural shortest path (NSP), a vector-valued INR approach of surface reconstruction that simultaneously approximates the distance function and its gradient. By parameterizing the NSP and leveraging a variable splitting method, we ensured the convergence of the magnitude of NSP in the $H^1$ norm. We also proposed a new loss function, proving that the exact shortest path corresponds to its global minimum. The NSP addresses key limitations of existing INR models, such as inaccurate gradient estimation and heavy dependency on point cloud data. Unlike to classical methods, the NSP uses the advantages of deep learning to operate without a mesh and to efficiently solve highly nonlinear minimization problems, thereby offering enhanced flexibility and efficiency. Extensive experiments demonstrate that the NSP outperforms existing INR methods in the distance function and gradient learning, achieving high-quality surface reconstructions with smoother surfaces and better preservation of geometric features, particularly for complex shapes like garments, cars, and 3D indoor scenes. The NSP also shows strong robustness to noisy and sparse point clouds, making it well-suited for real-world applications where data quality is often compromised.

While the proposed NSP demonstrates theoretical and practical improvements in surface reconstruction, there remain some limitations. Specifically, while the NSP successfully generates smooth surfaces, it still has difficulties to capture fine details with the desired level of accuracy. A potential avenue for future research would be to refine the method in order to preserve surface smoothness while more accurately capturing intricate geometric details. Additionally, point clouds are typically imperfect, which presents a challenge in distinguishing between regions that should correspond to holes or missing parts of the surface. At present, the NSP does not take this distinction into account. Addressing this issue would require the introduction of supplementary guidance as a regularization term, which could replace the existing area loss function. 
Furthermore, further investigation is required regarding the optimization of the weights between loss terms, as this could lead to more effective surface reconstruction in various scenarios.

\section{Acknowledgements}
All authors would like to thank Dr. Seong Weon Jeong, CEO, and Dr. Cheol-O Ahn, CTO of \textsc{Metariver Technology Co., Ltd}, for providing the advanced visualization tool, \texttt{SAMADII+}. This tool was used in checking surface meshes from selected points for the assessment of ongoing results until we developed the method in Section~\ref{subsec:extract_alg}.

This work was supported by the National Research Foundation of Korea under Grants [2021R1A2C3010887, RS-2024-00406127, RS-2024-00343226, RS-2024-00421203] and the Ministry of Science and ICT R\&D program of MSIT/IITP[2021-0-00077].
This project No. 2140/01/01 has received funding from the European Union´s Horizon 2020 research and innovation programme under the Marie Sk{\l}odowska-Curie grant agreement No. 945478.

\newpage
\appendix
\section{Detailed Proofs}
\subsection{Proof of Theorem \ref{thm:Sobolev_convergence}}\label{appen:pf_Sobolev}
Let $F_n$ converges to $F$ a.e. in $\Omega$. Since $F_n$ is nonzero a.e. in $\Omega$, the limit point $F$ is also nonzero a.e., and hence its MDD is well-defined a.e. in $\Omega$. Since $d_n$ is defined by the magnitude of $F_n$, $d_n$ converges to $d\coloneqq\left\Vert F\right\Vert$ a.e. in $\Omega$, because
\begin{equation}
	\left\vert d_n-d\right\vert=
	\Bigl\vert\left\Vert F_n\right\Vert - \left\Vert F\right\Vert \Bigr\vert \leq \left\Vert F_n-F\right\Vert\rightarrow0,
\end{equation}
as $n\rightarrow\infty$ a.e. Reverse triangle inequality is used in the last inequality.

For examining the convergence of $G_n$, let us remind $G_n=\frac{F_n}{d_n}$.
Since $d_n$ and $d$ are nonzero a.e., we can deduce that
\begin{align}
	\left\Vert\frac{F_n}{d_n}-\frac{F}{d}\right\Vert &= \frac{\left\Vert dF_n-d_nF\right\Vert}{\left\vert d_nd\right\vert} \nonumber \\
	& = \frac{\left\Vert d\left(F_n-F\right)-\left(d_n - d\right)F\right\Vert}{\left\vert d_nd\right\vert} \nonumber \\
	& \leq \frac{\left\vert d\right\vert\left\Vert F_n-F\right\Vert + \left\vert d_n - d\right\vert \left\Vert F\right\Vert}{\left\vert d_nd\right\vert}
	\rightarrow 0,
\end{align}
as $n\rightarrow\infty$ a.e. in $\Omega$.
By definition, $\left\Vert \frac{F_n}{d_n}\right\Vert = \left\Vert \frac{F_n}{\left\Vert F_n\right\Vert}\right\Vert=1$ a.e. in $\Omega$. Consequently, Lebesgue's dominated convergence theorem provides $\frac{F_n}{d_n}\rightarrow\frac{F}{d}$ in $L^2\left(\Omega\right)$. 
For the sake of notational simplicity, its limit point will henceforth be denoted by $G$.

If $\cL_{\text{GM}}\left(F_n\right)$ converges to zero, that is,
\[
\left\Vert \nabla d_n-G_n\right\Vert_{L^2\left(\Omega\right)}\rightarrow 0,
\]
as $n\rightarrow\infty$,
we have $\nabla d_n$ achieves $L^2$ convergence to the limit point of $G_n$.

Now, we show $d\in H^1\left(\Omega\right)$ with $\nabla d = G$.
For any test function $\phi\in C_c^\infty\left(\Omega\right)$, we have
\begin{align}
	\begin{split}
	\left\vert \int_\Omega G\phi\diff\bx - \int_\Omega \nabla d\phi\diff\bx\right\vert
	&= \left\vert\int_\Omega\left(G-\nabla d_n\right)\phi+\left(\nabla d_n-\nabla d\right)\phi\diff \bx\right\vert \nonumber\\
	&= \left\vert\int_\Omega\left(G-\nabla d_n\right)\phi+ \left(d-d_n\right)\nabla\phi\diff \bx\right\vert \nonumber\\
	&\leq \int_\Omega\Bigl\vert\left(G-\nabla d_n\right)\phi\Bigr\vert\diff \bx + \int_\Omega\Bigl\vert\left(d-d_n\right)\nabla\phi\Bigr\vert\diff \bx \nonumber\\
	&\leq \left\Vert G-\nabla d_n\right\Vert_{L^2\left(\Omega\right)}\left\Vert\phi\right\Vert_{L^2\left(\Omega\right)} + \left\Vert d-d_n\right\Vert_{L^2\left(\Omega\right)}\left\Vert \nabla\phi\right\Vert_{L^2\left(\Omega\right)} \nonumber\\
	& \leq C\left(\left\Vert G-\nabla d_n\right\Vert_{L^2\left(\Omega\right)}+\left\Vert d-d_n\right\Vert_{L^2\left(\Omega\right)}\right),
	\end{split}
\end{align}
for some constant $C=C\left(\phi,\Omega\right)$. Here, integration by parts is employed in the second equality. 
Since both $\left\Vert G-\nabla d_n\right\Vert_{L^2\left(\Omega\right)}$ and $\left\Vert d-d_n\right\Vert_{L^2\left(\Omega\right)}$ converge to zero as $n$ goes to infinity, we conclude that $\nabla d=G$.
Consequently, $d_n$ converges to $d\in H^1\left(\Omega\right)$ with $\nabla d=G$ in $H^1\left(\Omega\right)$ norm.

\subsection{Proof of Theorem \ref{thm:global_min}}\label{appen:pf_global_min}
Let $F^\ast$ be the optimal solution to \eqref{eq:minimization_NSP}. Since $F^\ast$ is nonzero a.e., its MDD is well-defined a.e. and we denote it by define $d^{\ast}= \left\Vert F^\ast\right\Vert$ and $G^\ast= F^\ast / \left\Vert F^\ast\right\Vert$. Set its zero level set by $\Gamma'\coloneqq\left\{\bx\in\Omega\mid d^\ast\left(\bx\right)=0\right\}$ and let $d$ be the distance function of $\Gamma'$.
Since $\cL_\Gamma\left(F^\ast\right)=0$, which implies $d^\ast=0$ a.e. in $\Gamma$, the target surface $\Gamma$ is contained in $\Gamma'$ except for the measure zero.

From $\cL_{\text{GM}}\left(F^\ast\right)=0$, $\nabla d^\ast=G^\ast$ and hence $\nabla d^\ast$ has the unit length a.e. in $\Omega$.
For any point $\bx\in\Omega$, the ESP property of the distance function writes that
\[
\bx-d\left(\bx\right)\nabla d\left(\bx\right)\in\Gamma',
\]
which implies
\[
d^\ast\left(\bx-d\left(\bx\right)\nabla d\left(\bx\right)\right)=0.
\]
By mean value property, we can deduce that
\begin{align}
	\begin{split}
	d^\ast\left(\bx\right) & = \left\vert 0-d^\ast\left(\bx\right)\right\vert \nonumber\\
	& = \left\vert d^\ast\left(\bx-d\left(\bx\right)\nabla d\left(\bx\right)\right) - d^\ast\left(\bx\right)\right\vert \nonumber\\
	& = \Bigl\vert d\left(\bx\right)\nabla d\left(\bx\right)^{\text{T}}\nabla d^\ast\left(\bx-\bv\right)\Bigr\vert \nonumber\\
	& \leq \left\vert d\left(\bx\right)\right\vert \left\Vert\nabla d\left(\bx\right)\right\Vert \left\Vert\nabla d^\ast\left(\bx-\bv\right)\right\Vert \nonumber\\
	& = d\left(\bx\right),
	\end{split}
\end{align}
for some $\bv\in\bR^3$ lies in the line segment between $\mathbf{0}$ and $d\left(\bx\right)\nabla d\left(\bx\right)$.
Therefore, we attain $d^\ast\left(\bx\right)\leq d\left(\bx\right)$ for almost every $\bx\in\Omega$.

For the reverse inequality, note that $d^\ast$ also project $\bx$ into $\Gamma'$ by
\[
\bx-d^\ast\left(\bx\right)\nabla d^\ast\left(\bx\right)\in\Gamma',
\]
since $d^\ast$ is the optimal solution of \eqref{eq:minimization_NSP}. 
This indicates that
\[
d\left(\bx-d^\ast\left(\bx\right)\nabla d^\ast\left(\bx\right)\right)=0.
\]
By interchanging the roles of $d$ and $d^\ast$ and applying the same calculations as above, we can derive the desired inequality $d\left(\bx\right)\leq d^\ast\left(\bx\right)$.

Combining both inequalities, we have that 
\[
d^\ast\left(\bx\right) = d\left(\bx\right),
\]
concluding the proof.

\section{Experimental Setup}\label{appen:exp_detail}
All experiments of NSP are conducted using an MLP \eqref{eq:MLP} with a depth of $L=6$ and a width of $512$, incorporating a skip connection at the third hidden layer and utlizing the softplus activation function $\sigma\left(x\right)=\frac{1}{\beta}\log\left(1+e^{\beta x}\right)$ with $\beta=100$, as in previous studies \citep{park2019deepsdf, gropp2020implicit, lipman2021phase}. The network is trained for for 60,000 epochs using the Adam optimizer \citep{kingma2014adam} with an initial learning rate of $10^{-3}$ decayed by a factor of $0.99$ every $2,000$ epochs. The proposed training loss $\cL _\text{total}$ \eqref{eq:loss_final} is a weighted sum of four loss terms with three positive constants $\lambda_{\text{SP}}$, $\lambda_{\text{GM}}$, and $\lambda_{\text{MA}}$. In all experiments, we use $\lambda_{\text{SP}}=0.01$ and $\lambda_{\text{GM}}=0.06$. For the MGN dataset \citep{bhatnagar2019multi}, $\lambda_{\text{MA}}=0.08$ is applied, while $\lambda_{\text{MA}}=0.15$ and $\lambda_{\text{MA}}=0.3$ are utilized for the ShapeNet \citep{shapenet2015} and indoor scene \citep{zhou2013dense} datasets, respectively. 
All derivatives included in the loss function are computed using an auto-differentiation library.
All terms of the training loss function were computed using Monte Carlo approximation. In each epoch, 20,000 points on the surface were uniformly randomly sampled from the given point cloud, and 20,000 collocation points for the computational domain were sampled from the domain with uniform distribution. In cases where the number of points in the given point cloud was fewer than 20,000, the entire set of points was used to calculate the manifold loss in each epoch.
The point cloud is dilated and translated to fit within the domain $\Omega=\left[-1,1\right]^3$. All numerical experiments are implemented on a single NVIDIA RTX 3090 GPU.

Experiments involving NDF \citep{chibane2020neural}, CAP \citep{zhou2024cap}, and CSP \citep{venkatesh2021deep} are conducted using their official codes\footnote{\label{ndf}\url{https://github.com/jchibane/ndf}}
\footnote{\label{cap}\url{https://github.com/junshengzhou/CAP-UDF}}
\footnote{\label{csp}\url{https://github.com/rahulvenkk/csp-net}}(MIT License) and adhering closely to the configurations outlined in the respective papers. Ball pivoting was employed for mesh extraction in CSP, as it performed better than the original mesh extraction algorithm.
\bibliography{mybib}

\end{document}